\documentclass[lettersize,journal]{IEEEtran}
\usepackage{amsmath,amsfonts,amssymb}
\usepackage{algorithmic}
\usepackage{algorithm}
\usepackage{array}
\usepackage[caption=false,font=normalsize,labelfont=sf,textfont=sf]{subfig}
\usepackage{textcomp}
\usepackage{stfloats}
\usepackage{url}
\usepackage{verbatim}
\usepackage{graphicx}
\usepackage{cite}
\usepackage{color}
\usepackage{hyperref}
\usepackage{relsize}
\usepackage{tabularx}
\usepackage{multirow}
\usepackage{booktabs}
\usepackage{diagbox}

\newcommand\blfootnote[1]{%
  \begingroup
  \renewcommand\thefootnote{}\footnote{#1}%
  \addtocounter{footnote}{-1}%
  \endgroup
}

\definecolor{red}{rgb}{1,0.2,0.2}
\definecolor{or}{rgb}{1,0.5,0.25}
\definecolor{green}{rgb}{0, 1, 0}
\definecolor{bl}{rgb}{0, 0, 1}
\definecolor{teal}{rgb}{0, 0.7, 0.6}

\begin{document}

\title{PRIMEdit: Probability Redistribution for\\Instance-aware Multi-object Video Editing\\with a New Benchmark Dataset}

\author{Samuel~Teodoro\textsuperscript{1*}, Agus~Gunawan\textsuperscript{1*}, Soo~Ye~Kim\textsuperscript{2}, Jihyong~Oh\textsuperscript{3\textdagger}, and Munchurl~Kim\textsuperscript{1\textdagger},~\IEEEmembership{Senior Member,~IEEE,}

\textsuperscript{1} School of Electrical Engineering, Korea Advanced Institute of Science and Technology \\
\textsuperscript{2} Adobe Research, San Jose, California \\
\textsuperscript{3} Department of Imaging Science, GSAIM, Chung-Ang University \\
}

\markboth{Journal of \LaTeX\ Class Files,~Vol.~14, No.~8, August~2021}%
{Shell \MakeLowercase{\textit{et al.}}: A Sample Article Using IEEEtran.cls for IEEE Journals}

\IEEEpubid{\begin{minipage}{\textwidth}\ \\[12pt]
\footnotesize
\copyright~2026 IEEE. Personal use of this material is permitted. Permission from IEEE must be obtained for all other uses, in any current or future media, including reprinting/republishing this material for advertising or promotional purposes, creating new collective works, for resale or redistribution to servers or lists, or reuse of any copyrighted component of this work in other works.\\
Published in IEEE Trans. Pattern Anal. Mach. Intell., vol. 48, no. 11, pp. 15571--15588, Nov. 2026. DOI: 10.1109/TPAMI.2026.3722239
\end{minipage}}

\maketitle

\begin{abstract}
Recent Artificial Intelligence (AI)-based video editing has enabled users to edit videos through simple text prompts, significantly simplifying the editing process.
However, recent zero-shot video editing techniques focus on global or single-object edits, which can lead to unintended changes in other parts of the video.
When editing multiple objects only within their localized regions, existing methods face challenges, such as unfaithful editing, editing leakage, and lack of suitable evaluation datasets and metrics.
To overcome these limitations, we propose \textbf{P}robability \textbf{R}edistribution for \textbf{I}nstance-aware \textbf{M}ulti-object Video \textbf{Edit}ing (\textbf{PRIMEdit}).
PRIMEdit is a zero-shot framework that introduces two key modules: (i) Instance-centric Probability Redistribution (IPR) to ensure precise localization and faithful editing and (ii) Disentangled Multi-instance Sampling (DMS) to prevent editing leakage.
Additionally, we present the MIVE dataset for multi-instance video editing, including its subsets categorized by instance counts and sizes for controlled analysis across different levels of scene complexity.
We also introduce the Cross-Instance Accuracy (CIA) Score that is \textit{firstly} proposed to evaluate the editing leakage in multi-instance video editing tasks, which exhibits strong correlation with human evaluation as measured by Spearman's $\rho = 0.86$.
Our extensive qualitative, quantitative, and user study evaluations demonstrate that PRIMEdit significantly outperforms recent state-of-the-art methods in terms of editing faithfulness, accuracy, and leakage prevention, setting a new benchmark for multi-instance video editing.
\end{abstract}

\begin{IEEEkeywords}
Video editing, multi-instance video editing, diffusion, attention manipulation
\end{IEEEkeywords}

\section{Introduction}
\label{sec:intro}

\IEEEPARstart{T}{he} popularity of short-form social media videos has grown significantly \cite{rise2014vandersmissen, unlocking2024kim}, yet editing them remains time-consuming \cite{automation2021soe} and often requires professional assistance \cite{short2023liu}.
These challenges have driven the development of AI-based video editing (VE) tools \cite{unlocking2024kim}, enabled by advances in generative \cite{ldm2022rombach, magvit2023yu, align2023blattmann} and visual language models \cite{clip2021radford}.
By allowing users to specify edits through simple text prompts \cite{text2live2022bar, fatezero2023qi}, these tools make editing faster and accessible.
\blfootnote{*Co-first authors (equal contribution)}
\blfootnote{\textdagger Co-corresponding authors}
\blfootnote{We provide sample input and edited videos for our method along with baseline methods in our project page: \url{https://kaist-viclab.github.io/primedit-site/}}

Recent VE methods leverage pre-trained text-to-image (T2I) models \cite{ldm2022rombach}.
Compared with training on large-scale datasets \cite{avid2024zhang, consistent2023cheng} or fine-tuning on single videos \cite{tuneavideo2023wu, videop2p2024liu, text2live2022bar}, zero-shot methods \cite{controlvideo2023zhang, flatten2023cong, rave2023kara, fresco2024yang, tokenflow2023geyer, fatezero2023qi, groundavideo2024jeong, stablevideo2023chai} continue to gain attraction for their efficiency and the availability of pre-trained T2I models.
Most zero-shot approaches perform global editing that modifies the entire scene \cite{controlvideo2023zhang, fresco2024yang} or single-object editing that may unintentionally affect other parts of the video \cite{pix2video2023ceylan, videop2p2024liu}.
However, users often require precise object-level editing without altering the rest of the video, e.g., replacing explicit content, like cigarette, to create family-friendly versions \cite{deepcens2023yuksel}.

\begin{figure}[t]
    \centering
    \includegraphics[width=0.95\linewidth]{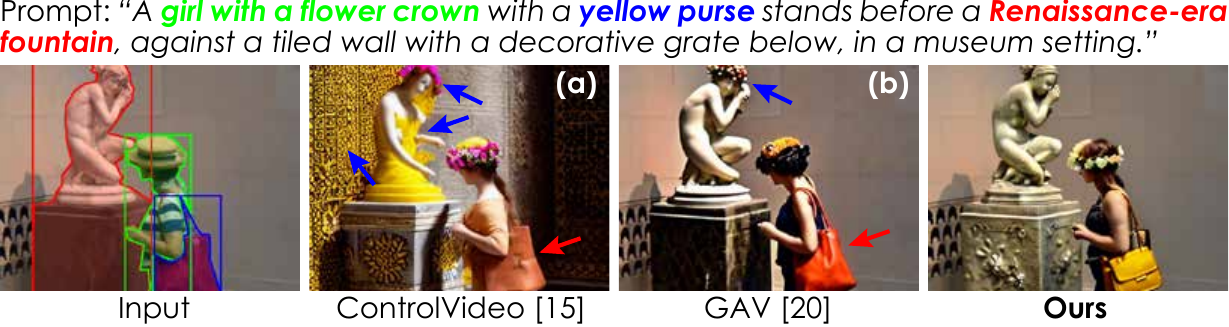}
    \vspace{-0.3cm}
    \caption{
        Limitations of previous SOTA methods. (a) ControlVideo \cite{controlvideo2023zhang} relies on single global captions, and (b) GAV \cite{groundavideo2024jeong} depends on bounding box conditions that can sometimes overlap. Both are susceptible to unfaithful editing (\textcolor{red}{red arrow}) and attention leakage (\textcolor{blue}{blue arrow}).
        Note that the colored words in \texttt{Prompt}  correspond to the instances to be edited in \texttt{Input}.
        }
    \label{fig:1_sota_weaknesses}
    \vspace{-0.55cm}
\end{figure}

Local VE addresses this problem by accurately manipulating specific objects in videos.
However, adopting pre-trained T2I models for this task is challenging, as they lack fine-grained control and require additional training \cite{avid2024zhang} or attention manipulation \cite{videograin2025yang, videop2p2024liu} for spatial control.
The problem is exacerbated when models perform multiple local edits simultaneously using a single long caption.
Fig. \ref{fig:1_sota_weaknesses} illustrates the limitations of previous SOTA VE methods.
As shown in Fig. \ref{fig:1_sota_weaknesses}-(a), ControlVideo \cite{controlvideo2023zhang}, due to the lack of explicit local control, suffers from (i) unfaithful editing (e.g., the red bag not transforming into a yellow purse) and (ii) attention leakage \cite{dynamic2023yang}, where edits intended for one object unintentionally affect others (e.g., both wall and statue becoming yellow).
Ground-A-Video (GAV) \cite{groundavideo2024jeong} in Fig. \ref{fig:1_sota_weaknesses}-(b) shows that simultaneous multi-object VE is feasible using a T2I model \cite{ldm2022rombach} fine-tuned with grounding conditions \cite{gligen2023li}.
However, GAV also suffers from attention leakage \cite{dynamic2023yang}, especially when object bounding boxes overlap (e.g., the statue also gaining a flower crown).

\IEEEpubidadjcol

Very recently, VideoGrain \cite{videograin2025yang} attempted to address attention leakage using object masks and attention modulation \cite{densediffusion2023kim}.
However, modulation in spatio-temporal attention (STA) introduces significant memory and computational overhead, especially as the number of video frames increases (see Table \ref{table:10_runtime_memory_comparison}).
Moreover, to describe edited scenes, VideoGrain relies on global edit captions, which may be unavailable or difficult to generate for scenes with a large number of objects.

Furthermore, both GAV \cite{groundavideo2024jeong} and VideoGrain \cite{videograin2025yang} are evaluated on limited datasets, insufficient for comprehensive testing of multi-instance VE across diverse viewpoints, instance sizes, and numbers of instances.
In addition, the global editing metrics used in these methods fail to accurately measure local editing quality, which is essential for multi-instance VE.

From the above, four critical challenges remain in multi-instance VE:
\textbf{(i) Attention leakage}.
The lack of local control in pre-trained T2I models, imprecise input conditions \cite{groundavideo2024jeong}, and reliance on a single global edit caption \cite{controlvideo2023zhang} prevent previous methods from effectively disentangling edit prompts;
\textbf{(ii) Unfaithful editing}.
The absence of techniques to enhance editing faithfulness in multi-instance VE often leads to inaccurate edits;
\textbf{(iii) Lack of fast and intuitive multi-instance VE methods}.
Although modulation in \cite{videograin2025yang} enables per-object editing, it slows the generative process, and reliance on full-scene descriptions makes editing cumbersome, especially in cluttered videos.
\textbf{(iv) Lack of evaluation datasets and metrics}.
Recent methods \cite{videograin2025yang, groundavideo2024jeong} are evaluated on limited datasets using metrics inadequate for assessing local VE quality.

\begin{figure*}[t]
  \centering 
  \includegraphics[width=0.9\linewidth]{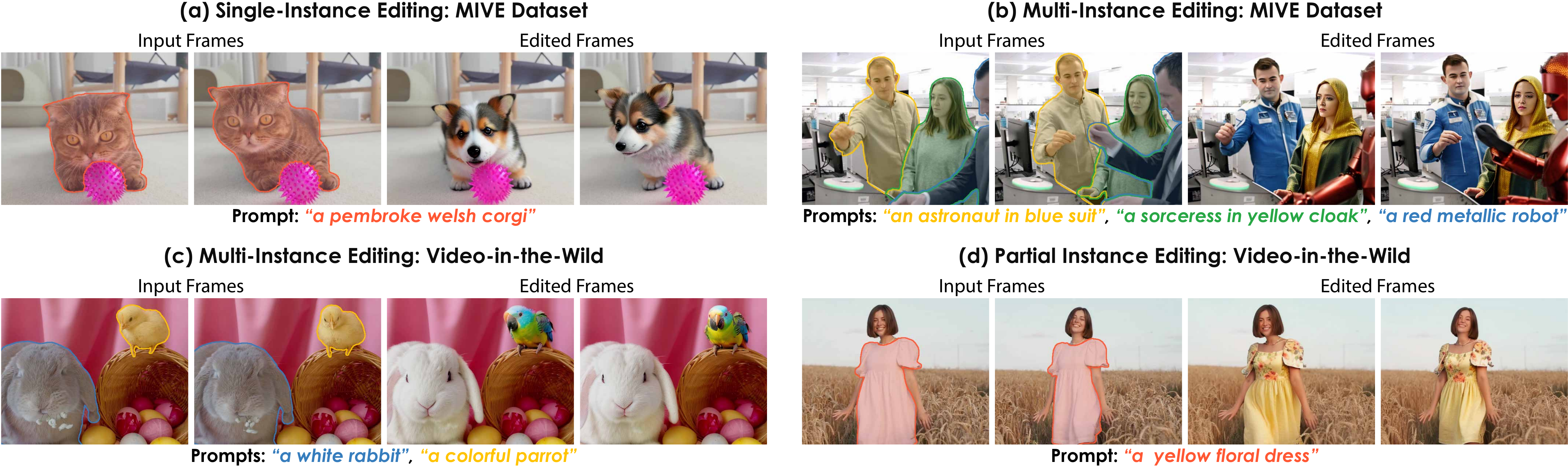}
  \vspace{-0.3cm}
  \caption{
      Given a video, instance masks, and target instance captions, our PRIMEdit enables faithful and disentangled edits in (a) single- and (b)-(c) multi-instance levels, as well as an applicability to more fine-grained (d) partial instance level without additional training.
      Unlike previous methods, our PRIMEdit does not rely on global edit captions, but leverages individual instance captions.
      Object masks are color-coded to match its corresponding edit caption.
  }
  \vspace{-0.45cm}
  \label{fig:2_main_figure}
\end{figure*}

To address these challenges, we present \textbf{P}robability \textbf{R}edistribution for \textbf{I}nstance-aware \textbf{M}ulti-object Video \textbf{Edit}ing (\textbf{PRIMEdit}), an intuitive zero-shot multi-instance VE framework that achieves faithful edits and reduces attention leakage by disentangling multi-instance edits.
PRIMEdit readily adopts existing T2I and T2V models within a unified framework and achieves multi-instance editing through two key modules:
(i) To enhance editing faithfulness and increase the likelihood of objects appearing within their masks, we introduce Instance-centric Probability Redistribution (IPR, Sec. \ref{subsec:ipr}) in the cross-attention layers; and
(ii) Inspired by prior works \cite{instancediffusion2024wang, noisecollage2024shirakawa}, we design Disentangled Multi-instance Sampling (DMS, Sec. \ref{subsec:dms}), which \textit{significantly} reduces attention leakage.
Fig. \ref{fig:2_main_figure} presents edited results across scenarios, including Single-Instance (a), Multi-Instance (b, c), and Partial Instance (d) editing.
We further evaluate PRIMEdit on our newly proposed 200-video MIVE dataset using standard VE metrics and a novel Cross-Instance Accuracy (CIA) Score to quantify attention leakage, comparing against previous SOTA VE methods.

Our contributions are as follows:
\begin{itemize}
    \item We propose a novel zero-shot multi-instance video editing framework, PRIMEdit, which enables fast and intuitive multi-instance editing for videos;
    \item We propose to disentangle multi-instance video editing through (i) Instance-centric Probability Redistribution (IPR), which enhances localization and faithfulness, and (ii) Disentangled Multi-instance Sampling (DMS), which reduces attention leakage;
    \item We introduce a new evaluation benchmark comprising a novel metric, the Cross-Instance Accuracy (CIA) Score, strongly correlated with human evaluation (Spearman’s $\rho = 0.86$), and a new 200-video dataset, MIVE, with varying numbers and sizes of instances, accompanied by instance-level captions and masks. The CIA Score is the \textit{first} metric to quantify attention leakage in multi-instance editing, and MIVE is the \textit{first} benchmark for \textit{multi-instance} video editing;
    \item Extensive experiments demonstrate the effectiveness of PRIMEdit in disentangling multiple edits across instances and achieving faithful editing with better temporal consistency, \textit{significantly} outperforming recent SOTA methods in local video editing metrics, CIA Score, and user study.
\end{itemize}

\section{Related Works}
\label{sec:related_works}

\subsection{Global Video Editing}
\noindent
Recent advances in diffusion models \cite{ddpm2020ho, ddim2020song, score2020song} have accelerated progress in T2I \cite{dalle2022ramesh, glide2021nichol, imagen2022saharia, diffusion2021dhariwal, ldm2022rombach} and T2V \cite{align2023blattmann, videoprobabilisticdm2023yu, lavie2023wang, imagenvideo2022ho, latent2022he, structure2023esser, modelscope2023wang} models for generative tasks.
These breakthroughs led to numerous video editing (VE) frameworks built on pre-trained backbones.
Existing methods either fine-tune models for editing (training-based) \cite{avid2024zhang, videocomposer2024wang, structure2023esser}, optimize on a single input video \cite{tuneavideo2023wu, videop2p2024liu}, or enable VE without additional training or optimization (zero-shot) \cite{controlvideo2023zhang, flatten2023cong, rave2023kara, fresco2024yang, tokenflow2023geyer, groundavideo2024jeong, fatezero2023qi, videograin2025yang}.
Our work belongs to the zero-shot category, allowing editing without additional training or optimization.

\noindent
\textbf{T2I-based Text-guided Video Editing.}
In T2I-based VE, prior methods \cite{fatezero2023qi, tuneavideo2023wu, tokenflow2023geyer, flatten2023cong, rave2023kara, groundavideo2024jeong, controlvideo2023zhang, fresco2024yang, slicedit2024cohen} rely on pre-trained T2I models \cite{ldm2022rombach, gligen2023li, controlnet2023zhang}.
Most methods focus on temporal consistency, as pre-trained T2I models struggle to maintain coherence across frames despite their strong editing capabilities.

To address temporal inconsistency, Tune-A-Video \cite{tuneavideo2023wu} proposes a one-shot tuning strategy and a spatio-temporal attention mechanism that extends spatial self-attention in latent diffusion models to capture inter-frame correlations.
FateZero \cite{fatezero2023qi} builds on this by enabling VE without tuning, storing intermediate attention maps during inversion and reusing them during editing to preserve structural and motion consistency.

Following these early works, many approaches improved temporal consistency by modifying spatio-temporal attention while preserving editing fidelity.
For example, FLATTEN \cite{flatten2023cong} and COVE \cite{cove2024wang} incorporate optical flow and diffusion feature correspondences, respectively, into the spatio-temporal attention computation, enabling patches with similar trajectories to attend across frames.
FRESCO \cite{fresco2024yang} further disentangles spatio-temporal attention into intra-frame and inter-frame components to better capture spatial and temporal dependencies.

Another line of work modifies input formats or processing strategies to improve temporal coherence.
RAVE \cite{rave2023kara} uses latent grids and noise shuffling, while TokenFlow \cite{tokenflow2023geyer} propagates diffusion features using inter-frame correspondences.
ControlVideo \cite{controlvideo2023zhang} uses ControlNet \cite{controlnet2023zhang} with spatio-temporal attention for structural alignment.
Slicedit \cite{slicedit2024cohen} processes spatial and spatiotemporal slices during sampling.
In contrast, our work adopts a T2V model to maintain temporal consistency while focusing on localized multi-instance editing fidelity.

\noindent
\textbf{T2V-based Text-guided Video Editing.}
Recently, T2V-based VE methods have emerged with the availability of public T2V models \cite{animatediff2023guo, align2023blattmann, preservecorrelationnoiseprior2024ge, cogvideox2025yang, wan2025wan, kong2024hunyuanvideo, gentron2024chen}.
Unlike T2I-based VE, T2V-based methods inherently preserve temporal consistency.
Recent works \cite{dreammotion2024jeong, dmt2023yatim, eve2024singer, mou2024revideo, xiao2024moft, burgert2025go, conmo2025gao, videop2p2024liu} focus instead on developing new editing techniques, including fine-tuning the T2V model \cite{eve2024singer}, latent optimization \cite{dreammotion2024jeong, gao2025unity}, or transferring motion from input videos during generation \cite{dmt2023yatim, burgert2025go}.
For example, DreamMotion \cite{dreammotion2024jeong} uses score distillation with spatial and temporal self-similarity matching to maintain structural correspondence and temporal smoothness, respectively.
DMT \cite{dmt2023yatim} optimizes latents using a space-time feature loss derived from spatial marginal means of input latents from a T2V model \cite{modelscope2023wang, zeroscope2023cerspense}, enabling motion transfer during inference.
In contrast, rather than latent optimization via distillation or feature losses, we modify the cross-attention and sampling process of a pre-trained T2V model \cite{animatediff2023guo} for precise localized editing.
Notably, our method also extends to T2I-based VE (see \textit{Appendix}).

\subsection{Local Video Editing}

\noindent
\textbf{Local Video Editing.}
Local VE aims to modify specific regions of input frames without modifying other parts.
Similar to image generation and editing, the demand for fine-grained control in videos has led to several local VE methods \cite{avid2024zhang, videop2p2024liu, groundavideo2024jeong, videograin2025yang, conmo2025gao, dreammotion2024jeong}.
These methods typically adapt spatial control techniques from image generation to VE.
In image generation, spatial control is achieved either by training additional adapters \cite{controlnet2023zhang, spatext2023avrahami, gligen2023li, reco2023yang, instancediffusion2024wang} or by zero-shot techniques such as attention manipulation \cite{densediffusion2023kim, p2p2022hertz}, optimization \cite{multidiffusion2023bar, attend2023chefer, training2024chen, boxdiff2023xie}, and multi-branch sampling \cite{noisecollage2024shirakawa}.

In particular, AVID \cite{avid2024zhang} adds motion modules to a T2I inpainting model (Stable Diffusion \cite{ldm2022rombach}) and trains a ControlNet-based structure guidance for tasks such as retexturing and uncropping.
In the zero-shot setting, Video-P2P \cite{videop2p2024liu} achieves single-object-level localized editing by adopting cross-attention control \cite{p2p2022hertz} which combines cross-attention maps from the reconstruction branch to preserve unedited parts and the editing branch to edit a specific region.
More recently, DreamMotion \cite{dreammotion2024jeong} employs a T2V model and performs editing via score distillation sampling.
Although capable of local control, these methods are designed for single-object editing and often fail when handling multiple objects simultaneously (see \textit{Appendix} for Video-P2P results).

\noindent
\textbf{Multi-Instance Local Video Editing.}
To overcome the limitations of local VE, which either handles single-object editing but is prone to attention leakage or performs sequential object-by-object editing but accumulates noise, multi-instance VE aims to edit multiple objects \textit{simultaneously}.
\textit{Multi-instance} VE remains relatively underexplored, with few works \cite{groundavideo2024jeong, videograin2025yang} adapting image generation techniques \cite{gligen2023li, densediffusion2023kim} for object localization.
GAV \cite{groundavideo2024jeong} leverages GLIGEN \cite{gligen2023li} and integrates gated self-attention layers into transformer blocks for spatial control.
However, GLIGEN requires retraining these layers for each new T2I model (e.g., SDv1.5 and SDv2.1), reducing flexibility.
VideoGrain \cite{videograin2025yang} uses masks and attention modulation \cite{densediffusion2023kim} for the spatio-temporal attention (STA) and cross-attention layers.
However, manipulating STA increases computational costs as the number of edited frames increases.
Our work introduces a novel redistribution technique that operates only in cross-attention layers, enabling seamless integration with any T2I or T2V model and spatial control with minimal overhead.
Table~\ref{table:1_sota_feature_comparison} presents a more detailed comparison between our work and two previous multi-instance VE SOTA methods.
Compared to GAV and VideoGrain, our PRIMEdit enables mask-based multi-instance VE using a simpler localization mechanism, IPR, applied only in the U-Net's cross-attention layers, and supports integration with T2I and T2V models.

\begin{table}[t]
    \small
    \caption{
        Comparison of our PRIMEdit against previous multi-instance video editing state-of-the-art methods.
    }
    \vspace{-0.5cm}
    \begin{center}
        \setlength\tabcolsep{3pt}{
            \renewcommand{\arraystretch}{1.1}
            \scalebox{0.7}{
              \begin{tabular}{l | c | c | c}
                \toprule
                
                \multirow{2}{*}{ \diagbox[]{Features}{Methods} } & \multirow{2}{*}{GAV \cite{groundavideo2024jeong}} & \multirow{2}{*}{VideoGrain \cite{videograin2025yang}} & \multirow{2}{*}{\textbf{PRIMEdit (Ours)}} \\
                & & & \\
                
                \midrule
                
                Multi-instance & \checkmark & \checkmark & \checkmark \\
                
                \midrule
                
                Conditions & bounding boxes & masks & masks \\
                
                \midrule
                
                \multirow{2}{*}{Local control} & \multirow{2}{*}{Grounding \cite{gligen2023li}} & Attention & Instance-centric Probability \\
                
                & & Modulation \cite{videograin2025yang, densediffusion2023kim} & Redistribution (Sec. \ref{subsec:ipr}) \\
                
                \midrule
                
                U-Net local & gated & spatio-temporal attention & cross-attention \\
                
                \hspace{2mm} control & self-attention & \& cross-attention & only \\
                
                \midrule
                
                Background & \multirow{2}{*}{$\times$} & \multirow{2}{*}{\checkmark} & \multirow{2}{*}{\checkmark} \\
                
                \hspace{2mm} preservation & & & \\
                
                \midrule
                
                T2I-compatible & \checkmark & \checkmark & \checkmark \\
                
                \midrule
                
                T2V-compatible & $\times$ & $\times$ & \checkmark \\
                
                \bottomrule
              \end{tabular}
            }
        }
    \end{center}
    \vspace{-0.6cm}
    \label{table:1_sota_feature_comparison}
\end{table}
\section{Proposed Method: PRIMEdit}
\label{sec:method}

\subsection{Preliminaries}
\label{subsec:preliminaries}

\noindent
Given a set of input frames $f^{1:N}$, each frame $f^{k}$ is encoded into a clean latent code $z_0^k= \mathcal{E}(f^k)$ using the encoder $\mathcal{E}$ of the Latent Diffusion Model (LDM) \cite{ldm2022rombach}.
DDIM inversion \cite{ddim2020song, diffusion2021dhariwal} is applied to map the clean latent $z_0^k$ to a noised latent $\hat{z}_T^k$ through reversed diffusion timesteps $t: 1 \rightarrow T$ using the U-Net $\epsilon_\theta$ of the LDM as:
\begin{equation}
\begin{split}
    \resizebox{.89\linewidth}{!}{$\hat{z}_{t}^k = \sqrt{\alpha_t}\dfrac{\hat{z}_{t-1}^{k} - \sqrt{1 - \alpha_{t-1}}\epsilon_{\theta}(I_{t-1})}{\sqrt{\alpha_{t-1}}} +
    \sqrt{1 - \alpha_{t}}\epsilon_\theta(I_{t-1})$},
\end{split}
\end{equation}
where $\alpha_t$ is a noise scheduling parameter \cite{ldm2022rombach} and $I_{t}=(\hat{z}_{t}, t, c, e)$ represents the noisy latent at timestep $t$ with text prompt $c$ as input.
After inversion, $\hat{z}_{t}^k$ at $t=T$ is used as input to the DDIM denoising process \cite{ddim2020song} to perform editing:
\begin{equation}
\begin{split}
   \resizebox{.89\hsize}{!}{$\hat{z}_{t-1}^k = \sqrt{\alpha_{t-1}}\dfrac{{\hat{z}_{t}^{k}} - \sqrt{1 - \alpha_{t}}\epsilon_{\theta}(I_{t})}{\sqrt{\alpha_{t}}} +
   \sqrt{1 - \alpha_{t-1}}\epsilon_\theta(I_{t})$}.
\end{split}
\end{equation}
ControlNet \cite{controlnet2023zhang} condition $e$ can be added as additional guidance for the sampling and can be obtained from any structured information (e.g., depth maps).
The edited frame $\hat{f}^{k}=\mathcal{D}(\hat{z}_{0}^k)$ is obtained using the decoder $\mathcal{D}$ of the LDM. 
A classifier-free guidance \cite{cfg2022ho} scale of $s_{cfg} = 1$ and a larger scale $s_{cfg} \gg 1$ are used during inversion and denoising, respectively.

\subsection{Overall Framework}
\label{subsec:overall_framework}

\begin{figure*}[t]
    \centering 
    \includegraphics[width=1.0\linewidth]{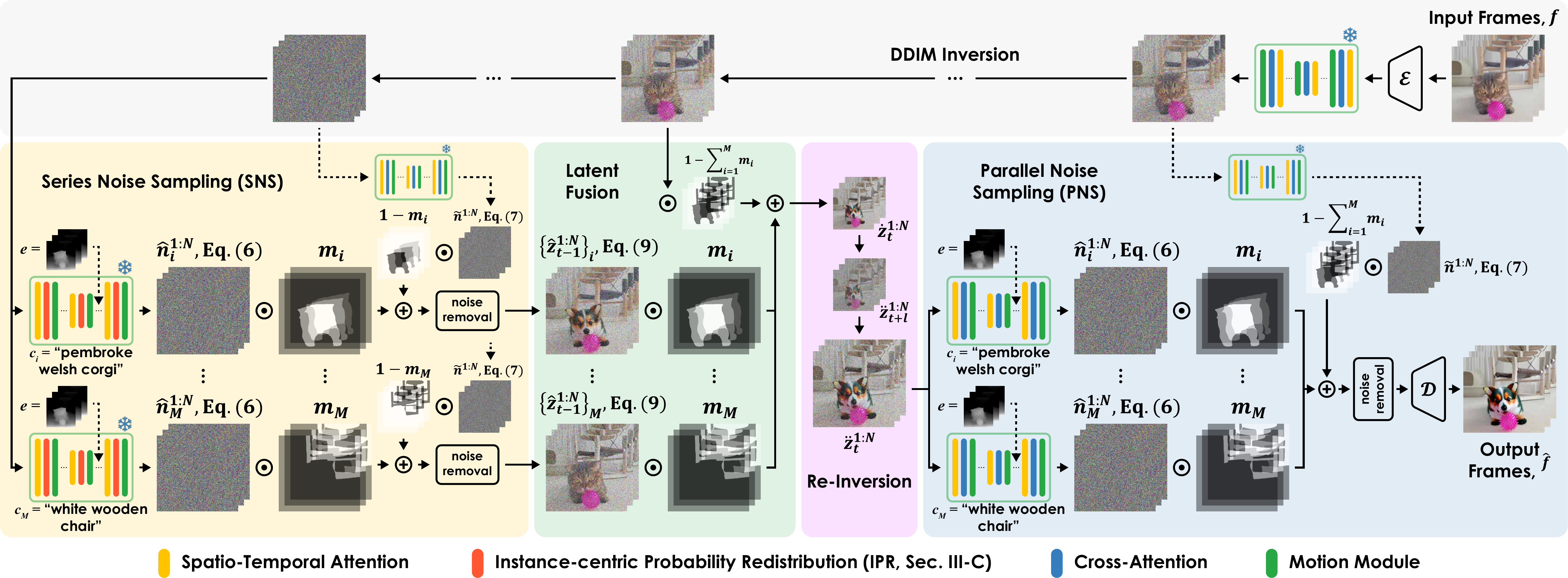}
    \vspace{-0.7cm}
    \caption{
        The overall framework of our PRIMEdit.
        For a given $M$ multi-instances, each instance has a caption $c_i$ and instance masks $\boldsymbol{m_i}$ for editing.
        Our Disentangled Multi-instance Sampling (DMS, Sec. \ref{subsec:dms}) consists of series noise sampling (SNS, yellow box), latent fusion (green box) to fuse different instance latents, re-inversion (pink box) to harmonize the latents after fusion, and parallel noise sampling (PNS, blue box).
        In addition, our Instance-centric Probability Redistribution (IPR, Sec. \ref{subsec:ipr}) provides better spatial control.
    }
    \vspace{-0.5cm}
    \label{fig:3_overall_framework}
\end{figure*}

\begin{table}[t]
    \caption{
        Key Notations and Acronyms in PRIMEdit.
    }
    \vspace{-0.2cm}
    \label{table:method_notations}
    \centering
    \setlength{\tabcolsep}{5pt}
    {
    \scalebox{0.85}{
        \begin{tabular}{l p{6cm}}
            \toprule
            Symbol / Acronym & Description \\
            \midrule
            
            PRIMEdit & Proposed multi-instance video editing framework. \\
            
            LDM & Latent Diffusion Model. \\
            
            $\mathcal{E}$ / $\mathcal{D}$ & VAE encoder / decoder of the LDM. \\
            
            $\epsilon_\theta$ & U-Net noise prediction network. \\
            
            $f^{1:N}$ & Input video frames ($N$ total frames). \\
    
            $M$ & Number of instances in the video. \\
            
            $z_t^{1:N}$ & Latent representations at timestep $t$. \\
            
            $\tilde{z}_t^{1:N}$ & Inverted latents obtained via DDIM inversion. \\
    
            $\hat{n}_{i}^{1:N}$ & Estimated instance noise. \\
    
            $\ddot{z}_{t}^{1:N}$ & Re-inverted noise. \\
            
            $\hat{z}_t^{1:N}$ & Edited latent representations during denoising. \\
            
            $c_i$ & Target edit caption for the $i$-th instance. \\
    
            $c_{\varnothing}$ & Empty caption. \\
            
            $e$ & ControlNet conditioning (e.g., depth maps). \\
            
            $\boldsymbol{m_i}$ & Instance masks across frames for the $i$-th instance. \\
    
            $\delta(\boldsymbol{m_i})$ & Dilated instance mask used for latent fusion. \\
            
            $T$ & Total number of denoising steps. \\
             
            IPR & Instance-centric Probability Redistribution. \\
            
            DMS & Disentangled Multi-instance Sampling. \\
            
            SNS & Series Noise Sampling. \\
            
            PNS & Parallel Noise Sampling. \\
            
            \bottomrule
        \end{tabular}
    }}
    \vspace{-0.5cm}
\end{table}

\noindent
In this work, we tackle multi-instance video editing (VE) by disentangling edits for \textit{multiple instances}.
Disentangling means that instances are edited separately in the early sampling stages and combined in later steps.
Given an input video $\boldsymbol{f} = f^{1:N}$ with $N$ frames, we define $M$ instances for multi-instance VE, where each instance is treated as an entity (or object) throughout the video $\boldsymbol{f}$.
The target edit for instance $i$ is defined as $g_i = \{\boldsymbol{m_i}, c_i\}$, where $\boldsymbol{m_i} = m_{i}^{1:N}$ denotes the set of $N$ instance masks over $\boldsymbol{f}$ (with $m_{i}^{k}$ the mask of the $i$-th instance in frame $f^{k}$), and $c_i$ is the corresponding edit caption for the $i$-th instance.
The complete set of target edits is $\boldsymbol{g} =\{g_i\}_{i=1}^{M}$.
For the $M$ instances in $\boldsymbol{f}$, we edit each instance $i$ according to $c_i$, keeping regions outside $m_{i}^{1:N}$ unchanged.

Fig. \ref{fig:3_overall_framework} illustrates the overall framework of PRIMEdit.
Starting from the initial latents $z_0^{1:N} = \mathcal{E}(f^{1:N})$ obtained via the VAE encoder $\mathcal{E}$ \cite{ldm2022rombach}, we employ the 3D U-Net \cite{ldm2022rombach, animatediff2023guo} to invert the latents using DDIM Inversion \cite{ddim2020song}.
This produces a sequence of inverted latents, $\{\boldsymbol{\tilde{z}_t}\}_{t=0}^T = \{\tilde{z}_t^{1:N}\}_{t=0}^T$, stored for later use, where $T$ denotes the number of denoising steps.

The vanilla cross-attention in the U-Net \cite{ldm2022rombach} cannot guarantee that each instance’s target edit remains within its masks $\boldsymbol{m_i}$ (see Fig. \ref{fig:9_ipr_ablation_study}-(b)).
To address this, we introduce Instance-centric Probability Redistribution (IPR, Sec. \ref{subsec:ipr}) in the cross-attention to improve edit placement within $\boldsymbol{m_i}$.

For the sampling, we introduce Disentangled Multi-instance Sampling (DMS, Sec. \ref{subsec:dms}), inspired by image generation methods \cite{instancediffusion2024wang, noisecollage2024shirakawa}, to disentangle multi-instance VE and reduce attention leakage.
Each instance is independently modified via Series Noise Sampling (SNS, yellow box in Fig. \ref{fig:3_overall_framework}), and the resulting denoised latents are harmonized through Parallel Noise Sampling (PNS), preceded by Latent Fusion and Re-inversion (blue, green, and pink boxes in Fig. \ref{fig:3_overall_framework}).
During sampling, we apply ControlNet \cite{controlnet2023zhang} with depth maps $\boldsymbol{d} = d^{1:N}$ extracted using MiDaS \cite{midas2020ranftl} as conditions.
Table~\ref{table:method_notations} provides a concise summary of all key notations and acronyms used in our PRIMEdit along with their definitions.

\subsection{Instance-centric Probability Redistribution (IPR)}
\label{subsec:ipr}

\begin{figure}[t]
   \centering
   \scalebox{0.9}{
      \includegraphics[width=0.92\linewidth]{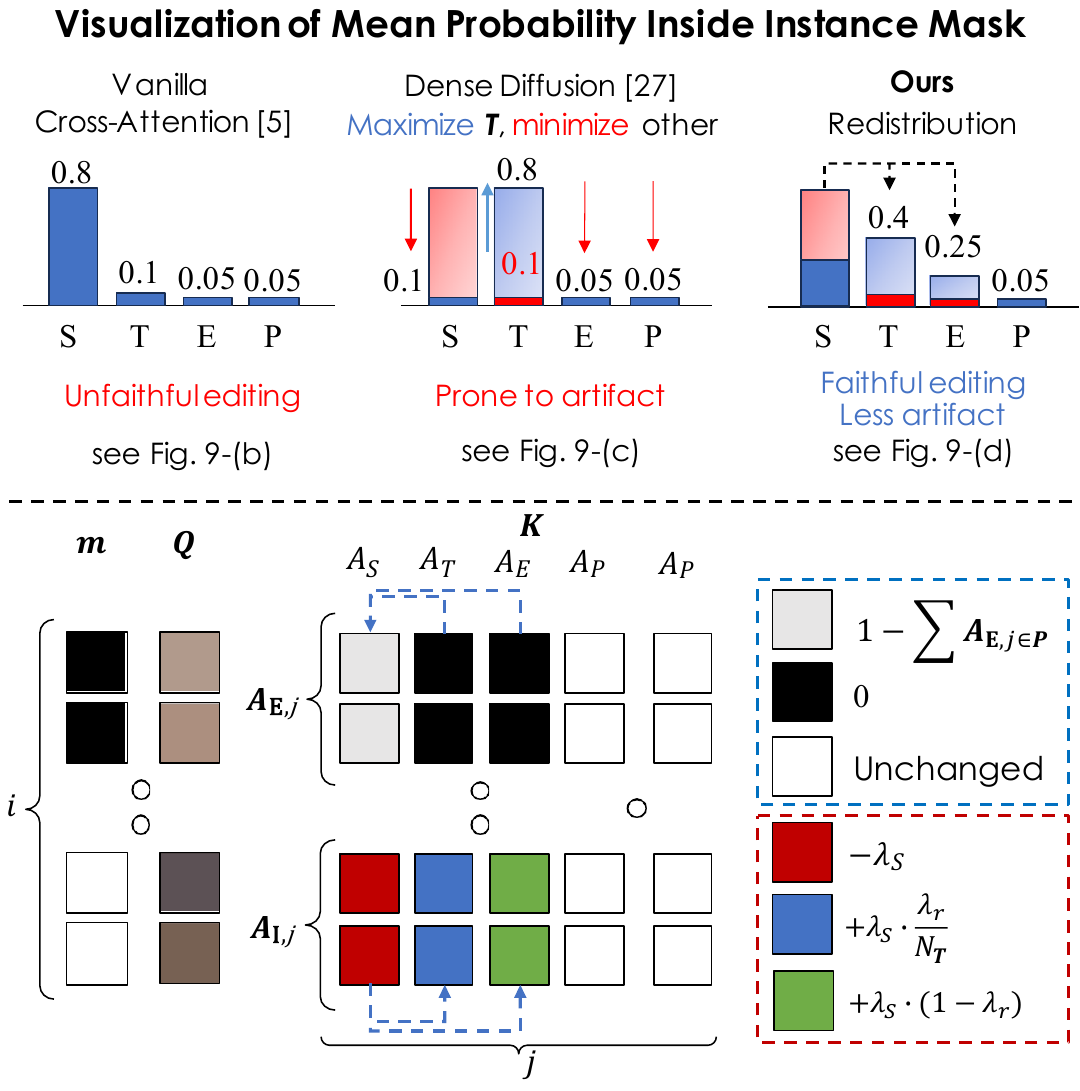}
   }
   \vspace{-0.3cm}
   \caption{
        A comparative illustration of our IPR versus Vanilla Cross-Attention \cite{ldm2022rombach} and DenseDiffusion \cite{densediffusion2023kim} (top). The attention manipulation details of our IPR is shown at the bottom.
    }
   \label{fig:4_ipr}
   \vspace{-0.5cm}
\end{figure}

\noindent
Our sampling requires edits to stay within instance masks, but vanilla cross-attention \cite{ldm2022rombach} struggles to localize instance-level modifications (see Fig. \ref{fig:9_ipr_ablation_study}-(b)).
To address this, we propose Instance-centric Probability Redistribution (IPR), inspired by attention modulation \cite{densediffusion2023kim}.
IPR ensures faithful edits while operating only in cross-attention layers, avoiding the computational overhead of modulating \cite{densediffusion2023kim} spatio-temporal layers \cite{videograin2025yang} for high-quality results (see Fig. \ref{fig:9_ipr_ablation_study}-(c)).

Fig. \ref{fig:4_ipr} illustrates our IPR (bottom) and a comparison with Vanilla Cross-Attention \cite{ldm2022rombach} and DenseDiffusion \cite{densediffusion2023kim} (top).
In the bottom part of Fig. \ref{fig:4_ipr}, we consider single-instance editing with target caption $c = c_i$.
The caption $c$, containing $n$ tokens, is encoded into embeddings $\mathbf{e} = \{e_j\}_{j=1}^n$ using CLIP \cite{clip2021radford}, where $e_j \in \mathbb{R}^{1 \times d}$ denotes the $j$-th token embedding.
These embeddings serve as keys $\boldsymbol{K} = \{K_j\}_{j=1}^n$ in cross-attention, with $K_j = e_j = \{S, \boldsymbol{T}, E, \boldsymbol{P}\}$ representing the start token $S$, text tokens $\boldsymbol{T}$, end token $E$, and padding tokens $\boldsymbol{P}$.
The cross-attention map $\boldsymbol{A} \in \mathbb{R}^{hw \times n}$ between image queries $\boldsymbol{Q} \in \mathbb{R}^{hw \times d}$ and keys $\boldsymbol{K} \in \mathbb{R}^{n \times d}$ is computed as:
\begin{equation}
    \boldsymbol{A} = \text{Softmax}(\boldsymbol{Q}\boldsymbol{K}^{T} \mathbin{/}\sqrt{d}) \in [0, 1]^{hw \times n},
\end{equation}
where $A_{i,j}$ of $\boldsymbol{A}$ denotes the correlation between the $i$-th image feature and the $j$-th token embedding $e_j$.
We propose to manipulate each $A_{i,j}$ through redistribution to localize edits and enhance faithfulness.

In IPR, we split $A_{i,j}$ for each instance into two sets based on mask $m$:
Exterior of the instance $\boldsymbol{A}_{\boldsymbol{\text{E}},j}=\{A_{i,j}; m_i = 0 \text{ for \textit{i}-th image feature}\}$ and Interior of the instance $\boldsymbol{A}_{\boldsymbol{\text{I}},j}=\{A_{i,j}; m_i = 1 \text{ for \textit{i}-th image feature}\}$ for the $j\text{-th}$ text embedding $e_j$.
IPR is motivated by two observations:
(i) Modifying the padding token attention $A_{i,j \in \boldsymbol{P}}$ causes artifacts, so we keep them unchanged;
(ii) Increasing the probability of the start-of-sequence token $S$ reduces editing faithfulness, while decreasing $S$ and reallocating it to $\boldsymbol{T}$ and $E$ improves faithfulness.
Hence, for $\boldsymbol{A}_{\boldsymbol{\text{E}},j}$, we set $\boldsymbol{T}$ and $E$ to zero and reallocate their values to $S$ to prevent edits outside the mask (blue dashed box in Fig. \ref{fig:4_ipr}).
In addition, for $\boldsymbol{A}_{\boldsymbol{\text{I}},j}$, we reduce $S$ and redistribute it to $\boldsymbol{T}$ and $E$ (red dashed box in Fig. \ref{fig:4_ipr}), enhancing faithfulness.

More specifically, we reduce $\boldsymbol{A}_{\boldsymbol{\text{I}},j=S}$ by $\lambda_S$ which linearly decays to 0 from $t=T$ to $t=1$, as follows:
\begin{equation}
\begin{gathered}
\label{eq:ipr_lambda_s}
    \resizebox{.89\linewidth}{!}{$\lambda_S = (t \mathbin{/} T) \cdot (\text{min}(\text{mean}(A_{\boldsymbol{\text{I}}, j=S}), \text{min}(A_{\boldsymbol{\text{I}}, j=S})) + W)$}
\end{gathered}
\end{equation}
where $W$, which improves editing fidelity for small objects, is a warm-up value that is linearly decayed from $\lambda$ to 0 in the early sampling steps $t < 0.1T$.
Increasing $\lambda$ enhances editing faithfulness but may introduce artifacts.
Empirically, we find $\lambda = 0.5$ provides the best trade-off between faithfulness and artifacts.
We then redistribute $\lambda_S$ to $\boldsymbol{T}$ and $E$ with ratios $\lambda_r$ and $1 - \lambda_r$, respectively.
The attention probabilities, $\boldsymbol{A}_{\boldsymbol{\text{I}}, j \in \boldsymbol{T}}$ and $\boldsymbol{A}_{\boldsymbol{\text{I}}, j=E}$, of $\boldsymbol{\text{T}}$ and $E$,  are updated as follows:
\begin{equation}
    \begin{gathered}
        \boldsymbol{A}_{\boldsymbol{\text{I}}, j \in \boldsymbol{T}} = \boldsymbol{A}_{\boldsymbol{\text{I}}, j \in \boldsymbol{T}} + \lambda_S \cdot \lambda_r \mathbin{/} N_{\boldsymbol{T}} \\
        \boldsymbol{A}_{\boldsymbol{\text{I}}, j=E} = \boldsymbol{A}_{\boldsymbol{\text{I}}, j=E} + \lambda_S \cdot (1 - \lambda_r) 
    \end{gathered}
\end{equation}
where $N_{\boldsymbol{T}}$ denotes the number of text tokens.
Increasing $\lambda_r$ enhances editing details for specific tokens, while decreasing it improves overall fidelity.
We empirically set $\lambda_r = 0.5$ for all experiments.
We compare the attention map visualizations for the vanilla cross-attention \cite{ldm2022rombach}, DenseDiffusion \cite{densediffusion2023kim} and our IPR in Sec. \ref{subsec:ipr_analysis} (Fig. \ref{fig:14_ipr_analysis}).

\subsection{Disentangled Multi-instance Sampling (DMS)}
\label{subsec:dms}

\noindent
To reduce attention leakage, we propose Disentangled Multi-instance Sampling (DMS), which edits each instance independently and fuses the resulting denoised latents in the final sampling steps, reducing the undesirable cross-instance interference.
As shown in Fig. \ref{fig:3_overall_framework}, DMS consists of two strategies: (i) Series Noise Sampling (SNS, yellow box) and (ii) Parallel Noise Sampling (PNS, blue box), preceded by latent fusion (green box) and re-inversion (pink box).
In SNS, we independently edit each instance $i$ using its caption $c_i$ and mask $\boldsymbol{m_i}$ by first estimating the instance noise $\hat{n}_{i}^{1:N}$ for $\boldsymbol{f} = f^{1:N}$.
$\hat{n}_{i}^{1:N}$ is obtained by:
\begin{equation}
    \hat{n}_{i}^{1:N} = \epsilon_{\theta}(\{z_{t}^{1:N}\}_i, c_i, \boldsymbol{m_i}, e, t),
\end{equation}
where $\{z_{t}^{1:N}\}_i$ is the noisy latent of the $i$-th instance at timestep $t$, and $e$ represents the depth condition obtained via MiDas \cite{midas2020ranftl}.
Then, we compute the background noise $\tilde{n}^{1:N}$ using the inverted latents as:
\begin{equation}
    \tilde{n}^{1:N} = \epsilon_{\theta}(\tilde{z}_t^{1:N}, c_{\varnothing}, t),
\end{equation}
where $c_{\varnothing}$ is an empty caption to reconstruct the background that is not intended to be edited.
We then fuse the estimated instance noise $\hat{n}_{i}^{1:N}$ with the inverted latent noise $\tilde{n}^{1:N}$ to obtain edited instance noise $\{\bar{n}_{t-1}^{1:N}\}_{i}$ using dilated masks $\delta(\boldsymbol{m_i})$.
$\{\bar{n}_{t-1}^{1:N}\}_{i}$ is computed as:
\begin{equation}
   \{\bar{n}_{t-1}^{1:N}\}_i = \hat{n}_{i}^{1:N} \odot \delta(\boldsymbol{m_i}) + \tilde{n}^{1:N} \odot (1 - \delta(\boldsymbol{m_i})),
\end{equation}
where $\odot$ denotes element-wise multiplication, and $\delta(\boldsymbol{m_i})$ is obtained by applying a $5 \times 5$ dilation kernel $\delta$ to $\boldsymbol{m_i}$, enlarging the instance regions and producing more coherent boundaries.
We apply $\delta$ because when $\boldsymbol{m_i}$ is smaller than the target edit for the $i$-th instance, some pixels from the boundaries of the original instance remain unedited, as they are treated part of the background.
We then perform one DDIM \cite{ddim2020song} denoising step from $t \rightarrow t-1$ to obtain instance latents $\{\hat{z}_{t-1}^{1:N}\}_{i}$ as:
\begin{equation}
    \label{eq:nps}
    \{\hat{z}_{t-1}^{1:N}\}_i = \text{DDIM}(\{\hat{z}_{t}^{1:N}\}_i, \{\bar{n}_{t-1}^{1:N}\}_i, t).
\end{equation}
SNS requires the edited instance $i$ to appear within $\boldsymbol{m_i}$, as the background is replaced with the inverted latent noise at each step to ensure it remains unchanged. 
Our proposed IPR (Sec. \ref{subsec:ipr}) provides this necessary spatial control.

While we can perform multi-instance VE by solely using SNS until the end of denoising, the edited frames suffer from temporal inconsistency (see Table \ref{table:4_ablations}).
Thus, we propose a two-step approach to mitigate this as follows:

\noindent
\textbf{(i) Step 1 - Latent Fusion}: As shown in the green box of Fig. \ref{fig:3_overall_framework}, we fuse the multiple instance latents $\{\dot{z}_{t}^{1:N}\}$ sampled independently from SNS via:
\begin{equation}
   \dot{z}_{t}^{1:N} = \Sigma_{i=1}^M(\{\hat{z}_{t}^{1:N}\}_i \odot \delta(\boldsymbol{m_i})  + \tilde{z}_{t}^{1:N} \odot \boldsymbol{m_B},
\end{equation}
where $\boldsymbol{m_B} = \boldsymbol{1} - \Sigma_{i=1}^M \delta(\boldsymbol{m_i})$ denotes background masks for $\boldsymbol{f} = f^{1:N}$;

\noindent
\textbf{(ii) Step 2 - Re-inversion}: We propose re-inversion as shown in the pink box of Fig. \ref{fig:3_overall_framework} by performing DDIM inversion for $l$ steps after Step 1 to obtain $\ddot{z}_{t+l}^{1:N}$ followed by PNS (discussed below) from timestep $t+l \rightarrow t$ to yield the final fused latent $\ddot{z}_{t}^{1:N}$ as follows:
\begin{equation}
    \label{eq:nps}
    \ddot{z}_{t}^{1:N} = \text{DDIM}(\ddot{z}_{t+l}^{1:N}, t).
\end{equation}

After mitigating the temporal inconsistency, we perform PNS for the remaining denoising steps.
PNS harmonizes the edited instance latents from SNS while maintaining decoupled sampling with their respective target captions.
In this strategy, we use the re-inverted fused latents $\ddot{z}_{t}^{1:N}$ to estimate each instance noise $\hat{n}_{i}^{1:N} = \epsilon_{\theta}(\ddot{z}_{t}^{1:N}, c_i, \boldsymbol{m_i}, e, t)$ using instance caption $c_i$.
We estimate the background noise as $\tilde{n}^{1:N} = \epsilon_{\theta}(\tilde{z}_t^{1:N}, c_{\varnothing}, t)$ using the inverted latents $\tilde{z}_{t}^{1:N}$ and empty caption $c_{\varnothing}$.
We combine $\hat{n}_{i}^{1:N}$ and $\tilde{n}^{1:N}$ to a single noise via masking, and perform one DDIM step to obtain $\hat{z}_{t-1}^{1:N}$ as:
\begin{equation}
    \label{eq:nps}
    \hat{z}_{t-1}^{1:N} = \text{DDIM}(\hat{z}_{t}^{1:N}, \Sigma_{i=1}^M(\hat{n}_i^{1:N} \odot \delta(\boldsymbol{m_i})) + \tilde{n}^{1:N} \odot \boldsymbol{m_B}, t).
\end{equation}
Lastly, we decode $\hat{z}_{t-1}^{1:N}$ using VAE decoder $\mathcal{D}$ to obtain the final edited frames $\hat{\boldsymbol{f}} = \mathcal{D}(\hat{z}_{0}^{1:N})$ as shown in Fig. \ref{fig:3_overall_framework}.
\section{Evaluation Data and Metric}
\label{sec:eval_data_metric}

\begin{figure*}[t]
  \centering
    \includegraphics[width=0.98\linewidth]{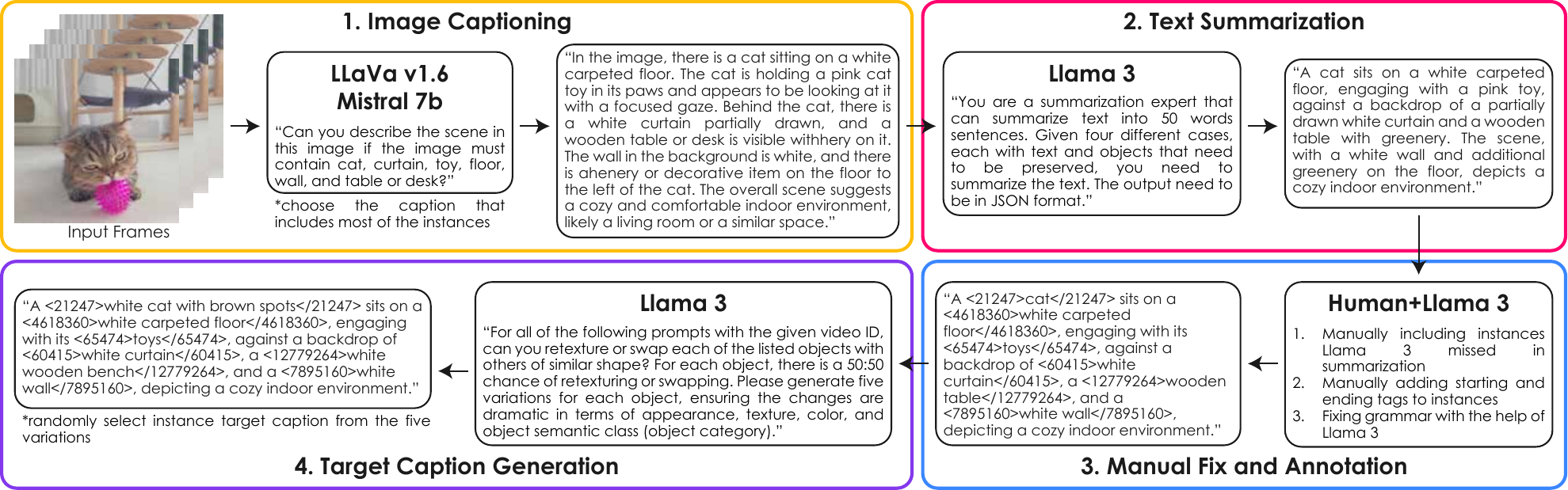}
    \vspace{-0.25cm}
    \caption{
        An illustration of the four-step video caption generation pipeline of our MIVE dataset.
        (1) Image Captioning - The caption generation process starts by prompting LLaVA \cite{llava1-5_2024liu} to generate caption for each video that includes all instances in the video.
        Since LLaVa can only accept images, we perform the prompting for each frame and select one representative caption that includes the most instances.
        (2) Text Summarization - We utilize Llama 3 \cite{llama2024dubey} to summarize the initial caption generated by LLaVa.
        (3) Manual Fix and Annotation - We manually include all of the instances of interest that are not included in the caption and manually add tags to map the instance captions to corresponding segmentation masks.
        (4) Target Caption Generation - We utilize Llama 3 to generate target captions by retexturing or swapping instances similar to \cite{avid2024zhang} for each instance.
    }
   \label{fig:5_dataset}
   \vspace{-0.2cm}
\end{figure*}

\begin{table*}[t!]
    \small
    \caption{
        Comparison of our MIVE dataset with existing text-guided video editing datasets (top), and the subset categorization of MIVE dataset according to difficulty levels (bottom). $\text{MIVE}_E$, $\text{MIVE}_M$, $\text{MIVE}_H$ and $\text{MIVE}_X$ indicate \texttt{Easy Video}, \texttt{Medium Video}, \texttt{Hard Video} and \texttt{eXtreme Video} subsets, respectively.
    }
    \vspace{-0.4cm}
    \begin{center}
        \setlength\tabcolsep{6pt}{
            \renewcommand{\arraystretch}{1.1}
            \scalebox{0.75}{
              \begin{tabular}{l | c c c c c c c c c c c}
                \toprule
                \multirow{2}{*}{Dataset Name} & Number & Number of & Number of & Number of Object & Number of Instances & Instance & Instance & Range of Average Instance \\        
                & of Clips & Frames per Clip & Objects per Clip & Classes & per Object Class & Captions & Masks & Mask Size Per Video (\%) \\
                \midrule
                TGVE \cite{loveutgve2023wu} \& TGVE+ \cite{eve2024singer} & 76 & $32-128$ & 1-2 & No Info & 1-2 & $\checkmark$ & $\times$ & No Masks \\
                VideoGrain \cite{videograin2025yang} & 13 & $16-32$ & 1-5 & 7 & 1-2 & \checkmark & \checkmark & $0.01 \sim 92.34$ \\
                MIVE Dataset (\textbf{Ours}) & 200 & $12-46$ & 1-9 & 54 & 1-17 & \checkmark & \checkmark & $0.02 \sim 77.35$ \\
                \midrule
                $\text{MIVE}_E$ & 200 & 12-46 & 1 & 37 & 1 & $\checkmark$ & $\checkmark$ & $0.32 \sim 77.35$  \\
                $\text{MIVE}_M$ & 116 & 12-46 & 1-3 & 45 & 1-2 & $\checkmark$ & $\checkmark$ &$0.02 \sim 77.35$  \\
                $\text{MIVE}_H$ & 66 & 13-46 & 2-6 & 47 & 1-5 & $\checkmark$ & $\checkmark$ & $0.02 \sim 76.72$ \\
                $\text{MIVE}_X$ & 18 & 15-26 & 2-9 & 29 & 1-17 & $\checkmark$ & $\checkmark$ & $0.05 \sim 40.39$ \\
                \bottomrule
              \end{tabular}
            }
        }
    \end{center}
    \vspace{-0.6cm}
    \label{table:2_dataset_statistics}
\end{table*}

\subsection{MIVE Dataset Construction}
\label{subsec:dataset_construction}

\noindent
Existing datasets \cite{loveutgve2023wu, eve2024singer, videograin2025yang, groundavideo2024jeong, davis2017pont} are unsuitable for multi-instance video editing (VE) tasks.
TGVE \cite{loveutgve2023wu} and TGVE+ \cite{eve2024singer} contain few videos, limited objects with few instances per object class, and no instance masks.
VideoGrain's \cite{videograin2025yang} dataset similarly includes few objects and limited instances per class.
GAV’s \cite{groundavideo2024jeong} dataset is only partially available, and DAVIS \cite{davis2017pont} does not distinguish multiple instances within the same class, making it unsuitable for multi-instance VE.

To address this gap, we introduce MIVE, the \textit{first} evaluation dataset for multi-instance VE tasks.
MIVE is built from VIPSeg \cite{vipseg2022miao}, a video panoptic segmentation dataset, with each video center-cropped to $512 \times 512$.
We retain clips that contain at least 12 frames and where all instances remain visible across frames.
To ensure diversity, we remove videos containing only one of the 40 most frequently occurring object classes (e.g., only persons).
This results in a final set of 200 videos.

As VIPSeg lacks source captions, we generate them using a visual language model, LLaVA \cite{llava1-5_2024liu}, and an LLM, Llama 3 \cite{llama2024dubey}.
We show our four-step caption generation pipeline in Fig. \ref{fig:5_dataset}:
(1) Image Captioning,
(2) Text Summarization,
(3) Manual Fix and Annotation, and
(4) Target Caption Generation.
As shown in Fig. \ref{fig:5_dataset}, we first prompt LLaVA to describe each frame while including known objects, then select the caption covering the most instances as the representative video caption.
Next, we prompt Llama 3 \cite{llama2024dubey} to summarize the caption with fewer words and provide the output in JSON format (omitted in Fig. \ref{fig:5_dataset} for clarity).
As some objects may be missed, we manually refine the captions and insert start/end tags for each instance to serve as reference for their segmentation masks.

In addition to the source captions, we generate a target caption for each instance using Llama 3 to prompt edits such as retexturing or swapping, similar to \cite{avid2024zhang}.
We instruct Llama 3 to generate five candidate target captions for each instance, and randomly select one for the final target instance caption.
We then modify the \textit{global source caption} by replacing its \textit{instance captions} with the selected targets to form the global target caption.
Although we focus on ``\texttt{thing}'' instances, we also generate captions for the ``\texttt{stuff}'' and background elements to enable future extensions with LLM-based target edits.
Sample frames and captions are provided in the \textit{Appendix}.

We further categorize MIVE dataset by (i) the numbers of instances to edit and (ii) the sizes of instances.
For the number of instances per video, we define four difficulty levels:
\texttt{Easy} ($\text{MIVE}_E$), one edited instance (largest);
\texttt{Medium} ($\text{MIVE}_M$), 1–3 instances;
\texttt{Hard} ($\text{MIVE}_H$), 4–7 instances; and
\texttt{eXtreme} ($\text{MIVE}_X$), more than 7 instances.
For instance size, we follow the COCO Dataset convention \cite{coco2014lin}: small ($<32^2$ pixels), medium ($32^2$–$96^2$), and large ($>96^2$).
MIVE dataset contains 69 small, 297 medium, and 434 large instances, with each instance area averaged across frames.
Under both criteria, PRIMEdit consistently outperforms baselines on key multi-instance VE metrics (see Tables \ref{table:5_quantitative_comparison_instance_size} and \ref{table:6_quantitative_comparison_instance_number}).

Table~\ref{table:2_dataset_statistics} compares MIVE with existing video editing datasets (top) and summarizes MIVE by the number of edited instances per video (bottom).
Compared to prior datasets, MIVE provides the largest number of evaluation videos and greater diversity in objects per video and instances per object class.
MIVE also covers a wide range of instance sizes, from small objects of 25 pixels ($\sim$0.01\% of a frame) to large objects occupying up to 98.68\% of the video.

\subsection{Cross-Instance Accuracy (CIA) Score}
\label{subsec:cia}

\begin{figure*}[t]
  \centering
    \includegraphics[width=0.85\linewidth]{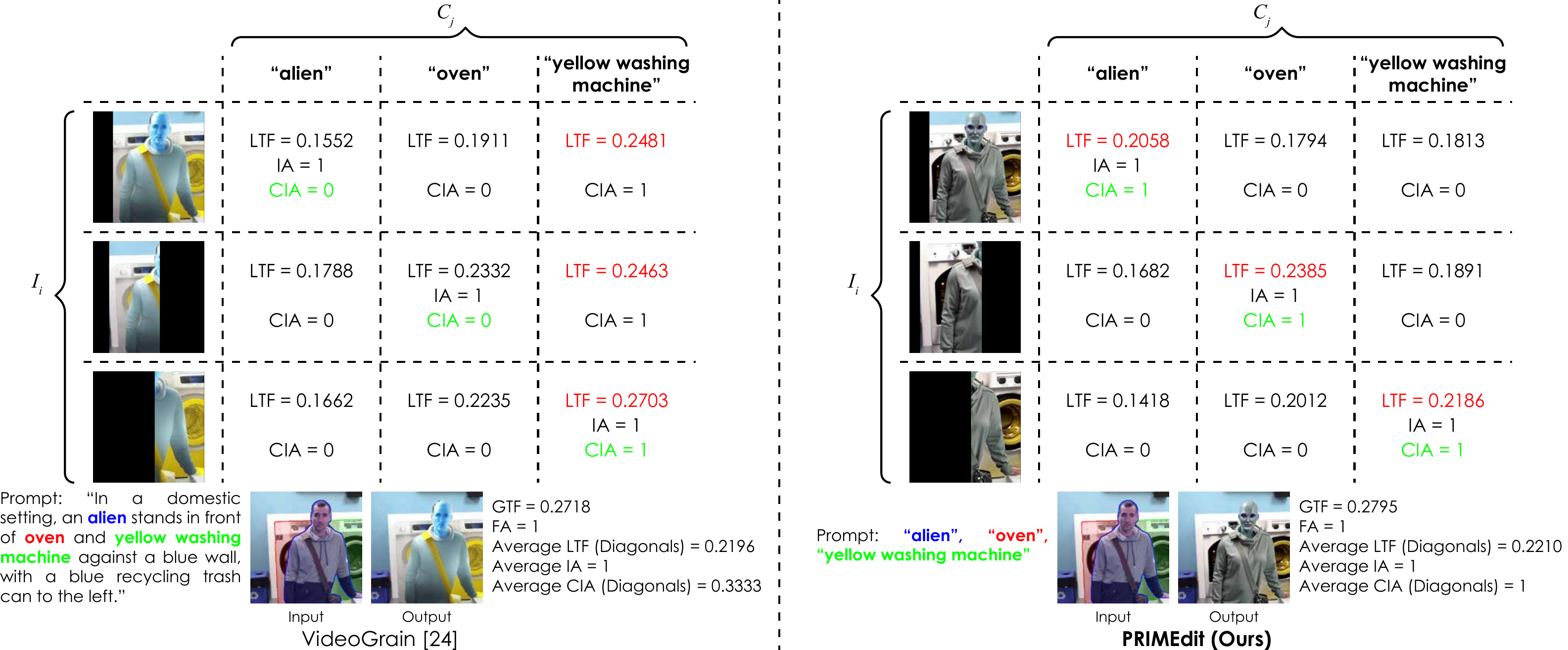}
    \vspace{-0.25cm}
    \caption{
        An illustration of Cross-Instance Accuracy (CIA) Score computation. We calculate the Local Textual Faithfulness (LTF) between each cropped instance and all the instance captions. For each row, we assign 1 to the maximum LTF (shown in \textcolor{red}{red}) and 0 to the rest. The CIA Score is calculated as the mean of the diagonal elements (shown in \textcolor{green}{green}).
    }
   \label{fig:6_cia}
   \vspace{-0.5cm}
\end{figure*}

\noindent
We propose a new metric, the Cross-Instance Accuracy (CIA) Score, to address limitations of existing VE metrics, particularly their inability to capture editing leakage, often referred to as attention leakage.
Global Textual Faithfulness (GTF) \cite{videograin2025yang, rave2023kara, flatten2023cong, tokenflow2023geyer} and Frame Accuracy (FA) \cite{fresco2024yang} fail to detect cases where a prompt for one instance affects others, as they cannot capture the nuances of individual instances within the global captions and frames.
Instance Accuracy (IA), defined as the percentage of instances more similar to their target instance caption than source instance caption, only measures target caption alignment and ignores interference from other instance captions.
Local Textual Faithfulness (LTF) \cite{instancediffusion2024wang, spatext2023avrahami} evaluates alignment between a cropped instance and its target instance caption but also overlooks leakage from other captions.
Background Preservation (BP) measures background changes but cannot detect leakage affecting instances.
We further observe cases where the LTF between a cropped instance and another instance’s target caption exceeds the score with its own instance caption, revealing cross-instance leakage.


To address all the above issues on GTF, LTF, FA, IA and BP, our CIA  Score is defined as follows:
For each cropped instance $i$ and each target instance caption $j$, we obtain the corresponding CLIP image embedding $I_i$ and text embedding $C_j$, respectively.
We compute the cosine similarity $S(I_i, C_j)$ between all pairs of image embeddings $I_i$ and the text embeddings $C_j$ for all $m$ instances $(i, j \in \{1,...,m\})$, producing an $m \times m$ matrix given by $\mathbf{S} = [ S(I_i, C_j) ]$ for $i,j=1, 2,..., m$.
Fig.~\ref{fig:6_cia} visualizes matrix $\mathbf{S}$ and illustrates how CIA is computed for both VideoGrain~\cite{videograin2025yang} and our PRIMEdit.

For each row of $\mathbf{S}$, we assign a value of 1 to the highest similarity score and 0 to the rest, ideally yielding a matrix with ones along the diagonal and zeros elsewhere.
We adopt this binary formulation because CIA evaluates exclusive instance–caption alignment rather than similarity magnitude.
Although LTF provides continuous similarity scores, even the sum or average of diagonal LTF values may fail to reveal severe cross-instance leakage.
As shown in Fig.~\ref{fig:6_cia}, VideoGrain achieves an average diagonal LTF of 0.2196, comparable to PRIMEdit’s 0.2210, yet in several rows, the off-diagonal similarities for VideoGrain exceed the diagonal entries, indicating incorrect caption dominance.
Consequently, continuous diagonal aggregation masks such violations.
By enforcing a winner-takes-all decision per row, CIA directly penalizes these cases, which is reflected in the average CIA scores (0.3333 for VideoGrain versus 1.0 for PRIMEdit), clearly distinguishing the methods despite similar diagonal LTF averages.

The CIA Score is calculated as the mean of the diagonal elements as:
\begin{equation}
    \text{CIA} = (1 \mathbin{/} m) \cdot \Sigma_{i=1}^{m} S(I_i, C_i).
\end{equation}

As illustrated in Fig.~\ref{fig:6_cia}, CIA clearly captures editing leakage in VideoGrain. For example, the second cropped instance is more influenced by the caption ``yellow washing machine,'' resulting in a CIA value of 0.

\section{Experiments}
\label{sec:experiments}

\begin{table*}[t]
    \caption{
        Evaluation metrics. Acronyms, full names, and concise descriptions of the metrics used in our experiments.
    }
    \vspace{-0.2cm}
    \label{table:metric_summary}
    \centering
    \setlength{\tabcolsep}{4pt}
    \scalebox{0.85}{
        \begin{tabular}{l l p{12cm}}
        \toprule
        Metric & Full Name & What It Measures \\
        \midrule
        GTC \cite{videograin2025yang, rave2023kara, flatten2023cong, fresco2024yang} & Global Temporal Consistency & Temporal consistency across consecutive frames (overall video smoothness). \\
        
        GTF \cite{videograin2025yang, rave2023kara, flatten2023cong, tokenflow2023geyer} & Global Textual Faithfulness & Alignment between edited frames and the global edit prompt. \\
        
        FA \cite{fresco2024yang} & Frame Accuracy & Percentage of frames more aligned with the global target prompt than the global source prompt. \\
        
        LTC & Local Temporal Consistency & Temporal consistency of each cropped instance across consecutive frames. \\
        
        LTF \cite{instancediffusion2024wang, spatext2023avrahami} & Local Textual Faithfulness & Alignment between each cropped instance and its corresponding target instance caption. \\
        
        IA & Instance Accuracy & Percentage of instances more aligned with the target instance caption than the source instance caption. \\
        
        CIA (Ours) & Cross-Instance Accuracy & Correct instance--caption assignment; penalizes cross-instance leakage. \\
        
        SSIM \cite{ssim} & Structural Similarity Index Measure & Background structural similarity. \\
        
        LPIPS \cite{slicedit2024cohen, dreammotion2024jeong} & Learned Perceptual Image Patch Similarity & Perceptual similarity for background consistency. \\
        \bottomrule
        \end{tabular}
    }
    \vspace{-0.5cm}
\end{table*}

\noindent
\textbf{Implementation details}.
We evaluate PRIMEdit on our MIVE dataset (Sec. \ref{subsec:dataset_construction}), editing 12–32 frames per video, using a single NVIDIA RTX A6000 GPU.
We implement our PRIMEdit on top of a baseline that integrates Stable Diffusion \cite{ldm2022rombach} v1.5, AnimateDiff \cite{animatediff2023guo} motion modules and ControlNet \cite{controlnet2023zhang} with depth \cite{midas2020ranftl} as our condition.
We perform 100 DDIM inversion steps with empty text \cite{flatten2023cong}, followed by $T=50$ denoising steps, using a CFG scale of 12.5 \cite{cfg2022ho}.
IPR is applied in the first 10\% of denoising steps, after which we revert to vanilla cross-attention \cite{ldm2022rombach}.
SNS is used for the first 40\% of steps, followed by PNS, with re-inversion steps set to $l=2$.

\noindent
\textbf{Evaluation metrics}.
To evaluate the edited frames, we report standard metrics:
(i) Global Temporal Consistency (GTC) \cite{videograin2025yang, rave2023kara, flatten2023cong, fresco2024yang}: average cosine similarity of CLIP \cite{clip2021radford} image embeddings between consecutive frames,
(ii) Global Textual Faithfulness (GTF) \cite{videograin2025yang, rave2023kara, flatten2023cong, tokenflow2023geyer}: average similarity between the frames and global edit prompts, and
(iii) Frame Accuracy (FA) \cite{fresco2024yang}: percentage of frames with greater similarity to the target than the source prompt.

Global metrics assess overall frame quality but they overlook nuances in individual instance edits, which is essential for multi-instance tasks.
To address this, we use Local Temporal Consistency (LTC), Local Textual Faithfulness (LTF) \cite{instancediffusion2024wang, spatext2023avrahami}, and Instance Accuracy (IA).
We crop each instance using the bounding boxes inferred from their masks and add padding to preserve their aspect ratios.
LTF \cite{instancediffusion2024wang, spatext2023avrahami} is calculated as the average cosine similarity between the CLIP \cite{clip2021radford} image embeddings of each cropped instance and the text embeddings of its instance caption.
LTC is similarly measured as the average cosine similarity between cropped instances across consecutive frames.
The IA is the percentage of instances with a higher similarity to the target instance caption than to the source instance caption.
We also quantify cross-instance leakage through our proposed CIA Score (Sec. \ref{subsec:cia}) and background leakage through SSIM \cite{ssim} and LPIPS \cite{slicedit2024cohen, dreammotion2024jeong}.
Table~\ref{table:metric_summary} provides a concise summary of all evaluation metrics and their definitions.

\begin{figure*}[t]
  \centering
    \includegraphics[width=0.97\linewidth]{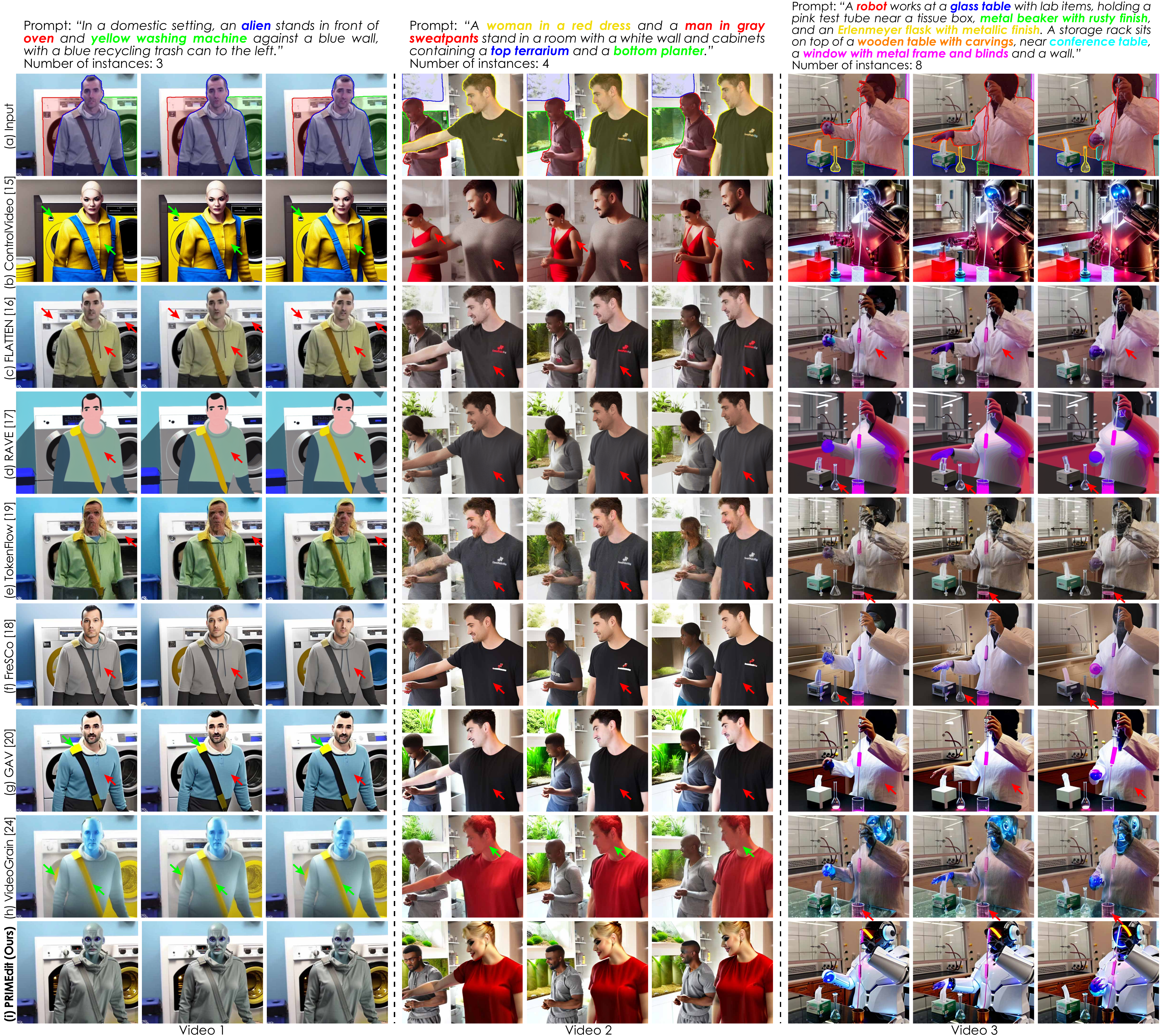}
    \vspace{-0.35cm}
    \caption{
        Qualitative comparison for three videos (with increasing difficulty from left to right) in our MIVE dataset.
        (a) shows the color-coded masks overlaid on the input frames to match the corresponding instance captions.
        (b)-(d) use global target captions for editing.
        (e) uses global source caption, instance source captions, global target caption, and instance target captions along with bounding boxes (omitted in (a) for better visualization).
        (f) uses global source caption, instance source captions, global target caption, and instance target captions, along with instance masks.
        Our PRIMEdit in (g) uses instance captions and masks.
        Unfaithful editing examples are shown in \textcolor{red}{red arrow} and attention leakage are shown in \textcolor{green}{green arrow}.
    }
    \label{fig:7_sotacomparison}
    \vspace{-0.5cm}
\end{figure*}

\begin{table*}[t]
    \small
    \caption{
        Quantitative results for multi-instance video editing.
        The best and second best scores are shown in \textcolor{red}{\textbf{red}} and \textcolor{blue}{\underline{blue}}, respectively.
    }
    \vspace{-0.4cm}
    \begin{center}
    \setlength\tabcolsep{6pt}{
        \renewcommand{\arraystretch}{1.1}
        \scalebox{0.72}{
          \begin{tabular}{l | c | c | c c c | c c c | c c c | c c c}
            \toprule
            \multirow{2}{*}{Methods} & \multirow{2}{*}{Venue} & Editing & \multicolumn{3}{c}{Global Scores} & \multicolumn{3}{|c|}{Local Scores} & \multicolumn{3}{c|}{Leakage Scores} & \multicolumn{3}{c}{User Study} \\
            & & Scopes & GTC $\uparrow$ & GTF $\uparrow$ & FA $\uparrow$ & LTC $\uparrow$ & LTF $\uparrow$ & IA $\uparrow$ & CIA (\textbf{Ours}) $\uparrow$ & SSIM $\uparrow$ & LPIPS $\downarrow$ & TC $\uparrow$ & TF $\uparrow$ & Leakage $\uparrow$ \\
            \midrule
            ControlVideo \cite{controlvideo2023zhang} & ICLR'24 & Global & \textcolor{red}{\textbf{0.9745}} & \textcolor{blue}{\underline{0.2738}} & \textcolor{red}{\textbf{0.8850}} & 0.9548 & 0.1960 & \textcolor{blue}{\underline{0.4944}} & 0.4963 & 0.5327 & 0.4242 & 0.0167 & 0.0656 & 0.0208 \\
            FLATTEN \cite{flatten2023cong} & ICLR'24 & Global & 0.9678 & 0.2389 & 0.2627 & 0.9507 & 0.1881 & 0.2463 & 0.5109 & 0.8220 & 0.1314 & \textcolor{blue}{\underline{0.3260}} & 0.0594 & \textcolor{blue}{\underline{0.1646}} \\
            RAVE \cite{rave2023kara} & CVPR'24 & Global & 0.9675 & 0.2728 & 0.5769 & \textcolor{blue}{\underline{0.9551}} & 0.1869 & 0.3504 & 0.4950 & 0.7699 & 0.2188 & 0.0667 & 0.0354 & 0.0458 \\
            TokenFlow \cite{tokenflow2023geyer} & ICLR'24 & Global & \textcolor{blue}{\underline{0.9686}} & 0.2571 & 0.5617 & 0.9477 & 0.1868 & 0.3499 & 0.5319 & 0.7991 & 0.1807 & 0.0125 & 0.0521 & 0.0531 \\
            FreSCo \cite{fresco2024yang} & CVPR'24 & Global & 0.9543 & 0.2528 & 0.4202 & 0.9324 & 0.1860 & 0.2960 & 0.5172 & 0.7567 & 0.2427 & 0.0823 & 0.0073 & 0.0104 \\
            GAV \cite{groundavideo2024jeong} & ICLR'24 & Local, Multiple & 0.9665 & 0.2574 & 0.5544 & 0.9518 & 0.1895 & 0.3733 & 0.5516 & 0.8618 & 0.0914 & 0.0792 & 0.0938 & 0.1000 \\
            VideoGrain \cite{videograin2025yang} & ICLR'25 & Local, Multiple & 0.9600 & \textcolor{red}{\textbf{0.2803}} & 0.7430 & 0.9423 & \textcolor{blue}{\underline{0.2026}} & 0.4798 & \textcolor{blue}{\underline{0.5868}} & \textcolor{blue}{\underline{0.8929}} & \textcolor{red}{\textbf{0.0488}} & 0.0802 & \textcolor{blue}{\underline{0.1177}} & 0.1542 \\
            \textbf{PRIMEdit (Ours)} & - & Local, Multiple & 0.9677 & 0.2632 & \textcolor{blue}{\underline{0.7707}} & \textcolor{red}{\textbf{0.9552}} & \textcolor{red}{\textbf{0.2048}} & \textcolor{red}{\textbf{0.5369}} & \textcolor{red}{\textbf{0.6705}} & \textcolor{red}{\textbf{0.9008}} & \textcolor{blue}{\underline{0.0586}} & \textcolor{red}{\textbf{0.3365}} & \textcolor{red}{\textbf{0.5687}} & \textcolor{red}{\textbf{0.4510}} \\
            \bottomrule
          \end{tabular}
        }
    } 
    \end{center}
    \vspace{-0.6cm}
    \label{table:3_sota_comparison}
\end{table*}

\subsection{Experimental Results}
\label{subsec:experimental_results}

\noindent
We compare our PRIMEdit against recent zero-shot video editing (VE) methods: (i) five global editing methods: ControlVideo \cite{controlvideo2023zhang}, FLATTEN \cite{flatten2023cong}, RAVE \cite{rave2023kara}, FreSCo \cite{fresco2024yang}, TokenFlow \cite{tokenflow2023geyer}, and (ii) two multi-object editing methods: GAV \cite{groundavideo2024jeong} and VideoGrain \cite{videograin2025yang}.
The five global editing methods use single global source and global target captions.
For the multi-object editing methods, GAV and VideoGrain utilize a combination of a global source caption, instance source captions, a global target caption, and instance target captions along with bounding box and mask conditions per instance.
Our PRIMEdit requires instance masks and local target captions only.
We exclude Video-P2P \cite{videop2p2024liu} since it edits one instance at a time, accumulating error when used in multi-instance editing scenarios.
We also exclude ConMo \cite{conmo2025gao}, which, despite its ability to disentangle individual object motions, lacks explicit local control in assigning instance captions to specific target instances (see \textit{Appendix}).
For each baseline above, we use its default settings.

\noindent
\textbf{Qualitative comparison}.
We present qualitative comparisons in Fig. \ref{fig:7_sotacomparison}.
As shown, ControlVideo, GAV, and VideoGrain suffer from attention leakage (e.g. \textit{``yellow''} affecting the \textit{``alien''} in Video 1 of Fig. \ref{fig:7_sotacomparison}-(b), (g), and (h)) while FLATTEN, RAVE, TokenFlow, FreSCo, and GAV exhibit unfaithful editing for all the examples (Fig. \ref{fig:7_sotacomparison}-(c) to Fig. \ref{fig:7_sotacomparison}-(g)).
Moreover, ControlVideo exhibits a mismatch in editing instances as shown in Video 2 of Fig. \ref{fig:7_sotacomparison}-(b), incorrectly assigning \textit{``woman in a red dress''} to the man on the left.
In contrast, our PRIMEdit confines the edits within the instance masks, faithfully adheres to instance captions, and does not exhibit mismatch in instance editing.
More demo videos as well as additional results for the DAVIS dataset are shown in the \textit{Appendix}.

\noindent
\textbf{Quantitative comparison}.
Table \ref{table:3_sota_comparison} shows the quantitative comparisons.
The SOTA methods yield (i) low LTF and IA scores, indicating less accurate instance-level editing, and (ii) low CIA scores, reflecting significant attention leakage.
In contrast, our PRIMEdit achieves the best scores in key multi-instance VE metrics (LTC, LTF, IA, and CIA), demonstrating its superior ability in maintaining temporal consistency, fine-grained textual alignment, and instance-aware modifications.
Note also that our PRIMEdit exhibits a robust performance for background preservation metrics, achieving the best SSIM and second-best LPIPS scores \textit{without} relying on an inpainting model like GAV \cite{groundavideo2024jeong}.
This further underscores its effectiveness in isolating instance edits while preserving and preventing the editing leakage in the background.

Despite achieving the highest scores in local metrics, leakage minimization, and background preservation, PRIMEdit exhibits a slight disadvantage in global metrics.
Our GTC score ranks fourth, which we attribute to prior SOTA methods favoring frame reconstruction, thereby preserving the high GTC of the original video.
Similarly, our GTF ranks fourth, while our FA ranks second only to ControlVideo.
We attribute this to CLIP's limitations in compositional reasoning \cite{tifa2023hu, crepe2024ma}, which hinders its ability to capture complex relationships between instances in the global caption and frames.
As a result, CLIP-based evaluations may lead to inflated GTF and FA scores despite inaccuracies in instance edits (e.g., swapped edits between two people in Video 2 of Fig. \ref{fig:7_sotacomparison}-(b)).

\begin{figure}[t]
  \centering
    \includegraphics[width=0.95\linewidth]{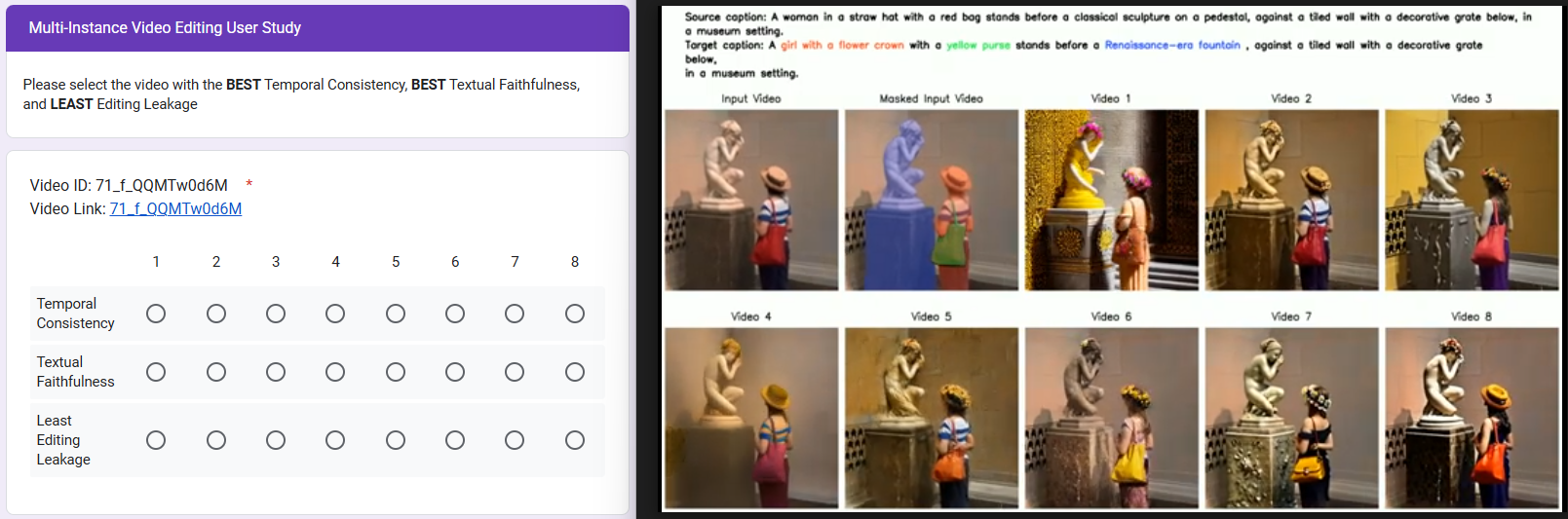}
    \vspace{-0.25cm}
    \caption{
        Our user study interface and questionnaire form.
        Participants are shown an input video with a source caption, an annotated video with a target caption, and 8 randomly ordered edited videos from PRIMEdit and 7 other SOTA VE methods.
        Each instance mask is color-coded to correspond with its target instance caption.
        Participants select the video with the \textbf{Best Temporal Consistency}, \textbf{Best Textual Faithfulness}, and \textbf{Least Editing Leakage}.
    }
   \label{fig:8_user_study}
   \vspace{-0.2cm}
\end{figure}

\noindent
\textbf{User study}. We conduct a user study to compare our PRIMEdit with the SOTA baselines, and report the results in Table \ref{table:3_sota_comparison}.
We select 30 videos from our dataset covering diverse scenarios (varying the number of objects per clip, the number of instances per object class, and the instance sizes).
We task 32 participants to select the method with the best temporal consistency (TC), textual faithfulness (TF), and minimal editing leakage (Leakage).
Fig. \ref{fig:8_user_study} shows our user study interface and the questionnaire form.
Our PRIMEdit demonstrates clear superiority against previous SOTA methods with 33.65\% chance of being selected as the best for TC, 56.87\% for TF, and 45.10\% for the least amount of leakage.

\subsection{Ablation Studies}

\begin{table}[t]
    \small
    \caption{
        Ablation study of core components (IPR (Sec. \ref{subsec:ipr}) and DMS (Sec. \ref{subsec:dms})) and module design choices.
    }
    \vspace{-0.4cm}
    \begin{center}
        \setlength\tabcolsep{3.5pt}{
            \renewcommand{\arraystretch}{1.1}
            \scalebox{0.75}{
              \begin{tabular}{l | l | c c c | c c c}
                \toprule
                \multicolumn{2}{c|}{Methods} & LTC $\uparrow$ & LTF $\uparrow$ & IA $\uparrow$ & CIA $\uparrow$ & SSIM $\uparrow$ & LPIPS $\downarrow$ \\

                \midrule

                \multirow{4}{*}{\rotatebox{90}{Core Comp.}}
                  & DMS only & \textcolor{red}{\textbf{0.9564}} & 0.2018 & 0.4920 & 0.6497 & 0.9007 & 0.0600 \\
                  & IPR + SNS only & 0.9503 & \textcolor{blue}{\underline{0.2047}} & 0.5288 & \textcolor{red}{\textbf{0.6743}} & \textcolor{red}{\textbf{0.9063}} & \textcolor{red}{\textbf{0.0500}} \\
                  & IPR + PNS only & 0.9542 & 0.2044 & \textcolor{blue}{\underline{0.5311}} & \textcolor{blue}{\underline{0.6718}} & 0.8993 & \textcolor{blue}{\underline{0.0585}} \\
                  & \textbf{Ours, Full} &\textcolor{blue}{\underline{0.9552}} & \textcolor{red}{\textbf{0.2048}} & \textcolor{red}{\textbf{0.5369}} & 0.6705 & \textcolor{blue}{\underline{0.9008}} & 0.0586 \\
                
                \midrule
                
                \multirow{3}{*}{\rotatebox{90}{IPR}}
                  & No Modulation \cite{ldm2022rombach} & \textcolor{red}{\textbf{0.9564}} & 0.2018 & 0.4920 & 0.6497 & \textcolor{blue}{\underline{0.9007}} & 0.0600 \\
                  & Dense Diffusion \cite{densediffusion2023kim} & 0.9530 & \textcolor{red}{\textbf{0.2078}} & \textcolor{red}{\textbf{0.5845}} & \textcolor{red}{\textbf{0.6774}} & 0.8977 & \textcolor{blue}{\underline{0.0604}} \\
                  & \textbf{Ours, Full} &\textcolor{blue}{\underline{0.9552}} & \textcolor{blue}{\underline{0.2048}} & \textcolor{blue}{\underline{0.5369}} & \textcolor{blue}{\underline{0.6705}} & \textcolor{red}{\textbf{0.9008}} & \textcolor{red}{\textbf{0.0586}} \\
                  
                \midrule			
                
                \multirow{4}{*}{\rotatebox{90}{DMS}}
                  & Only SNS & 0.9503 & 0.2047 & 0.5288 & \textcolor{red}{\textbf{0.6743}} & \textcolor{red}{\textbf{0.9063}} & \textcolor{red}{\textbf{0.0500}} \\
                  & Only PNS & \textcolor{blue}{\underline{0.9542}} & 0.2044 & 0.5311 & 0.6718 & 0.8993 & 0.0585 \\
                  & SNS + PNS (no re-inv) & 0.9535 & \textcolor{red}{\textbf{0.2051}} & 0.5323 & \textcolor{blue}{\underline{0.6728}} & \textcolor{blue}{\underline{0.9040}} & \textcolor{blue}{\underline{0.0525}} \\
                  & \textbf{Ours, Full} & \textcolor{red}{\textbf{0.9552}} & \textcolor{blue}{\underline{0.2048}} & \textcolor{red}{\textbf{0.5369}} & 0.6705 & 0.9008 & 0.0586 \\
                  
                \bottomrule
              \end{tabular}
            }
        } 
    \end{center}
    \vspace{-0.6cm}
    \label{table:4_ablations}
\end{table}

\begin{figure}[t]
    \centering
    \includegraphics[width=0.93\linewidth]{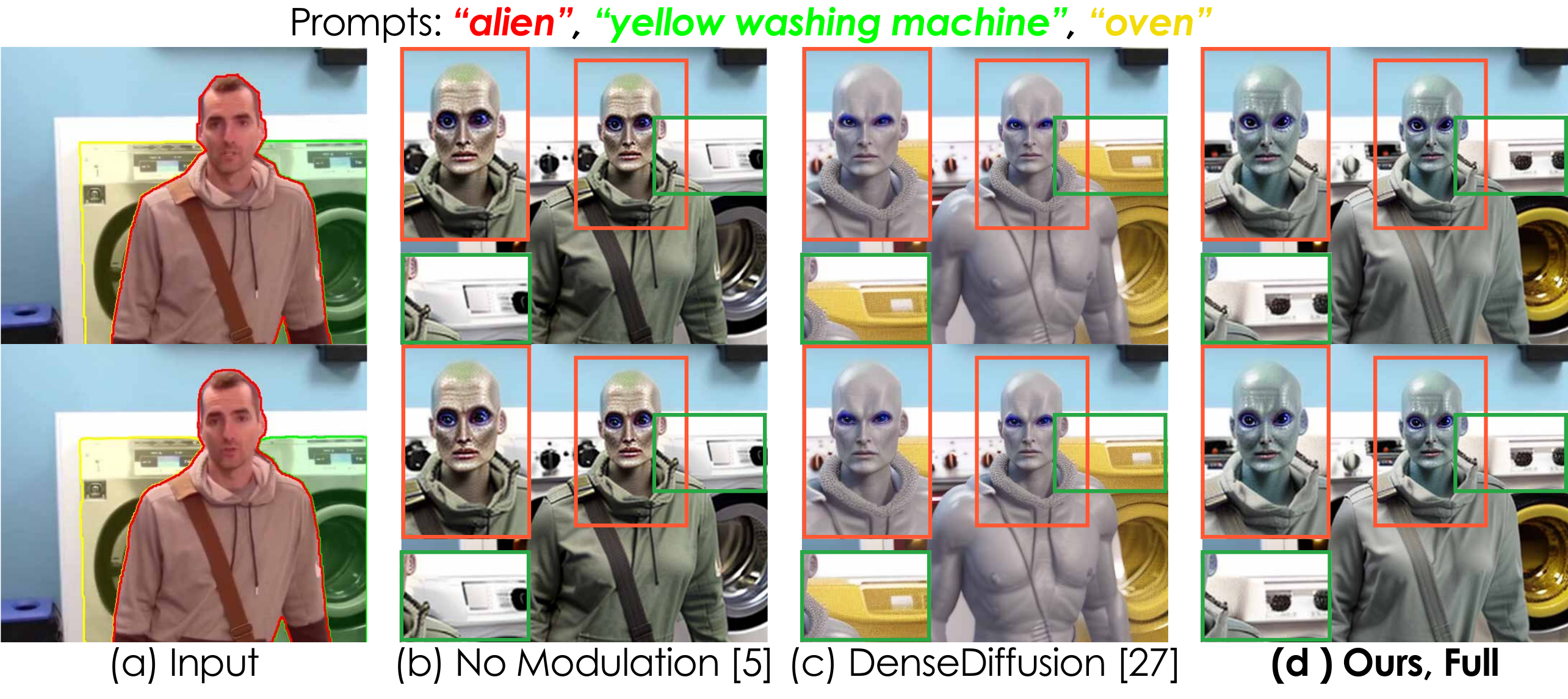}
    \vspace{-0.4cm}
    \caption{
        Ablation study on IPR (Sec. \ref{subsec:ipr}).
    }
    \label{fig:9_ipr_ablation_study}
    \vspace{-0.4cm}
\end{figure}

\noindent
In our ablation studies, we report key multi-instance VE metrics in Table~\ref{table:4_ablations} and provide Global Scores in the \textit{Appendix}.

\noindent
\textbf{Ablation study on core components}.
Table~\ref{table:4_ablations} (top) reports the quantitative results of the core component ablations.
We first disable IPR, equivalent to vanilla cross-attention \cite{ldm2022rombach}, and use a DMS-only baseline.
As expected, without local control, this setting yields lower LTF, IA, and CIA compared to IPR-based variants, although it achieves the highest LTC, indicating strong temporal consistency from full DMS harmonization.

Since IPR edits one instance at a time, we \textit{partially} activate DMS using SNS only or PNS only.
IPR${+}$SNS only achieves the best CIA, SSIM, and LPIPS, indicating that early disentangled editing with IPR reduces cross-instance and background leakage.
However, this setting yields the lowest LTC due to the lack of harmonization from the PNS.
IPR${+}$PNS only improves LTC and IA, suggesting that PNS harmonization enhances temporal consistency and instance discrimination.

Finally, activating both IPR and full DMS achieves the best overall balance, yielding the highest LTF and IA, second-best LTC and SSIM, and competitive CIA and LPIPS.
This demonstrates that combining IPR with full DMS enhances editing faithfulness and instance discrimination while preserving strong temporal coherence.

\noindent
\textbf{Ablation study on IPR}.
We conduct an ablation study on our IPR, with qualitative results presented in Fig. \ref{fig:9_ipr_ablation_study} and quantitative results in Table \ref{table:4_ablations} (middle).
`No Modulation' in Table \ref{table:4_ablations} refers to the vanilla cross-attention from the latent diffusion model \cite{ldm2022rombach}.
Note that in this ablation, we use DMS to isolate the contribution of IPR, since without DMS, `No Modulation' cannot perform multi-instance editing.
As shown in Fig. \ref{fig:9_ipr_ablation_study}-(b), the model tends to reconstruct the input frames when the modulation is omitted.
The reconstruction thereby preserves the high LTC of the input but suffers from unfaithful editing (low LTF and IA) and leakage (low CIA).
Introducing the cross-attention modulation of DenseDiffusion \cite{densediffusion2023kim} reduces LTC but improves faithfulness and minimizes leakage.
However, as presented in the green box of Fig. \ref{fig:9_ipr_ablation_study}-(c) and \cite{videocomposer2024wang}, the cross-attention modulation struggles to generate visually appealing results when the computationally expensive spatio-temporal attention modulation is absent.
In contrast, our IPR effectively reduces background artifacts (best SSIM and LPIPS), maintains temporal consistency (second-best), enhances edit faithfulness (second-best LTF and IA), and minimizes leakage (second-best CIA), while operating solely on the cross-attention layers.

\begin{figure}[t]
    \centering
    \includegraphics[width=0.95\linewidth]{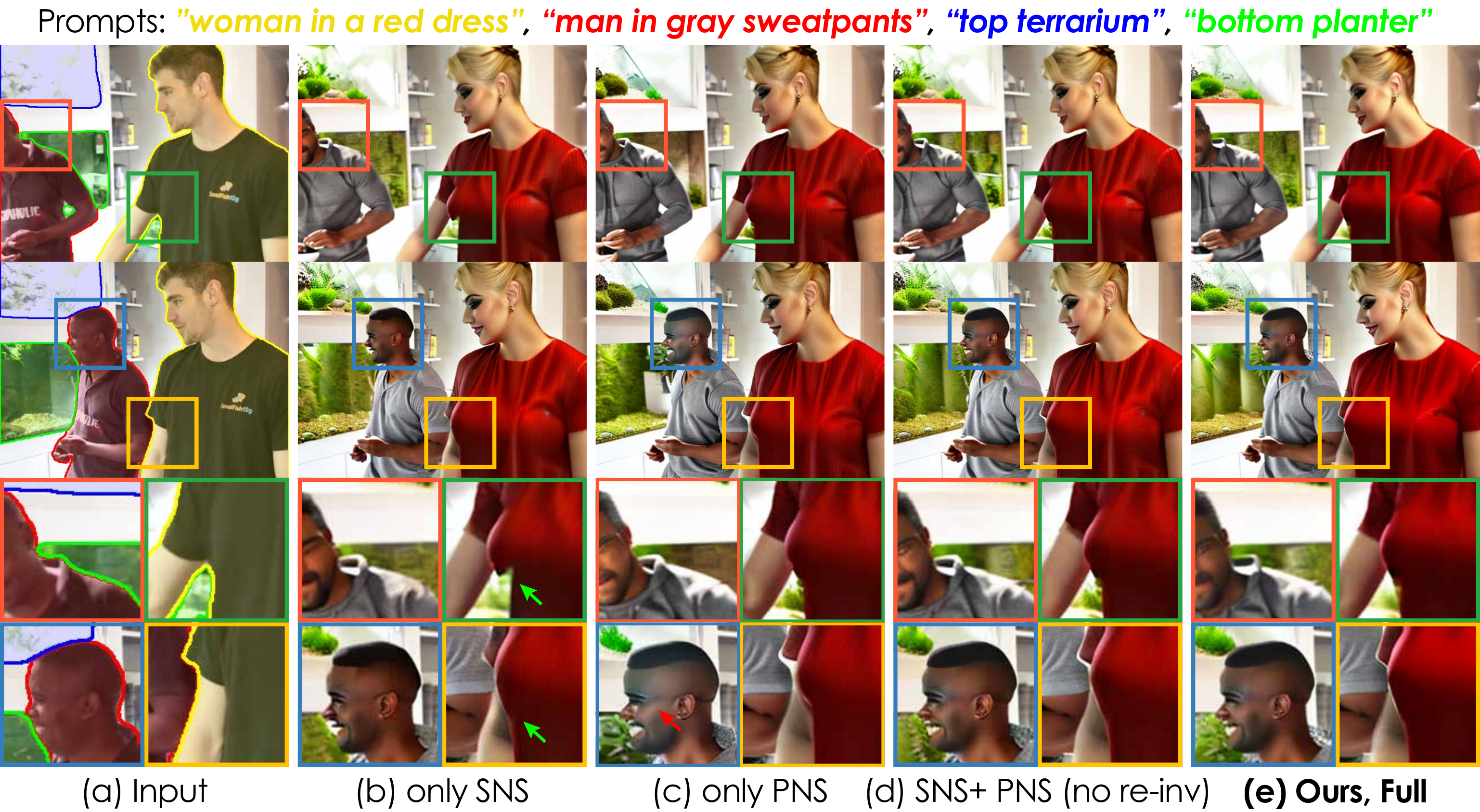}
    \vspace{-0.3cm}
    \caption{
        Ablation study on DMS (Sec. \ref{subsec:dms}).
    }
    \label{fig:10_dms_ablation_study}
    \vspace{-0.4cm}
\end{figure}

\noindent
\textbf{Ablation study on DMS}.
Fig. \ref{fig:10_dms_ablation_study} and Table \ref{table:4_ablations} (bottom) show the qualitative and quantitative ablations, respectively.
From Fig. \ref{fig:10_dms_ablation_study} and Fig. \ref{table:4_ablations}, we observe the following:
(i) using only SNS prevents excessive leakage (high CIA) but suffers from temporal inconsistency (lowest LTC) as indicated by the green arrow in Fig. \ref{fig:10_dms_ablation_study}-(b);
(ii) using only PNS improves LTC as objects are fused early in the denoising process but increases leakage (decrease in CIA), as indicated by the red arrow in Fig. \ref{fig:10_dms_ablation_study}-(c), where the green color from the plants leaks onto the man's face;
(iii) SNS followed by PNS mitigates the issues observed in (i) and (ii), improving both CIA and LTC; and
(iv) adding re-inversion improves IA and boosts LTC by harmonizing instances with a slight decline in CIA Score.
Overall, our final model shows a balanced trade-off between LTC, faithfulness, leakage, and background preservation.

\subsection{Analyses on Difficulty Levels of Dataset}
\label{subsec:difficulty_level}

\begin{table*}[t]
  \small
  \renewcommand{\arraystretch}{1.1}
  \caption{\
    Quantitative comparison based on instance size. We only show Local Scores since these are the only scores that can be computed depending on the instance sizes.
  }
  \vspace{-0.4cm}
  \begin{center}
  \setlength\tabcolsep{6pt}{
    \scalebox{0.75}{
      \begin{tabular}{l | c | c | c c c | c c c | c c c}
        \toprule
        \multirow{2}{*}{Methods} & \multirow{2}{*}{Venues} & Editing & \multicolumn{3}{c}{Local Scores (Small)} & \multicolumn{3}{|c|}{Local Scores (Medium)} & \multicolumn{3}{c}{Local Scores (Large)}\\
        & & Scopes & LTC $\uparrow$ & LTF $\uparrow$ & IA $\uparrow$ & LTC $\uparrow$ & LTF $\uparrow$ & IA $\uparrow$ & LTC $\uparrow$ & LTF $\uparrow$ & IA $\uparrow$ \\
        \midrule
        ControlVideo \cite{controlvideo2023zhang} & ICLR'24 & Global & 0.9202 & 0.1610 & 0.3308 & \textcolor{red}{\textbf{0.9552}} & 0.1993 & \textcolor{blue}{\underline{0.5258}} & 0.9554 & 0.2008 & \textcolor{blue}{\underline{0.5338}} \\
        FLATTEN \cite{flatten2023cong} & ICLR'24 & Global & 0.9128 & 0.1707 & 0.3228 & 0.9515 & 0.1890 & 0.2366 & 0.9521 & 0.1891 & 0.2285 \\
        RAVE \cite{rave2023kara} & CVPR'24 & Global & \textcolor{red}{\textbf{0.9255}} & 0.1710 & 0.3685 & \textcolor{red}{\textbf{0.9552}} & 0.1898 & 0.3580 & \textcolor{red}{\textbf{0.9557}} & 0.1898 & 0.3510 \\
        TokenFlow \cite{tokenflow2023geyer} & ICLR'24 & Global & 0.9209 & 0.1701 & \textcolor{blue}{\underline{0.4015}} & 0.9490 & 0.1883 & 0.3451 & 0.9490 & 0.1887 & 0.3405 \\
        FreSCo \cite{fresco2024yang} & CVPR'24 & Global & 0.9021 & 0.1681 & 0.3686 & 0.9347 & 0.1869 & 0.2948 & 0.9348 & 0.1867 & 0.2825\\
        GAV \cite{groundavideo2024jeong} & ICLR'24 & Local, Multiple & 0.9182 & 0.1716 & 0.3358 & 0.9514 & 0.1907 & 0.3810 & 0.9521 & 0.1908 & 0.3807 \\
        VideoGrain \cite{videograin2025yang} & ICLR'25 & Local, Multiple & 0.9191 & \textcolor{red}{\textbf{0.1750}} & \textcolor{red}{\textbf{0.4419}} & 0.9427 & \textcolor{blue}{\underline{0.2052}} & 0.4991 & 0.9427 & \textcolor{blue}{\underline{0.2065}} & 0.5025 \\
        \textbf{PRIMEdit (Ours)} & - & Local, Multiple & \textcolor{blue}{\underline{0.9235}} & \textcolor{blue}{\underline{0.1728}} & 0.3870 & \textcolor{blue}{\underline{0.9551}} & \textcolor{red}{\textbf{0.2078}} & \textcolor{red}{\textbf{0.5502}} & \textcolor{blue}{\underline{0.9556}} & \textcolor{red}{\textbf{0.2093}} & \textcolor{red}{\textbf{0.5604}} \\
        \bottomrule
      \end{tabular}
      }
  }
  \end{center}
  \vspace{-0.8cm}
  \label{table:5_quantitative_comparison_instance_size}
\end{table*}

\begin{figure*}[t]
    \centering
    \includegraphics[width=0.95\linewidth]{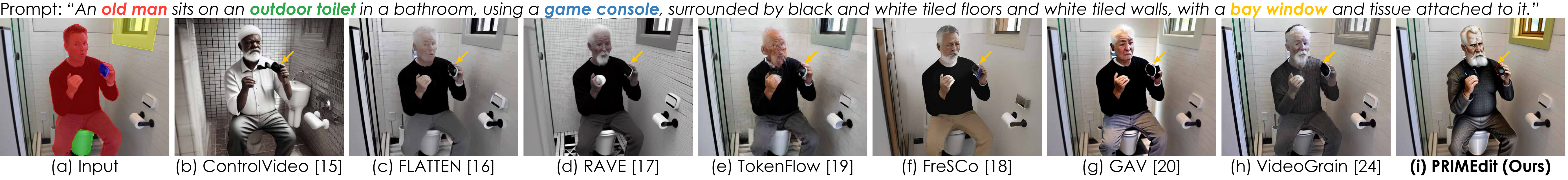}
    \vspace{-0.3cm}
    \caption{
        Qualitative comparison based on instance size.
        The yellow arrows highlight failure cases on small objects (e.g., “game console”), highlighting the difficulty of small object editing in diffusion-based video editing.
    }
    \label{fig:11_qualitative_instance_size}
    \vspace{-0.2cm}
\end{figure*}

\noindent
Here, we deeply analyze the performance of baseline methods and our PRIMEdit based on: (i) the instance sizes and (ii) the number of instances to know if there is any bias in the performance.
For the experiments on instance size variations, we evaluate the local scores for different object sizes according to the COCO dataset categories \cite{coco2014lin} on 200 videos of MIVE dataset.
Meanwhile, for experiments on various numbers of instances, we classify 200 videos into three categories, depending on the number of edited instances per video. 
In addition, we also perform additional experiments on single instance video editing, where we select the largest instance to be edited.

\noindent
\textbf{Results for different instance sizes.}
Table \ref{table:5_quantitative_comparison_instance_size} and Fig. \ref{fig:11_qualitative_instance_size} present quantitative and qualitative comparisons between our PRIMEdit and the baseline approaches across different instance sizes, respectively.
We utilize the categorization of instance sizes in the COCO dataset \cite{coco2014lin} as follows: (i) small instances have areas less than $32^2$ pixels, (ii) medium instances have areas between $32^2$ and $96^2$ pixels, and (iii) large instances have areas greater than $96^2$ pixels. 
Across all videos, there are 69 small, 297 medium, and 434 large instances.
The area of each instance is computed by averaging its areas across all frames per video. As shown in Table \ref{table:5_quantitative_comparison_instance_size}, the LTC scores closely align with the main quantitative comparison in Table \ref{table:3_sota_comparison}.
However, we note that our method shows superior temporal consistency compared to other multi-instance VE methods.

For medium and large instances, our PRIMEdit achieves the best performance in textual faithfulness (LTF and IA) and the second-best (comparable to the best) performance in LTC.
Both our PRIMEdit and VideoGrain significantly outperform the global video editing methods and the bounding box-based method (GAV) in LTF and IA metrics.
This highlights the advantage of using masks over bounding boxes as conditioning mechanisms for multi-instance VE, allowing for more precise and context-aware instance modifications.

We further observe a trade-off between editing fidelity and temporal consistency in the multi-instance VE setting, as reflected in the performance comparison (Table \ref{table:5_quantitative_comparison_instance_size}) for our PRIMEdit, GAV and VideoGrain.
There is a tendency of temporal consistency drop when we maximize the editing fidelity in multi-instance setting.
Nonetheless, our PRIMEdit still maintains competitive LTC performance against global editing methods while achieving the best editing fidelity in terms of LTF and IA scores.

For the small instances, all methods yield lower LTF scores compared to the medium and large instances as shown in Table \ref{table:5_quantitative_comparison_instance_size}.
This behavior is evident in the qualitative results, as indicated by the yellow arrows in Fig. \ref{fig:11_qualitative_instance_size}, where all methods struggle to edit the \textit{``game console''}.
This suggests that small object editing remains a challenge for diffusion-based VE methods, largely due to the spatial size reduction from the LDM's VAE encoder \cite{ldm2022rombach}.
Compared to other multi-instance methods, our PRIMEdit achieves better LTC performance while maintaining competitive LTF and IA scores.
One potential solution is to increase the weight of the IPR component to improve editing fidelity.
However, we find that doing so often results in reduced LTC performance.
Hence, for small instances, we adopt a balanced trade-off that prioritizes competitive LTC without compromising editing fidelity.

\begin{table*}[t]
  \small
  \renewcommand{\arraystretch}{1.1}
  \caption{\
    Quantitative comparison for multi-instance video editing on various number of instances.
  }
  \vspace{-0.4cm}
  \begin{center}
    \setlength\tabcolsep{6pt}{
      \scalebox{0.75}{
          \begin{tabular}{l | c | c | c c c | c c c | c c c}
            \toprule
            \multirow{2}{*}{Methods} & \multirow{2}{*}{Venues} & Editing & \multicolumn{3}{c}{Global Scores} & \multicolumn{3}{|c|}{Local Scores} & \multicolumn{2}{c}{Leakage Scores} \\
            & & Scopes & GTC $\uparrow$ & GTF $\uparrow$ & FA $\uparrow$ & LTC $\uparrow$ & LTF $\uparrow$ & IA $\uparrow$ & CIA (\textbf{Ours}) $\uparrow$ & SSIM $\uparrow$ & LPIPS $\downarrow$ \\
            \midrule
            \multicolumn{11}{c}{Editing on One (1) Instance ($\text{MIVE}_E$) - 200 Videos}\\
            \midrule
            ControlVideo \cite{controlvideo2023zhang} & ICLR'24 & Global & \textcolor{red}{\textbf{0.9755}} & \textcolor{blue}{\underline{0.2717}} & \textcolor{red}{\textbf{0.6922}} & \textcolor{blue}{\underline{0.9611}} & \textcolor{blue}{\underline{0.2048}} & \textcolor{blue}{\underline{0.5351}} & - & 0.4130 & 0.5090 \\
            FLATTEN \cite{flatten2023cong} & ICLR'24 & Global & 0.9699 & 0.2538 & 0.3140 & 0.9586 & 0.1826 & 0.1971 & - & 0.7783 & 0.1567 \\
            RAVE \cite{rave2023kara} & CVPR'24 & Global & 0.9691 & \textcolor{red}{\textbf{0.2774}} & 0.4243 & 0.9599 & 0.1857 & 0.3374 & - & 0.7044 & 0.2716 \\
            TokenFlow \cite{tokenflow2023geyer} & ICLR'24 & Global & 0.9715 & 0.2621 & 0.4413 & 0.9564 & 0.1822 & 0.3006 & - & 0.7546 & 0.2148 \\
            FreSCo \cite{fresco2024yang} & CVPR'24 & Global & 0.9563 & 0.2630 & 0.3200 & 0.9423 & 0.1805 & 0.2515 & - & 0.6894 & 0.3014 \\
            GAV \cite{groundavideo2024jeong} & ICLR'24 & Local, Multiple & 0.9701 & 0.2537 & 0.4453 & 0.9584 & 0.1851 & 0.3350 & - & 0.8340 & 0.0999 \\
            VideoGrain \cite{videograin2025yang} & ICLR'25 & Local, Multiple & 0.9659 & 0.2696 & 0.5474 & 0.9515 & 0.2020 & 0.4531 & - & \textcolor{blue}{\underline{0.8731}} & \textcolor{red}{\textbf{0.0570}} \\
            \textbf{PRIMEdit (Ours)} & - & Local, Multiple & \textcolor{blue}{\underline{0.9737}} & 0.2534 & \textcolor{blue}{\underline{0.6192}} & \textcolor{red}{\textbf{0.9614}} & \textcolor{red}{\textbf{0.2108}} & \textcolor{red}{\textbf{0.5562}} & - & \textcolor{red}{\textbf{0.8807}} & \textcolor{blue}{\underline{0.0661}} \\
            
            \midrule
            \multicolumn{11}{c}{Editing on 1-3 Instances ($\text{MIVE}_M$) - 116 Videos}\\
            \midrule
            ControlVideo \cite{controlvideo2023zhang} & ICLR'24 & Global & \textcolor{red}{\textbf{0.9730}} & \textcolor{blue}{\underline{0.2724}} & \textcolor{red}{\textbf{0.8839}} & 0.9512 & 0.2020 & \textcolor{blue}{\underline{0.5380}} & 0.6191 & 0.4879 & 0.4515 \\
            FLATTEN \cite{flatten2023cong} & ICLR'24 & Global & 0.9659 & 0.2417 & 0.2908 & 0.9484 & 0.1894 & 0.2557 & 0.6053 & 0.7902 & 0.1504 \\
            RAVE \cite{rave2023kara} & CVPR'24 & Global & 0.9661 & 0.2699 & 0.5437 & \textcolor{red}{\textbf{0.9532}} & 0.1886 & 0.3604 & 0.5971 & 0.7396 & 0.2379 \\
            TokenFlow \cite{tokenflow2023geyer} & ICLR'24 & Global & \textcolor{blue}{\underline{0.9686}} & 0.2580 & 0.5701 & 0.9463 & 0.1880 & 0.3476 & 0.6265 & 0.7720 & 0.1974 \\
            FreSCo \cite{fresco2024yang} & CVPR'24 & Global & 0.9535 & 0.2492 & 0.4140 & 0.9326 & 0.1877 & 0.2836 & 0.6084 & 0.7165 & 0.2811 \\
            GAV \cite{groundavideo2024jeong} & ICLR'24 & Local, Multiple & 0.9649 & 0.2528 & 0.5648 & 0.9481 & 0.1919 & 0.3903 & 0.6498 & 0.8397 & 0.0959 \\
            VideoGrain \cite{videograin2025yang} & ICLR'25 & Local, Multiple & 0.9596 & \textcolor{red}{\textbf{0.2775}} & 0.7486 & 0.9413 & \textcolor{blue}{\underline{0.2065}} & 0.5238 & \textcolor{blue}{\underline{0.6712}} & \textcolor{blue}{\underline{0.8736}} & \textcolor{red}{\textbf{0.0526}} \\
            \textbf{PRIMEdit (Ours)} & - & Local, Multiple & 0.9658 & 0.2629 & \textcolor{blue}{\underline{0.7897}} & \textcolor{blue}{\underline{0.9516}} & \textcolor{red}{\textbf{0.2096}} & \textcolor{red}{\textbf{0.5817}} & \textcolor{red}{\textbf{0.7567}} & \textcolor{red}{\textbf{0.8823}} & \textcolor{blue}{\underline{0.0633}} \\
            
            \midrule
            \multicolumn{11}{c}{Editing on 4-7 Instances ($\text{MIVE}_H$) - 66 Videos}\\
            \midrule
            ControlVideo \cite{controlvideo2023zhang} & ICLR'24 & Global & \textcolor{red}{\textbf{0.9760}} & 0.2774 & \textcolor{red}{\textbf{0.8815}} & \textcolor{red}{\textbf{0.9581}} & 0.1874 & \textcolor{blue}{\underline{0.4544}} & 0.3694 & 0.5830 & 0.3920\\
            FLATTEN \cite{flatten2023cong} & ICLR'24 & Global & \textcolor{blue}{\underline{0.9702}} & 0.2372 & 0.2303 & 0.9527 & 0.1854 & 0.2288 & 0.4206 & 0.8585 & 0.1099 \\
            RAVE \cite{rave2023kara} & CVPR'24 & Global & 0.9690 & \textcolor{blue}{\underline{0.2776}} & 0.5821 & 0.9571 & 0.1843 & 0.3399 & 0.3954 & 0.8060 & 0.1957\\
            TokenFlow \cite{tokenflow2023geyer} & ICLR'24 & Global & 0.9686 & 0.2559 & 0.5427 & 0.9487 & 0.1845 & 0.3631 & 0.4504 & 0.8342 & 0.1585\\
            FreSCo \cite{fresco2024yang} & CVPR'24 & Global & 0.9557 & 0.2589 & 0.4274 & 0.9319 & 0.1835 & 0.3179 & 0.4391 & 0.8068 & 0.1934 \\
            GAV \cite{groundavideo2024jeong} & ICLR'24 & Local, Multiple & 0.9679 & 0.2652 & 0.5557 & 0.9553 & 0.1861 & 0.3659 & 0.4692 & 0.8884 & 0.0837 \\
            VideoGrain \cite{videograin2025yang} & ICLR'25 & Local, Multiple &  0.9603 & \textcolor{red}{\textbf{0.2836}} & 0.6938 & 0.9429 & \textcolor{blue}{\underline{0.1972}} & 0.4185 & \textcolor{blue}{\underline{0.5096}} & \textcolor{blue}{\underline{0.9180}} & \textcolor{red}{\textbf{0.0425}} \\
            \textbf{PRIMEdit (Ours)} & - & Local, Multiple & 0.9691 & 0.2622 & \textcolor{blue}{\underline{0.7039}} & \textcolor{blue}{\underline{0.9578}} & \textcolor{red}{\textbf{0.1990}} & \textcolor{red}{\textbf{0.4897}} & \textcolor{red}{\textbf{0.6087}} & \textcolor{red}{\textbf{0.9229}} & \textcolor{blue}{\underline{0.0509}} \\
            
            \midrule
            \multicolumn{11}{c}{Editing on $>$7 Instances ($\text{MIVE}_X$) - 18 Videos}\\
            \midrule
            ControlVideo \cite{controlvideo2023zhang} & ICLR'24 & Global & \textcolor{red}{\textbf{0.9781}} & 0.2692 & \textcolor{blue}{\underline{0.9051}} & \textcolor{blue}{\underline{0.9651}} & 0.1885 & 0.3602 & 0.1698 & 0.6374 & 0.3660 \\
            FLATTEN \cite{flatten2023cong} & ICLR'24 & Global & 0.9717 & 0.2274 & 0.2007 & 0.9584 & 0.1898 & 0.2499 & 0.2339 & 0.8930 & 0.0881 \\
            RAVE \cite{rave2023kara} & CVPR'24 & Global & 0.9711 & \textcolor{blue}{\underline{0.2735}} & 0.7714 & 0.9602 & 0.1857 & 0.3242 & 0.2026 & 0.8323 & 0.1806 \\
            TokenFlow \cite{tokenflow2023geyer} & ICLR'24 & Global & 0.9682 & 0.2552 & 0.5778 & 0.9529 & 0.1878 & 0.3162 & 0.2212 & 0.8455 & 0.1545 \\
            FreSCo \cite{fresco2024yang} & CVPR'24 & Global & 0.9538 & 0.2536 & 0.4339 & 0.9322 & 0.1844 & 0.2954 & 0.2156 & 0.8319 & 0.1760 \\
            GAV \cite{groundavideo2024jeong} & ICLR'24 & Local, Multiple & 0.9715 & 0.2580 & 0.4825 & 0.9628 & 0.1870 & 0.2911 & 0.2210 & 0.9070 & 0.0912 \\
            VideoGrain \cite{videograin2025yang} & ICLR'25 & Local, Multiple & 0.9615 & \textcolor{red}{\textbf{0.2873}} & \textcolor{red}{\textbf{0.9286}} & 0.9478 & \textcolor{red}{\textbf{0.1966}} & \textcolor{blue}{\underline{0.4050}} & \textcolor{blue}{\underline{0.2516}} & \textcolor{blue}{\underline{0.9348}} & \textcolor{red}{\textbf{0.0463}} \\
            \textbf{PRIMEdit (Ours)} & - & Local, Multiple & \textcolor{blue}{\underline{0.9750}} & 0.2682 & 0.8937 & \textcolor{red}{\textbf{0.9689}} & \textcolor{blue}{\underline{0.1945}} & \textcolor{red}{\textbf{0.4209}} & \textcolor{red}{\textbf{0.3413}} & \textcolor{red}{\textbf{0.9390}} & \textcolor{blue}{\underline{0.0559}} \\
            
            \bottomrule
          \end{tabular}
          }
        }
  \end{center}
  \vspace{-0.6cm}
  \label{table:6_quantitative_comparison_instance_number}
\end{table*}

\begin{figure*}[t]
    \centering
    \includegraphics[width=0.93\linewidth]{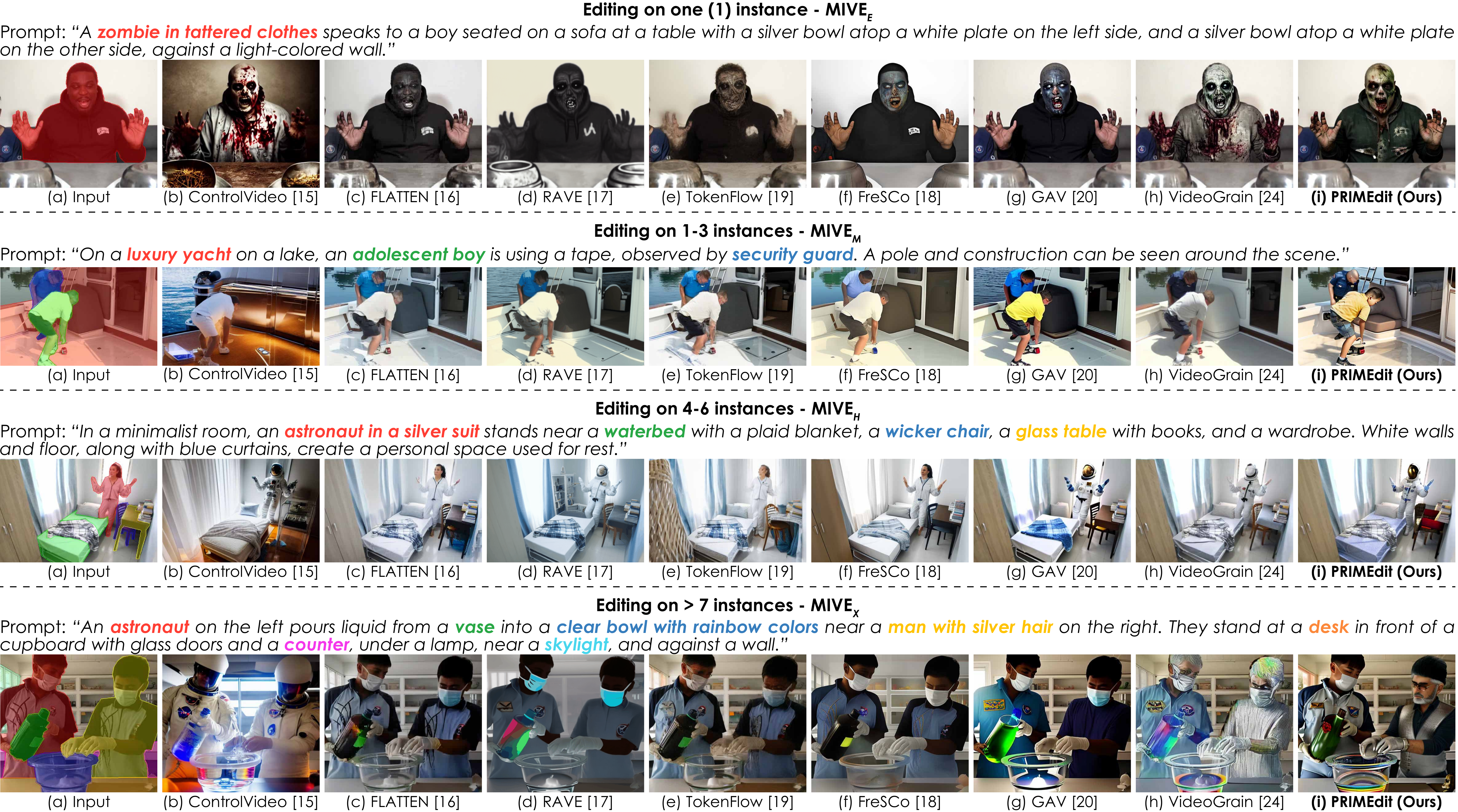}
    \vspace{-0.3cm}
    \caption{
        Qualitative comparison based on the number of instances.
    }
    \label{fig:12_qualitative_instance_count}
    \vspace{-0.2cm}
\end{figure*}

\noindent
\textbf{Results for different numbers of instances.} 
In addition to the comparisons across various instance sizes, we further analyze the performance for the number of edited instances per video. 
Table \ref{table:6_quantitative_comparison_instance_number} and Fig. \ref{fig:12_qualitative_instance_count} compare the performances of our PRIMEdit and other baseline methods for four different numbers of instance modifications: (i) \texttt{Easy Video} ($\text{MIVE}_E$) with a single and the largest edited instance in the video, (ii) \texttt{Medium Video} ($\text{MIVE}_M$) with 1 to 3 edited instances, (iii) \texttt{Hard Video} ($\text{MIVE}_H$) with 4 to 7 edited instances and (iv) \texttt{eXtreme Video} ($\text{MIVE}_X$) with more than 7 edited instances, as indicated in Sec. \ref{subsec:dataset_construction}.
Since $\text{MIVE}_E$ contains only one instance, the CIA Score is excluded from Table \ref{table:6_quantitative_comparison_instance_number}.

Regarding the Leakage Scores in Table \ref{table:6_quantitative_comparison_instance_number}, the results show no significant deviation from those in Table \ref{table:3_sota_comparison}.
In most cases, our PRIMEdit achieves the best performance, while VideoGrain consistently ranks second.

For the Local Scores in Table \ref{table:6_quantitative_comparison_instance_number}, the results are consistent with those in Table \ref{table:3_sota_comparison}. 
Our PRIMEdit almost always achieves the best LTC, LTF, and IA scores.
While exhibiting comparable LTC performance with the global video editing methods in some scenarios ($\text{MIVE}_M$ and $\text{MIVE}_H$), our PRIMEdit consistently outperforms the other two multi-instance editing methods (GAV and VideoGrain) across all local metrics.

We further observe a general trend where both LTF and IA decrease as the number of instances increases.
This trend is especially evident in the qualitative results presented in Fig. \ref{fig:12_qualitative_instance_count} for the global editing methods ((b)–(f)).
These methods lack explicit local control, often leading to editing failure and confusion about which instance to edit for each instance caption in the global target caption.
The multi-instance VE methods, on the other hand, can better assign captions to instances, with our PRIMEdit showing the best qualitative and quantitative performance.

Similar to the quantitative results for different instance sizes, we can also observe a trade-off between temporal consistency and editing fidelity in multi-instance settings across different numbers of instances.
Our PRIMEdit, however, effectively balances this trade-off, achieving a superior editing fidelity while maintaining competitive temporal consistency.

The Global Scores also remain consistent with the SOTA comparison in Table \ref{table:3_sota_comparison}, where our method demonstrates competitive GTC and achieves the second-best performance in FA ($\text{MIVE}_E$, $\text{MIVE}_M$, and $\text{MIVE}_H$).

Notably, the competitive performance of our PRIMEdit is not solely attributed to its multi-instance editing capability.
In the $\text{MIVE}_E$ scenario, where only a single instance is edited, our PRIMEdit still outperforms baselines specifically optimized for single-instance editing across all local metrics.
This highlights the versatility and robustness of our PRIMEdit in delivering accurate and faithful localized edits, regardless of the number of edited instances.

In summary, although small score deviations are observed across Tabs. \ref{table:3_sota_comparison}, \ref{table:5_quantitative_comparison_instance_size} and \ref{table:6_quantitative_comparison_instance_number}, our PRIMEdit maintains consistent performance across diverse scenarios. 
These results demonstrate the robustness of our method against variations in both instance sizes and the numbers of edited instances.

\subsection{Quantitative Evaluation on Challenging MIVE Subsets}

\begin{table*}[t]
    \small
    \caption{
        Quantitative evaluation on challenging subsets of the MIVE dataset: reflective surfaces, tiny objects (small instances), and extreme instance counts ($\text{MIVE}_{X}$, $>7$ edited instances). 
        FR denotes failure rate, defined as the percentage of instances with IA = 0 (i.e., the edited instance is more similar to its source caption than to its target caption).
    }
    \vspace{-0.4cm}
    \begin{center}
    \setlength\tabcolsep{6pt}{
        \renewcommand{\arraystretch}{1.1}
        \scalebox{0.80}{
          \begin{tabular}{l | c c c | c c | c c c}
            \toprule
            \multirow{2}{*}{Methods} & \multicolumn{3}{c|}{Reflective Surfaces} & \multicolumn{2}{c|}{Tiny Objects} & \multicolumn{3}{c}{$\text{MIVE}_{X}$} \\
            & FR $\downarrow$ & IA $\uparrow$ & CIA (\textbf{Ours}) $\uparrow$ & FR $\downarrow$ & IA $\uparrow$ & FR $\downarrow$ & IA $\uparrow$ & CIA (\textbf{Ours}) $\uparrow$ \\
            \midrule

            ControlVideo \cite{controlvideo2023zhang}
              & 0.5478 & 0.4522 & 0.3589
              & 0.6692 & 0.3308
              & 0.6398 & 0.3602 & 0.1698 \\
            GAV \cite{groundavideo2024jeong}
              & 0.5989 & 0.4011 & 0.3978
              & 0.6642 & 0.3358
              & 0.7089 & 0.2911 & 0.2210 \\
            VideoGrain \cite{videograin2025yang}
              & \textcolor{red}{\textbf{0.2978}} & \textcolor{red}{\textbf{0.7022}} & \textcolor{blue}{\underline{0.4056}}
              & \textcolor{red}{\textbf{0.5581}} & \textcolor{red}{\textbf{0.4419}}
              & \textcolor{blue}{\underline{0.5950}} & \textcolor{blue}{\underline{0.4050}} & \textcolor{blue}{\underline{0.2516}} \\
            \textbf{PRIMEdit (Ours)}
              & \textcolor{blue}{\underline{0.3811}} & \textcolor{blue}{\underline{0.6199}} & \textcolor{red}{\textbf{0.5311}}
              & \textcolor{blue}{\underline{0.6130}} & \textcolor{blue}{\underline{0.3870}}
              & \textcolor{red}{\textbf{0.5791}} & \textcolor{red}{\textbf{0.4209}} & \textcolor{red}{\textbf{0.3413}} \\
              
            \bottomrule
          \end{tabular}
        }
    }
    \end{center}
    \vspace{-0.6cm}
    \label{table:quantitative_mive_subset}
\end{table*}

\noindent
Beyond the analyses on the sizes and numbers of instances in Sec.~\ref{subsec:difficulty_level}, we further quantify performance on particularly challenging subsets of MIVE, namely: (i) videos containing reflective surfaces; (ii) tiny objects (small instances); and (iii) extreme number of instances ($>7$ instances).

For reflective scenes, we identify 5 videos containing visible mirror or glossy reflections where an edited instance appears both directly and via reflection.
For tiny objects, we follow the COCO \cite{coco2014lin} convention in Sec.~\ref{subsec:difficulty_level} and evaluate instances with average area $<32^2$ pixels.
For extreme number of instances, we use the $\text{MIVE}_X$ subset defined in Sec. IV-A, where more than seven instances are edited simultaneously.

Table~\ref{table:quantitative_mive_subset} reports the Failure Rate (FR), defined as the percentage of instances for which the Instance Accuracy (IA) equals 0 (i.e., the edited instance is more similar to its source caption than to its target caption), together with Instance Accuracy (IA) and Cross-Instance Accuracy (CIA) for PRIMEdit and ControlVideo (global editing with strong overall performance), GAV (instance-level editing with bounding boxes), and VideoGrain (instance-level editing with masks).
For the tiny objects, we only show FR and IA since these are the only scores that can be computed based on the instance size.
Qualitative results for the reflective scenes, tiny objects, and extreme number of instances are shown in Fig.~\ref{fig:16_limitations}, Fig~\ref{fig:11_qualitative_instance_size}, and Fig.~\ref{fig:12_qualitative_instance_count}, respectively.

Table~\ref{table:quantitative_mive_subset} reveals distinct failure modes across the three subsets.
For reflective scenes, VideoGrain achieves the lowest FR and highest IA, indicating strong instance-level alignment under visually duplicated structures.
PRIMEdit, however, achieves substantially higher CIA (0.5311 vs. 0.4056), suggesting improved robustness to cross-instance interference when objects appear both directly and via reflection.
For tiny objects, all methods degrade significantly, with failure rates above 0.55 and instance accuracy below 0.45, confirming the difficulty of editing small instances under limited latent resolution.
While VideoGrain slightly outperforms others in FR, PRIMEdit remains competitive, and the narrow performance gaps indicate that tiny object editing remains an open challenge for diffusion-based VE methods. 
In contrast, under extreme instance counts ($\text{MIVE}_{X}$), PRIMEdit demonstrates clear advantages across all metrics, achieving the lowest FR, highest IA, and highest CIA.
The margin in CIA over VideoGrain (0.3413 vs. 0.2516) highlights stronger resilience to attention entanglement as instance density increases.

\subsection{Analysis on DMS}

\begin{figure*}[t]
  \centering
    \includegraphics[width=0.75\linewidth]{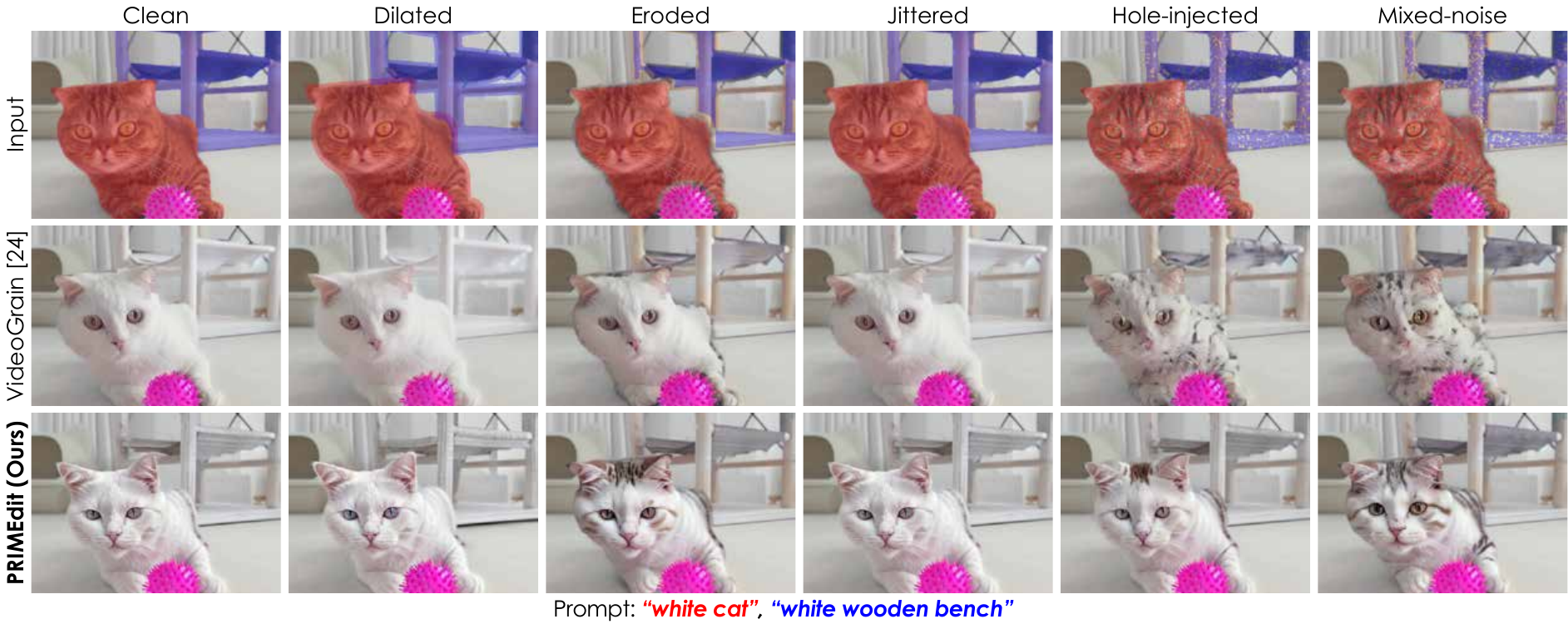}
    \vspace{-0.3cm}
    \caption{
        Editing results of Videograin \cite{videograin2025yang} and PRIMEdit under imprecise instance mask conditions.
    }
   \label{fig:13_dms_analysis}
   \vspace{-0.3cm}
\end{figure*}

\begin{table*}[t]
    \small
    \caption{
        Quantitative comparison between VideoGrain \cite{videograin2025yang} and PRIMEdit under noisy instance masks.
        For each metric, the percentage change relative to the Clean setting is reported in parentheses alongside the raw score.
        Within each column, the numerically largest and second-largest percentage changes are highlighted in \textcolor{red}{\textbf{red}} and \textcolor{blue}{\underline{blue}}, respectively.
    }
    \vspace{-0.4cm}
    \begin{center}
    \setlength\tabcolsep{1.9pt}{
        \renewcommand{\arraystretch}{1.1}
        \scalebox{0.67}{
          \begin{tabular}{l | c | c c c | c c c | c c c | c}
            \toprule
            \multirow{2}{*}{Methods} & \multirow{2}{*}{Mask} & \multicolumn{3}{c}{Global Scores} & \multicolumn{3}{|c|}{Local Scores} & \multicolumn{3}{c|}{Leakage Scores} & Mean \% \\
            & & GTC $\uparrow$ & GTF $\uparrow$ & FA $\uparrow$ & LTC $\uparrow$ & LTF $\uparrow$ & IA $\uparrow$ & CIA (\textbf{Ours}) $\uparrow$ & SSIM $\uparrow$ & LPIPS $\downarrow$ & Change $\uparrow$ \\
            
            \midrule

            VideoGrain \cite{videograin2025yang} & \multirow{2}{*}{Clean}
              & \textcolor{blue}{\underline{0.9600}} 
              & \textcolor{red}{\textbf{0.2803}} 
              & \textcolor{blue}{\underline{0.7430}} 
              & \textcolor{blue}{\underline{0.9423}} 
              & \textcolor{blue}{\underline{0.2026}} 
              & \textcolor{blue}{\underline{0.4798}} 
              & \textcolor{blue}{\underline{0.5868}} 
              & \textcolor{blue}{\underline{0.8929}} 
              & \textcolor{red}{\textbf{0.0488}}
              & - \\
            
            \textbf{PRIMEdit (Ours)} &
              & \textcolor{red}{\textbf{0.9677}} 
              & \textcolor{blue}{\underline{0.2632}} 
              & \textcolor{red}{\textbf{0.7707}} 
              & \textcolor{red}{\textbf{0.9552}} 
              & \textcolor{red}{\textbf{0.2048}} 
              & \textcolor{red}{\textbf{0.5369}} 
              & \textcolor{red}{\textbf{0.6705}} 
              & \textcolor{red}{\textbf{0.9008}} 
              & \textcolor{blue}{\underline{0.0586}}
              & - \\
              
            \midrule

            VideoGrain \cite{videograin2025yang} & \multirow{2}{*}{Dilated}
              & \textcolor{blue}{\underline{0.9604}} {\small \textcolor{blue}{\underline{(+0.05\%)}}}
              & \textcolor{red}{\textbf{0.2848}} {\small \textcolor{red}{\textbf{(+1.61\%)}}}
              & \textcolor{blue}{\underline{0.7732}} {\small \textcolor{red}{\textbf{(+4.07\%)}}}
              & \textcolor{blue}{\underline{0.9429}} {\small \textcolor{blue}{\underline{(+0.06\%)}}}
              & \textcolor{blue}{\underline{0.2046}} {\small \textcolor{red}{\textbf{(+0.95\%)}}}
              & \textcolor{blue}{\underline{0.4969}} {\small \textcolor{red}{\textbf{(+3.55\%)}}}
              & \textcolor{blue}{\underline{0.5912}} {\small \textcolor{red}{\textbf{(+0.74\%)}}}
              & \textcolor{blue}{\underline{0.8879}} {\small \textcolor{red}{\textbf{(-0.56\%)}}}
              & \textcolor{red}{\textbf{0.0565}} {\small \textcolor{blue}{\underline{(+15.87\%)}}}
              & \textcolor{red}{\textbf{+2.93}} \\
            
            \textbf{PRIMEdit (Ours)} &
              & \textcolor{red}{\textbf{0.9691}} {\small \textcolor{red}{\textbf{(+0.14\%)}}}
              & \textcolor{blue}{\underline{0.2659}} {\small \textcolor{blue}{\underline{(+1.06\%)}}}
              & \textcolor{red}{\textbf{0.7912}} {\small \textcolor{blue}{\underline{(+2.65\%)}}}
              & \textcolor{red}{\textbf{0.9571}} {\small \textcolor{red}{\textbf{(+0.20\%)}}}
              & \textcolor{red}{\textbf{0.2064}} {\small \textcolor{blue}{\underline{(+0.79\%)}}}
              & \textcolor{red}{\textbf{0.5513}} {\small \textcolor{blue}{\underline{(+2.68\%)}}}
              & \textcolor{red}{\textbf{0.6703}} {\small \textcolor{blue}{\underline{(-0.03\%)}}}
              & \textcolor{red}{\textbf{0.8936}} {\small \textcolor{blue}{\underline{(-0.80\%)}}}
              & \textcolor{blue}{\underline{0.0698}} {\small \textcolor{red}{\textbf{(+19.23\%)}}}
              & \textcolor{blue}{\underline{+2.88}} \\
              
            \midrule

            VideoGrain \cite{videograin2025yang} & \multirow{2}{*}{Eroded}
              & \textcolor{blue}{\underline{0.9602}} {\small \textcolor{red}{\textbf{(+0.02\%)}}}
              & \textcolor{red}{\textbf{0.2688}} {\small \textcolor{blue}{\underline{(-4.08\%)}}}
              & \textcolor{blue}{\underline{0.5884}} {\small \textcolor{blue}{\underline{(-20.81\%)}}}
              & \textcolor{blue}{\underline{0.9424}} {\small \textcolor{red}{\textbf{(+0.01\%)}}}
              & \textcolor{blue}{\underline{0.1975}} {\small \textcolor{blue}{\underline{(-2.51\%)}}}
              & \textcolor{blue}{\underline{0.4037}} {\small \textcolor{blue}{\underline{(-15.86\%)}}}
              & \textcolor{blue}{\underline{0.5775}} {\small \textcolor{red}{\textbf{(-1.59\%)}}}
              & \textcolor{blue}{\underline{0.8968}} {\small \textcolor{blue}{\underline{(+0.43\%)}}}
              & \textcolor{red}{\textbf{0.0414}} {\small \textcolor{red}{\textbf{(-15.07\%)}}}
              & \textcolor{blue}{\underline{-6.61}} \\
            
            \textbf{PRIMEdit (Ours)} &
              & \textcolor{red}{\textbf{0.9673}} {\small \textcolor{blue}{\underline{(-0.04\%)}}}
              & \textcolor{blue}{\underline{0.2571}} {\small \textcolor{red}{\textbf{(-2.29\%)}}}
              & \textcolor{red}{\textbf{0.7064}} {\small \textcolor{red}{\textbf{(-8.36\%)}}}
              & \textcolor{red}{\textbf{0.9542}} {\small \textcolor{blue}{\underline{(-0.11\%)}}}
              & \textcolor{red}{\textbf{0.2018}} {\small \textcolor{red}{\textbf{(-1.44\%)}}}
              & \textcolor{red}{\textbf{0.4788}} {\small \textcolor{red}{\textbf{(-10.83\%)}}}
              & \textcolor{red}{\textbf{0.6484}} {\small \textcolor{blue}{\underline{(-3.29\%)}}}
              & \textcolor{red}{\textbf{0.9061}} {\small \textcolor{red}{\textbf{(+0.59\%)}}}
              & \textcolor{blue}{\underline{0.0489}} {\small \textcolor{blue}{\underline{(-16.55\%)}}}
              & \textcolor{red}{\textbf{-4.70}} \\
              
            \midrule

            VideoGrain \cite{videograin2025yang} & \multirow{2}{*}{Jittered}
              & \textcolor{blue}{\underline{0.9588}} {\small \textcolor{blue}{\underline{(-0.12\%)}}}
              & \textcolor{red}{\textbf{0.2790}} {\small \textcolor{blue}{\underline{(-0.45\%)}}}
              & \textcolor{blue}{\underline{0.7330}} {\small \textcolor{blue}{\underline{(-1.35\%)}}}
              & \textcolor{blue}{\underline{0.9407}} {\small \textcolor{blue}{\underline{(-0.17\%)}}}
              & \textcolor{blue}{\underline{0.2021}} {\small \textcolor{blue}{\underline{(-0.25\%)}}}
              & \textcolor{blue}{\underline{0.4715}} {\small \textcolor{red}{\textbf{(-1.74\%)}}}
              & \textcolor{blue}{\underline{0.5847}} {\small \textcolor{blue}{\underline{(-0.37\%)}}}
              & \textcolor{blue}{\underline{0.8936}} {\small \textcolor{red}{\textbf{(+0.07\%)}}}
              & \textcolor{red}{\textbf{0.0475}} {\small \textcolor{blue}{\underline{(-2.51\%)}}}
              & \textcolor{blue}{\underline{-0.77}} \\
            
            \textbf{PRIMEdit (Ours)} &
              & \textcolor{red}{\textbf{0.9677}} {\small \textcolor{red}{\textbf{(+0.00\%)}}}
              & \textcolor{blue}{\underline{0.2628}} {\small \textcolor{red}{\textbf{(-0.13\%)}}}
              & \textcolor{red}{\textbf{0.7652}} {\small \textcolor{red}{\textbf{(-0.72\%)}}}
              & \textcolor{red}{\textbf{0.9550}} {\small \textcolor{red}{\textbf{(-0.02\%)}}}
              & \textcolor{red}{\textbf{0.2044}} {\small \textcolor{red}{\textbf{(-0.15\%)}}}
              & \textcolor{red}{\textbf{0.5275}} {\small \textcolor{red}{\textbf{(-1.74\%)}}}
              & \textcolor{red}{\textbf{0.6691}} {\small \textcolor{red}{\textbf{(-0.21\%)}}}
              & \textcolor{red}{\textbf{0.9014}} {\small \textcolor{red}{\textbf{(+0.07\%)}}}
              & \textcolor{blue}{\underline{0.0575}} {\small \textcolor{red}{\textbf{(-1.88\%)}}}
              & \textcolor{red}{\textbf{-0.53}} \\
            
            \midrule

            VideoGrain \cite{videograin2025yang} & Hole-
              & \textcolor{blue}{\underline{0.9511}} {\small \textcolor{blue}{\underline{(-0.93\%)}}}
              & \textcolor{red}{\textbf{0.2661}} {\small \textcolor{blue}{\underline{(-5.06\%)}}}
              & \textcolor{blue}{\underline{0.5961}} {\small \textcolor{blue}{\underline{(-19.77\%)}}}
              & \textcolor{blue}{\underline{0.9318}} {\small \textcolor{red}{\textbf{(-1.11\%)}}}
              & \textcolor{blue}{\underline{0.1950}} {\small \textcolor{blue}{\underline{(-3.75\%)}}}
              & \textcolor{blue}{\underline{0.3828}} {\small \textcolor{blue}{\underline{(-20.21\%)}}}
              & \textcolor{blue}{\underline{0.5667}} {\small \textcolor{blue}{\underline{(-3.43\%)}}}
              & \textcolor{blue}{\underline{0.8967}} {\small \textcolor{red}{\textbf{(+0.43\%)}}}
              & \textcolor{red}{\textbf{0.0426}} {\small \textcolor{blue}{\underline{(-12.64\%)}}}
              & \textcolor{blue}{\underline{-7.39}} \\
            
            \textbf{PRIMEdit (Ours)} & injected
              & \textcolor{red}{\textbf{0.9660}} {\small \textcolor{red}{\textbf{(-0.17\%)}}}
              & \textcolor{blue}{\underline{0.2591}} {\small \textcolor{red}{\textbf{(-1.55\%)}}}
              & \textcolor{red}{\textbf{0.7295}} {\small \textcolor{red}{\textbf{(-5.36\%)}}}
              & \textcolor{red}{\textbf{0.9539}} {\small \textcolor{blue}{\underline{(-0.14\%)}}}
              & \textcolor{red}{\textbf{0.2022}} {\small \textcolor{red}{\textbf{(-1.23\%)}}}
              & \textcolor{red}{\textbf{0.4945}} {\small \textcolor{red}{\textbf{(-7.89\%)}}}
              & \textcolor{red}{\textbf{0.6592}} {\small \textcolor{red}{\textbf{(-1.68\%)}}}
              & \textcolor{red}{\textbf{0.9027}} {\small \textcolor{blue}{\underline{(+0.21\%)}}}
              & \textcolor{blue}{\underline{0.0547}} {\small \textcolor{red}{\textbf{(-6.64\%)}}}
              & \textcolor{red}{\textbf{-2.72}} \\
              
            \midrule

            VideoGrain \cite{videograin2025yang} & Mixed-
              & \textcolor{blue}{\underline{0.9585}} {\small \textcolor{blue}{\underline{(-0.15\%)}}}
              & \textcolor{red}{\textbf{0.2571}} {\small \textcolor{blue}{\underline{(-8.26\%)}}}
              & \textcolor{blue}{\underline{0.4762}} {\small \textcolor{blue}{\underline{(-35.91\%)}}}
              & \textcolor{blue}{\underline{0.9405}} {\small \textcolor{blue}{\underline{(-0.19\%)}}}
              & \textcolor{blue}{\underline{0.1912}} {\small \textcolor{blue}{\underline{(-5.63\%)}}}
              & \textcolor{blue}{\underline{0.3299}} {\small \textcolor{blue}{\underline{(-31.24\%)}}}
              & \textcolor{blue}{\underline{0.5657}} {\small \textcolor{red}{\textbf{(-3.60\%)}}}
              & \textcolor{blue}{\underline{0.8996}} {\small \textcolor{red}{\textbf{(+0.75\%)}}}
              & \textcolor{red}{\textbf{0.0383}} {\small \textcolor{blue}{\underline{(-21.47\%)}}}
              & \textcolor{blue}{\underline{-11.75}} \\
            
            \textbf{PRIMEdit (Ours)} & noise
              & \textcolor{red}{\textbf{0.9663}} {\small \textcolor{red}{\textbf{(-0.14\%)}}}
              & \textcolor{blue}{\underline{0.2524}} {\small \textcolor{red}{\textbf{(-4.09\%)}}}
              & \textcolor{red}{\textbf{0.6488}} {\small \textcolor{red}{\textbf{(-15.82\%)}}}
              & \textcolor{red}{\textbf{0.9534}} {\small \textcolor{red}{\textbf{(-0.19\%)}}}
              & \textcolor{red}{\textbf{0.1991}} {\small \textcolor{red}{\textbf{(-2.77\%)}}}
              & \textcolor{red}{\textbf{0.4372}} {\small \textcolor{red}{\textbf{(-18.56\%)}}}
              & \textcolor{red}{\textbf{0.6429}} {\small \textcolor{blue}{\underline{(-4.11\%)}}}
              & \textcolor{red}{\textbf{0.9071}} {\small \textcolor{blue}{\underline{(+0.70\%)}}}
              & \textcolor{blue}{\underline{0.0470}} {\small \textcolor{red}{\textbf{(-19.68\%)}}}
              & \textcolor{red}{\textbf{-7.18}} \\
              
            \bottomrule
            
          \end{tabular}
        }
    }
    \end{center}
    \vspace{-0.6cm}
    \label{table:7_dms_analysis}
\end{table*}

\noindent
We compare the performance of PRIMEdit against VideoGrain under noisy instance masks.
To systematically evaluate robustness to mask imperfections, we conduct five experiments on the full MIVE dataset using synthetically corrupted masks:\\
\textbf{(i) Dilated masks}, where masks are randomly dilated using kernels with sizes selected from \{5, 7, 9, 11, 13, 15\} pixels;\\
\textbf{(ii) Eroded masks}, where masks are randomly eroded using kernels with sizes selected from \{5, 7, 9, 11, 13, 15\} pixels;\\
\textbf{(iii) Jittered masks}, where boundary perturbations are introduced to simulate imperfect annotations.
Specifically, we compute the Euclidean distance transform of each binary mask and define a narrow boundary band (5 pixels) around the object contour.
Gaussian noise (smoothed with a Gaussian kernel) is added within the boundary band.
The perturbed distance map is thresholded to obtain a binary mask, producing locally irregular contour distortions while preserving overall object structure.
We set the jitter strength to 30;\\
\textbf{(iv) Hole-injected masks}, where we simulate missing annotations by removing irregular interior regions from the foreground.
For each mask, we sample a Gaussian noise, smooth it with a $7\times7$ Gaussian kernel, and threshold it at 0.20 to obtain a binary pattern.
Pixels where both the original mask and the thresholded noise have values equal to 1 are set to 0, producing spatially irregular holes of varying shapes and sizes within the object.\\
\textbf{(v) Mixed-noise masks}, where dilation, erosion, boundary jittering, and hole injection are randomly combined to simulate more severe and realistic segmentation errors.

Table~\ref{table:7_dms_analysis} and Fig.~\ref{fig:13_dms_analysis} present quantitative and qualitative results, respectively.
From Table~\ref{table:7_dms_analysis}, PRIMEdit maintains higher LTC, LTF, and IA scores across all mask perturbation types.
As corruption intensifies from dilated, eroded, and jittered to hole-injected and mixed-noise settings, VideoGrain shows more pronounced drops in IA, particularly under hole-injected and mixed-noise masks.
This reflects increased inconsistency in applying the target attributes within the intended instance.
In contrast, PRIMEdit exhibits smaller IA reductions and shows more stable LTC and LTF values, indicating that local temporal coherence and instance-caption alignment remain intact even when masks are distorted or incomplete.

The qualitative results directly reflect these trends.
Under eroded masks in Fig.~\ref{fig:13_dms_analysis}, VideoGrain leaves portions of the cat partially unedited, producing visible inconsistency across the object boundaries.
This corresponds to the lower IA values observed in the table.
Under hole-injected and mixed-noise masks, patchy editing appears more frequently in VideoGrain outputs, which aligns with the larger IA degradation.
PRIMEdit, however, maintains more uniform editing across the cat and bench, visually confirming its higher IA and stable local metrics.

Regarding leakage sensitivity, the CIA scores provide the clearest distinction.
The differences in CIA are most pronounced under corruption types that introduce irregular or fragmented mask boundaries, such as jittered and hole-injected masks.
In these settings, PRIMEdit maintains a clearer advantage, indicating stronger resistance to cross-instance interference when mask precision is disrupted in non-uniform ways.
Even in corruption types where the relative percentage gaps are smaller (e.g., dilated or eroded masks), PRIMEdit still consistently achieves higher raw CIA scores than VideoGrain.
This pattern suggests that cross-instance leakage is more strongly triggered by irregular mask perturbations than by uniform boundary scaling.

Overall, the combined quantitative (shown in the last column of Table~\ref{table:7_dms_analysis}) and qualitative evidence demonstrates that PRIMEdit is less sensitive to mask inaccuracies (i.e., strong robustness against noisy masks enabled by the complementary interaction of IPR and DMS, where IPR stabilizes instance-level attention while DMS mitigates boundary distortions and missing regions).
It preserves local scores, maintains stronger instance-level alignment, and exhibits better leakage control across different perturbation types, especially under severe corruption such as hole-injected and mixed-noise masks.

\begin{figure}[t]
  \centering
    \includegraphics[width=0.78\linewidth]{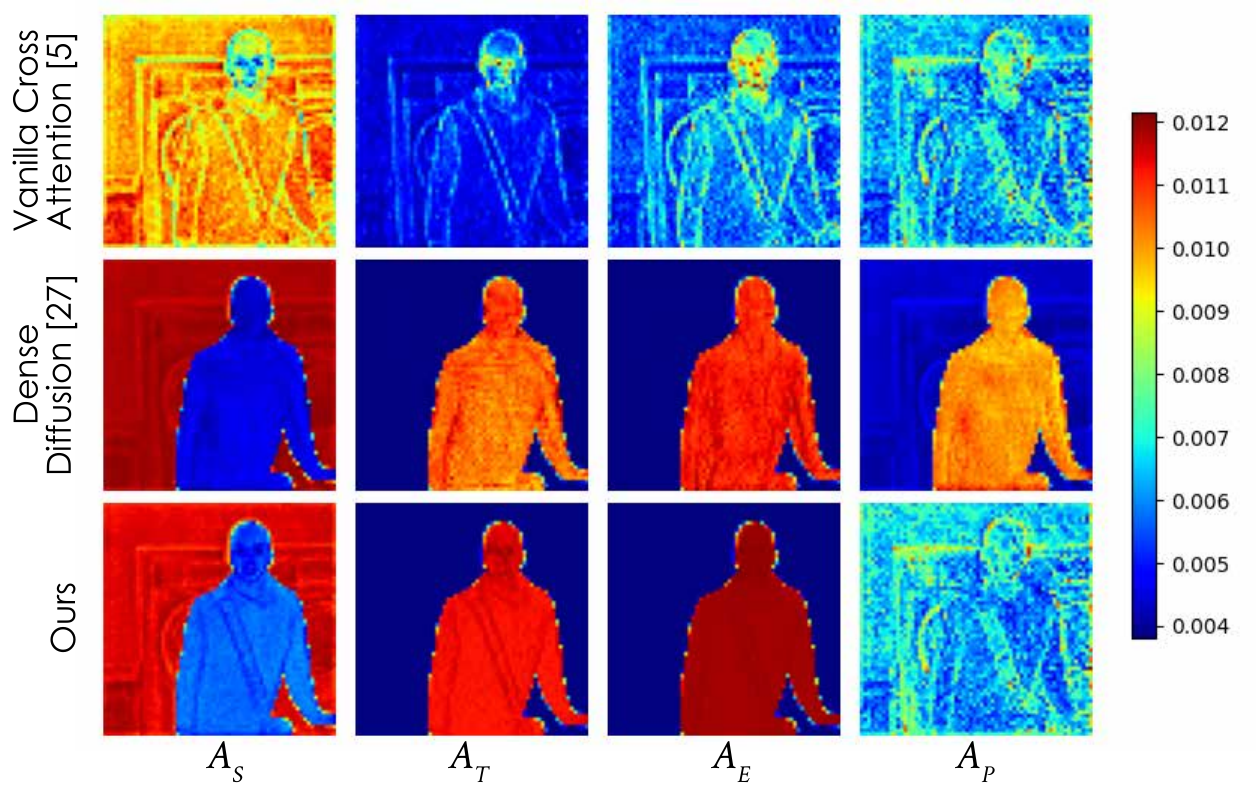}
    \vspace{-0.35cm}
    \caption{
        Attention weight visualization.
    }
   \label{fig:14_ipr_analysis}
   \vspace{-0.6cm}
\end{figure}

\subsection{Analysis on IPR}
\label{subsec:ipr_analysis}

\noindent
Fig. \ref{fig:14_ipr_analysis} presents the attention maps corresponding to the editing results in Fig. \ref{fig:9_ipr_ablation_study}.
Compared to DenseDiffusion \cite{densediffusion2023kim}, our IPR module better mitigates artifacts by smoothly distributing attention from the start (\textit{S}) to the end (\textit{E}) of the text token, whereas DenseDiffusion concentrates attention on the token (\textit{T}), making it more prone to artifacts (Fig. \ref{fig:9_ipr_ablation_study}-(c)). DenseDiffusion often fails to maintain object structures, as seen in Fig. \ref{fig:9_ipr_ablation_study}-(c) where the alien's body merges with its jacket (orange box), and the control buttons on the yellow washing machine lose their structures (green box).
Further, the attention modulation of DenseDiffusion adopted by VideoGrain, is not robust against noisy mask settings, as evident in the results of Fig. \ref{fig:13_dms_analysis}.
In contrast, our PRIMEdit preserves object integrity, as shown in Fig. \ref{fig:9_ipr_ablation_study}-(d), where the alien's jacket remains intact and the yellow washing machine retains its structure.

\begin{table}[t]
  \small
  \caption{
        A user-study on human evaluation for attention leakage.
  }
  \vspace{-0.4cm}
  \begin{center}
      \renewcommand{\arraystretch}{1.1}
      \setlength\tabcolsep{5pt}{
        \scalebox{0.77}{
          \begin{tabular}{l | c | c | c | c}
            \toprule
            \multirow{2}{*}{ \diagbox[]{Metrics}{Methods} } & \multirow{2}{*}{ControlVideo} & \multirow{2}{*}{GAV} & \multirow{2}{*}{VideoGrain} & \multirow{2}{*}{\textbf{PRIMEdit (Ours)}} \\
            & & & \\
            \midrule
            CIA $\uparrow$ & 0.4963 & 0.5516 & \textcolor{blue}{\underline{0.5868}}  & \textcolor{red}{\textbf{0.6705}} \\
            \midrule
            Human Eval. Ranks $\downarrow$ & 4 & 3 & \textcolor{blue}{\underline{2}} & \textcolor{red}{\textbf{1}} \\
            \bottomrule
          \end{tabular}
        }
      }
  \end{center}
  \label{table:8_cia_correlation}
  \vspace{-0.3cm}
\end{table}

\subsection{Correlation of CIA Score with Human Evaluation}

\noindent
We conduct an additional user study with 30 participants to evaluate the effectiveness of the CIA Score in quantifying attention leakage.
Participants ranked the severity of leakage in results from PRIMEdit and three representative baselines: ControlVideo (global editing with strong overall performance), GAV (instance-level editing with bounding boxes), and VideoGrain (instance-level editing with masks).
Table \ref{table:8_cia_correlation} reports the CIA Scores and human rankings for all methods.
The CIA Score rankings align correctly with those from the user study.
We further compute the Spearman rank correlation between CIA Score rankings and user evaluation rankings across all participants, obtaining a score of $\rho=0.86$, indicating strong correlation.
These results suggest that CIA Score effectively captures attention leakage in multi-instance VE and aligns well with human visual perception.

\subsection{Limitations and Discussions}

\begin{table*}[t]
  \small
  \renewcommand{\arraystretch}{1.1}
  \caption{
    Runtime (seconds) and GPU memory usage (GB) comparison for two categories: global video editing (top) and local video editing (bottom). We denote \textit{IS} as the number of inversion steps, \textit{DS} as the number of denoising steps, \textit{I} as the number of instances, and \textit{F} as the number of frames to edit.
  }
  \vspace{-0.4cm}
  \begin{center}
        \setlength\tabcolsep{7pt}{
        \scalebox{0.75}{
          \begin{tabular}{l | c | c | c | c | c | c}
            \toprule
             \multirow{2}{*}{Methods} & \multicolumn{4}{c|}{Time (s) $\downarrow$ / Memory (GB) $\downarrow$} & \multirow{2}{*}{\textit{IS}} &  \multirow{2}{*}{\textit{DS}} \\
             & \textit{I}=1, \textit{F}=15 & \textit{I}=1, \textit{F}=32 & \textit{I}=3, \textit{F}=15 & \textit{I}=3, \textit{F}=32 & & \\
            
            \midrule
            
            ControlVideo \cite{controlnet2023zhang}
              & \textcolor{red}{\textbf{106.80}} / \textcolor{red}{\textbf{7.97}} & \textcolor{blue}{\underline{359.73}} / \textcolor{red}{\textbf{11.35}} & \textcolor{red}{\textbf{106.96}} / \textcolor{red}{\textbf{7.85}} & \textcolor{blue}{\underline{360.11}} / \textcolor{red}{\textbf{11.21}} & 50 & 50 \\
            FLATTEN \cite{flatten2023cong}
              & 162.86 / \textcolor{blue}{\underline{11.07}} & 566.66 / 29.44 & 162.91 / \textcolor{blue}{\underline{11.13}} & 564.98 / 29.53 & 100 & 50 \\
            RAVE \cite{rave2023kara}
              & 206.59 / 12.23 & 409.62 / \textcolor{blue}{\underline{12.83}} & 208.04 / 12.22 & 409.49 / \textcolor{blue}{\underline{12.83}} & 50 & 50 \\
            TokenFlow \cite{tokenflow2023geyer}
              & 238.91 / 11.45 & 471.70 / 14.89 & 239.60 / 11.44 & 473.94 / 14.90 & 500 & 50 \\
            FreSCo \cite{fresco2024yang}
              & \textcolor{blue}{\underline{128.71}} / 14.62 & \textcolor{red}{\textbf{261.43}} / 14.61 & \textcolor{blue}{\underline{130.30}} / 14.24 & \textcolor{red}{\textbf{259.22}} / 14.64 & - & 20 \\
            
            \midrule
            
            GAV \cite{groundavideo2024jeong}
              & \textcolor{blue}{\underline{223.75}} / \textcolor{blue}{\underline{16.00}} & \textcolor{blue}{\underline{913.29}} / \textcolor{blue}{\underline{29.84}} & \textcolor{blue}{\underline{223.82}} / \textcolor{blue}{\underline{16.00}} & \textcolor{blue}{\underline{913.16}} / \textcolor{blue}{\underline{29.84}} & 20 & 50 \\
            VideoGrain \cite{videograin2025yang}
              & 299.70 / 22.89 & 1130.32 / 46.06 & 296.39 / 22.86 & 1129.76 / 46.06 & 50 & 50 \\
            \textbf{PRIMEdit (Ours)}
              & \textcolor{red}{\textbf{135.27}} / \textcolor{red}{\textbf{15.28}} & \textcolor{red}{\textbf{277.79}} / \textcolor{red}{\textbf{24.27}} & \textcolor{red}{\textbf{216.11}} / \textcolor{red}{\textbf{15.31}} & \textcolor{red}{\textbf{444.47}} / \textcolor{red}{\textbf{24.30}} & 100 & 50 \\
              
            \bottomrule
            
          \end{tabular}
        }
      }
  \end{center}
  \label{table:10_runtime_memory_comparison}
  \vspace{-0.7cm}
\end{table*}

\begin{figure}[t]
  \centering
    \includegraphics[width=0.95\linewidth]{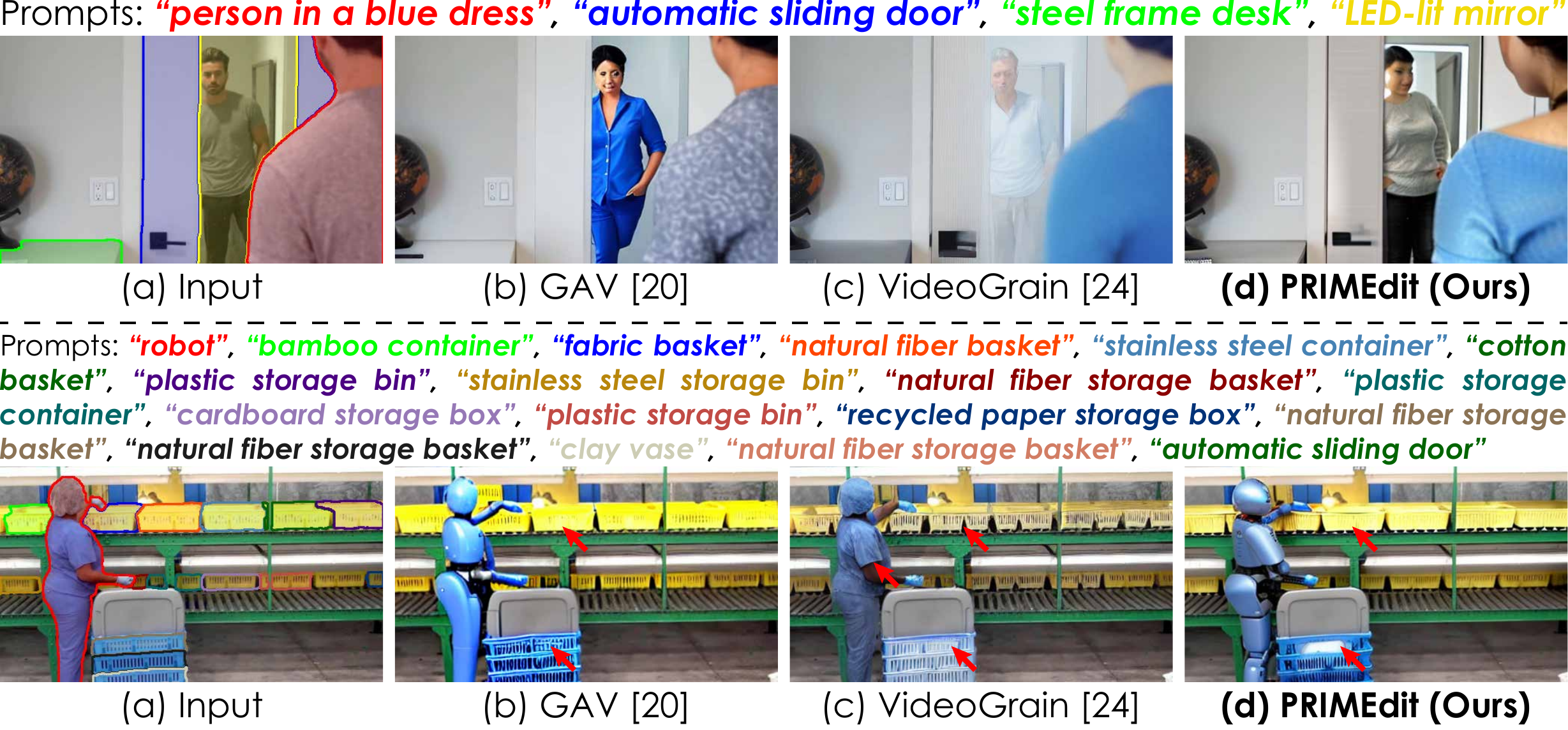}
    \vspace{-0.3cm}
    \caption{
        Limitations of our PRIMEdit and other SOTA multi-instance VE methods in editing scenes with reflective surfaces (top) and an excessive number of instances (bottom).
    }
   \label{fig:16_limitations}
   \vspace{-0.5cm}
\end{figure}

\noindent
Fig. \ref{fig:16_limitations} shows the limitations of our PRIMEdit and other SOTA methods in editing scenes with reflective surfaces and an excessive number of instances. 

As shown in the top row of Fig. \ref{fig:16_limitations}-(d), PRIMEdit struggles in scenes with reflective surfaces, where it cannot edit reflections because the masks cover only the instance itself.
This leads to inconsistencies between the edited instance and its reflection.
As shown in (b) and (c) in the top part of Fig. \ref{fig:16_limitations}, other SOTA methods exhibit the same limitation.
To address this, future works may incorporate a prior technique \cite{reflectingreality2024dhiman} to enforce the consistency between instances and their reflections.

Moreover, PRIMEdit and the other baselines struggle to handle complex scenes containing small objects to be edited.
As shown, they fail in scenes with many instances (18 objects in the bottom scene of Fig. \ref{fig:16_limitations}).
This limitation arises from the inherent difficulty of U-Net–based LDMs in handling small objects, as the input is compressed into low-resolution latents by the VAE encoder and further downsampled within the U-Net.
As a result, very small objects and their masks may vanish within the low-resolution U-Net features, hindering accurate editing.
Editing in the pixel space or a higher-resolution latent space could mitigate this issue at the cost of increased computational cost.
Both the reflection consistency and the small-sized object editing in complex scenes remain open problems for future research in multi-instance VE.

\subsection{Runtime and GPU Memory Usage Comparison}


\noindent
Table~\ref{table:10_runtime_memory_comparison} reports the runtime and GPU memory usage of PRIMEdit and other baselines on a single RTX A6000 GPU.
We evaluate four conditions:
(i) one instance with 15 frames,
(ii) one instance with 32 frames,
(iii) three instances with 15 frames, and
(iv) three instances with 32 frames.
These settings isolate two key scaling factors in multi-instance VE: the number of instances and the number of frames, while keeping the denoising steps fixed.

The baselines are grouped into global (top) and local (bottom) editing methods.
As shown in Table~\ref{table:10_runtime_memory_comparison}, larger numbers of instances or longer sequences lead to increased computational and memory demands for our PRIMEdit, which is expected for diffusion-based video generation models.
However, our PRIMEdit still runs faster and uses less memory, compared to other multi-instance VE baselines.

Compared to the five global editing baselines, our PRIMEdit achieves competitive results.
It should be noted, however, that the global video editing methods are incapable of localized multi-instance VE, as evidenced by their subpar performance (Local Scores and Leakage Scores) in Table \ref{table:3_sota_comparison}.

\section{Conclusion}
\label{sec:conclusion}

\noindent
In this paper, we present PRIMEdit, a novel multi-instance video editing framework, featuring Instance-centric Probability Redistribution (IPR) and Disentangled Multi-instance Sampling (DMS), to elaborately handle instance-level video editing.
PRIMEdit produces faithful, disentangled edits with minimal attention leakage, outperforming existing SOTA methods in qualitative, quantitative, and user studies on our \textit{newly} proposed MIVE dataset, the \textit{first} large-scale dataset dedicated to multi-instance video editing tasks.
We further propose a new Cross-Instance Accuracy (CIA) Score to quantify the attention leakage from one instance to others, which often occurs in multi-instance video editing but has been overlooked by existing video editing metrics.
Our user study supports the effectiveness of the CIA Score, showing the quality consistency with human's visual perception.

\section*{Acknowledgments}
This work was supported by Institute of Information and communications Technology Planning and Evaluation (IITP) grant funded by the Korean government [Ministry of Science and ICT (Information and Communications Technology)] (Project Number: RS2022-00144444, Project Title: Deep Learning Based Visual Representational Learning and Rendering of Static and Dynamic Scenes).

\bibliographystyle{IEEEtran}
\bibliography{main.bib}

@String(CVPR= {IEEE Conf. Comput. Vis. Pattern Recog.})

@String(ICCV= {Int. Conf. Comput. Vis.})

@String(ECCV= {Eur. Conf. Comput. Vis.})

@String(TIP  = {IEEE Trans. Image Process.})

@String(ICLR = {Int. Conf. Learn. Represent.})

@String(CVPR  = {CVPR})

@String(ICCV  = {ICCV})

@String(ECCV  = {ECCV})

@String(TIP   = {IEEE TIP})

@String(ICLR  = {ICLR})

@inproceedings{groundavideo2024jeong,
    title={Ground-A-Video: Zero-shot Grounded Video Editing using Text-to-image Diffusion Models},
    author={Hyeonho Jeong and Jong Chul Ye},
    booktitle={ICLR},
    year={2024},
    url={https://openreview.net/forum?id=28L2FCtMWq}
}

@inproceedings{instancediffusion2024wang,
  title={Instancediffusion: Instance-level control for image generation},
  author={Wang, Xudong and Darrell, Trevor and Rambhatla, Sai Saketh and Girdhar, Rohit and Misra, Ishan},
  booktitle={CVPR},
  pages={6232--6242},
  year={2024}
}

@inproceedings{gligen2023li,
  title={Gligen: Open-set grounded text-to-image generation},
  author={Li, Yuheng and Liu, Haotian and Wu, Qingyang and Mu, Fangzhou and Yang, Jianwei and Gao, Jianfeng and Li, Chunyuan and Lee, Yong Jae},
  booktitle={CVPR},
  pages={22511--22521},
  year={2023}
}

@inproceedings{flatten2023cong,
    title={{FLATTEN}: optical {FL}ow-guided {ATTEN}tion for consistent text-to-video editing},
    author={Yuren Cong and Mengmeng Xu and christian simon and Shoufa Chen and Jiawei Ren and Yanping Xie and Juan-Manuel Perez-Rua and Bodo Rosenhahn and Tao Xiang and Sen He},
    booktitle={ICLR},
    year={2024},
    url={https://openreview.net/forum?id=JgqftqZQZ7}
}

@inproceedings{fresco2024yang,
  title={FRESCO: Spatial-Temporal Correspondence for Zero-Shot Video Translation},
  author={Yang, Shuai and Zhou, Yifan and Liu, Ziwei and Loy, Chen Change},
  booktitle={CVPR},
  pages={8703--8712},
  year={2024}
}

@article{cove2024wang,
    title={COVE: Unleashing the Diffusion Feature Correspondence for Consistent Video Editing},
    author={Wang, Jiangshan and Ma, Yue and Guo, Jiayi and Xiao, Yicheng and Huang, Gao and Li, Xiu},
    journal={arXiv preprint arXiv:2406.08850},
    year={2024}
}

@inproceedings{tuneavideo2023wu,
  title={Tune-a-video: One-shot tuning of image diffusion models for text-to-video generation},
  author={Wu, Jay Zhangjie and Ge, Yixiao and Wang, Xintao and Lei, Stan Weixian and Gu, Yuchao and Shi, Yufei and Hsu, Wynne and Shan, Ying and Qie, Xiaohu and Shou, Mike Zheng},
  booktitle={ICCV},
  pages={7623--7633},
  year={2023}
}

@inproceedings{densediffusion2023kim,
  title={Dense text-to-image generation with attention modulation},
  author={Kim, Yunji and Lee, Jiyoung and Kim, Jin-Hwa and Ha, Jung-Woo and Zhu, Jun-Yan},
  booktitle={ICCV},
  pages={7701--7711},
  year={2023}
}

@inproceedings{noisecollage2024shirakawa,
  title={NoiseCollage: A Layout-Aware Text-to-Image Diffusion Model Based on Noise Cropping and Merging},
  author={Shirakawa, Takahiro and Uchida, Seiichi},
  booktitle={CVPR},
  pages={8921--8930},
  year={2024} 
}

@inproceedings{controlvideo2023zhang,
    title={ControlVideo: Training-free Controllable Text-to-video Generation},
    author={Yabo Zhang and Yuxiang Wei and Dongsheng Jiang and Xiaopeng Zhang and Wangmeng Zuo and Qi Tian},
    booktitle={ICLR},
    year={2024},
    url={https://openreview.net/forum?id=5a79AqFr0c}
}

@inproceedings{rave2023kara,
  title={Rave: Randomized noise shuffling for fast and consistent video editing with diffusion models},
  author={Kara, Ozgur and Kurtkaya, Bariscan and Yesiltepe, Hidir and Rehg, James M and Yanardag, Pinar},
  booktitle={CVPR},
  pages={6507--6516},
  year={2024}
}

@inproceedings{tokenflow2023geyer,
    title={TokenFlow: Consistent Diffusion Features for Consistent Video Editing},
    author={Michal Geyer and Omer Bar-Tal and Shai Bagon and Tali Dekel},
    booktitle={ICLR},
    year={2024},
    url={https://openreview.net/forum?id=lKK50q2MtV}
}

@inproceedings{ldm2022rombach,
  title={High-resolution image synthesis with latent diffusion models},
  author={Rombach, Robin and Blattmann, Andreas and Lorenz, Dominik and Esser, Patrick and Ommer, Bj{\"o}rn},
  booktitle={CVPR},
  pages={10684--10695},
  year={2022}

}

@inproceedings{controlnet2023zhang,
  title={Adding conditional control to text-to-image diffusion models},
  author={Zhang, Lvmin and Rao, Anyi and Agrawala, Maneesh},
  booktitle={ICCV},
  pages={3836--3847},
  year={2023}
}

@inproceedings{clip2021radford,
  title={Learning transferable visual models from natural language supervision},
  author={Radford, Alec and Kim, Jong Wook and Hallacy, Chris and Ramesh, Aditya and Goh, Gabriel and Agarwal, Sandhini and Sastry, Girish and Askell, Amanda and Mishkin, Pamela and Clark, Jack and others},
  booktitle={ICML},
  pages={8748--8763},
  year={2021},
  organization={PMLR}
}

@inproceedings{fatezero2023qi,
  title={Fatezero: Fusing attentions for zero-shot text-based video editing},
  author={Qi, Chenyang and Cun, Xiaodong and Zhang, Yong and Lei, Chenyang and Wang, Xintao and Shan, Ying and Chen, Qifeng},
  booktitle={ICCV},
  pages={15932--15942},
  year={2023}
}

@inproceedings{stablevideo2023chai,
  title={Stablevideo: Text-driven consistency-aware diffusion video editing},
  author={Chai, Wenhao and Guo, Xun and Wang, Gaoang and Lu, Yan},
  booktitle={ICCV},
  pages={23040--23050},
  year={2023}
}

@inproceedings{text2live2022bar,
  title={Text2live: Text-driven layered image and video editing},
  author={Bar-Tal, Omer and Ofri-Amar, Dolev and Fridman, Rafail and Kasten, Yoni and Dekel, Tali},
  booktitle={ECCV},
  pages={707--723},
  year={2022},
  organization={Springer}
}

@inproceedings{avid2024zhang,
  title={AVID: Any-Length Video Inpainting with Diffusion Model},
  author={Zhang, Zhixing and Wu, Bichen and Wang, Xiaoyan and Luo, Yaqiao and Zhang, Luxin and Zhao, Yinan and Vajda, Peter and Metaxas, Dimitris and Yu, Licheng},
  booktitle={CVPR},
  pages={7162--7172},
  year={2024}
}

@inproceedings{rise2014vandersmissen,
  title={The rise of mobile and social short-form video: an in-depth measurement study of vine},
  author={Vandersmissen, Baptist and Godin, Fr{\'e}deric and Tomar, Abhineshwar and De Neve, Wesley and Van de Walle, Rik},
  booktitle={Workshop on Social Multimedia and Storytelling (SoMuS 2014)},
  volume={1198},
  pages={1--10},
  year={2014}
}

@inproceedings{magvit2023yu,
  title={Magvit: Masked generative video transformer},
  author={Yu, Lijun and Cheng, Yong and Sohn, Kihyuk and Lezama, Jos{\'e} and Zhang, Han and Chang, Huiwen and Hauptmann, Alexander G and Yang, Ming-Hsuan and Hao, Yuan and Essa, Irfan and others},
  booktitle={CVPR},
  pages={10459--10469},
  year={2023}
}

@inproceedings{align2023blattmann,
  title={Align your latents: High-resolution video synthesis with latent diffusion models},
  author={Blattmann, Andreas and Rombach, Robin and Ling, Huan and Dockhorn, Tim and Kim, Seung Wook and Fidler, Sanja and Kreis, Karsten},
  booktitle={CVPR},
  pages={22563--22575},
  year={2023}
}

@inproceedings{unlocking2024kim,
  title={Unlocking Creator-AI Synergy: Challenges, Requirements, and Design Opportunities in AI-Powered Short-Form Video Production},
  author={Kim, Jini and Kim, Hajun},
  booktitle={Proceedings of the CHI Conference on Human Factors in Computing Systems},
  pages={1--23},
  year={2024}
}

@article{short2023liu,
  title={Short-form videos for public library marketing: performance analytics of Douyin in China},
  author={Liu, Ying and Chiu, Dickson KW and Ho, Kevin KW},
  journal={Applied Sciences},
  volume={13},
  number={6},
  pages={3386},
  year={2023},
  publisher={MDPI}
}

@inproceedings{automation2021soe,
  title={Automation in Video Editing: Assisted Workflows in Video Editing.},
  author={Soe, Than Htut},
  booktitle={AutomationXP@ CHI},
  year={2021}
}

@inproceedings{videop2p2024liu,
  title={Video-p2p: Video editing with cross-attention control},
  author={Liu, Shaoteng and Zhang, Yuechen and Li, Wenbo and Lin, Zhe and Jia, Jiaya},
  booktitle={CVPR},
  pages={8599--8608},
  year={2024}
}

@inproceedings{consistent2023cheng,
    title={Consistent Video-to-Video Transfer Using Synthetic Dataset},
    author={Jiaxin Cheng and Tianjun Xiao and Tong He},
    booktitle={ICLR},
    year={2024},
    url={https://openreview.net/forum?id=IoKRezZMxF}
}

@article{dynamic2023yang,
  title={Dynamic prompt learning: Addressing cross-attention leakage for text-based image editing},
  author={Yang, Fei and Yang, Shiqi and Butt, Muhammad Atif and van de Weijer, Joost and others},
  journal={NeurIPS},
  volume={36},
  pages={26291--26303},
  year={2023}
}

@article{ddpm2020ho,
  title={Denoising diffusion probabilistic models},
  author={Ho, Jonathan and Jain, Ajay and Abbeel, Pieter},
  journal={NeurIPS},
  volume={33},
  pages={6840--6851},
  year={2020}
}

@inproceedings{ddim2020song,
    title={Denoising Diffusion Implicit Models},
    author={Jiaming Song and Chenlin Meng and Stefano Ermon},
    booktitle={ICLR},
    year={2021},
    url={https://openreview.net/forum?id=St1giarCHLP}
}

@inproceedings{score2020song,
    title={Score-Based Generative Modeling through Stochastic Differential Equations},
    author={Yang Song and Jascha Sohl-Dickstein and Diederik P Kingma and Abhishek Kumar and Stefano Ermon and Ben Poole},
    booktitle={ICLR},
    year={2021},
    url={https://openreview.net/forum?id=PxTIG12RRHS}
}

@article{dalle2022ramesh,
  title={Hierarchical text-conditional image generation with clip latents},
  author={Ramesh, Aditya and Dhariwal, Prafulla and Nichol, Alex and Chu, Casey and Chen, Mark},
  journal={arXiv preprint arXiv:2204.06125},
  volume={1},
  number={2},
  pages={3},
  year={2022}
}

@inproceedings{glide2021nichol,
  title={{GLIDE}: Towards Photorealistic Image Generation and Editing with Text-Guided Diffusion Models},
  author={Nichol, Alexander Quinn and Dhariwal, Prafulla and Ramesh, Aditya and Shyam, Pranav and Mishkin, Pamela and Mcgrew, Bob and Sutskever, Ilya and Chen, Mark},
  booktitle={ICML},
  pages={16784--16804},
  year={2022},
}

@article{imagen2022saharia,
  title={Photorealistic text-to-image diffusion models with deep language understanding},
  author={Saharia, Chitwan and Chan, William and Saxena, Saurabh and Li, Lala and Whang, Jay and Denton, Emily L and Ghasemipour, Kamyar and Gontijo Lopes, Raphael and Karagol Ayan, Burcu and Salimans, Tim and others},
  journal={NeurIPS},
  volume={35},
  pages={36479--36494},
  year={2022}
}

@article{diffusion2021dhariwal,
  title={Diffusion models beat gans on image synthesis},
  author={Dhariwal, Prafulla and Nichol, Alexander},
  journal={NeurIPS},
  volume={34},
  pages={8780--8794},
  year={2021}
}

@inproceedings{videoprobabilisticdm2023yu,
  title={Video probabilistic diffusion models in projected latent space},
  author={Yu, Sihyun and Sohn, Kihyuk and Kim, Subin and Shin, Jinwoo},
  booktitle={CVPR},
  pages={18456--18466},
  year={2023}
}

@article{lavie2023wang,
  title={Lavie: High-quality video generation with cascaded latent diffusion models},
  author={Wang, Yaohui and Chen, Xinyuan and Ma, Xin and Zhou, Shangchen and Huang, Ziqi and Wang, Yi and Yang, Ceyuan and He, Yinan and Yu, Jiashuo and Yang, Peiqing and others},
  journal={arXiv preprint arXiv:2309.15103},
  year={2023}
}

@article{imagenvideo2022ho,
  title={Imagen video: High definition video generation with diffusion models},
  author={Ho, Jonathan and Chan, William and Saharia, Chitwan and Whang, Jay and Gao, Ruiqi and Gritsenko, Alexey and Kingma, Diederik P and Poole, Ben and Norouzi, Mohammad and Fleet, David J and others},
  journal={arXiv preprint arXiv:2210.02303},
  year={2022}
}

@article{latent2022he,
  title={Latent video diffusion models for high-fidelity long video generation},
  author={He, Yingqing and Yang, Tianyu and Zhang, Yong and Shan, Ying and Chen, Qifeng},
  journal={arXiv preprint arXiv:2211.13221},
  year={2022}
}

@inproceedings{structure2023esser,
  title={Structure and content-guided video synthesis with diffusion models},
  author={Esser, Patrick and Chiu, Johnathan and Atighehchian, Parmida and Granskog, Jonathan and Germanidis, Anastasis},
  booktitle={ICCV},
  pages={7346--7356},
  year={2023}
}

@article{videocomposer2024wang,
  title={Videocomposer: Compositional video synthesis with motion controllability},
  author={Wang, Xiang and Yuan, Hangjie and Zhang, Shiwei and Chen, Dayou and Wang, Jiuniu and Zhang, Yingya and Shen, Yujun and Zhao, Deli and Zhou, Jingren},
  journal={NeurIPS},
  volume={36},
  year={2024}
}

@inproceedings{pix2video2023ceylan,
  title={Pix2video: Video editing using image diffusion},
  author={Ceylan, Duygu and Huang, Chun-Hao P and Mitra, Niloy J},
  booktitle={ICCV},
  pages={23206--23217},
  year={2023}
}

@inproceedings{multidiffusion2023bar,
  title={{M}ulti{D}iffusion: Fusing Diffusion Paths for Controlled Image Generation},
  author={Bar-Tal, Omer and Yariv, Lior and Lipman, Yaron and Dekel, Tali},
  booktitle={ICML},
  pages={1737--1752},
  year={2023},
}

@inproceedings{reco2023yang,
  title={Reco: Region-controlled text-to-image generation},
  author={Yang, Zhengyuan and Wang, Jianfeng and Gan, Zhe and Li, Linjie and Lin, Kevin and Wu, Chenfei and Duan, Nan and Liu, Zicheng and Liu, Ce and Zeng, Michael and others},
  booktitle={CVPR},
  pages={14246--14255},
  year={2023}
}

@article{attend2023chefer,
  title={Attend-and-excite: Attention-based semantic guidance for text-to-image diffusion models},
  author={Chefer, Hila and Alaluf, Yuval and Vinker, Yael and Wolf, Lior and Cohen-Or, Daniel},
  journal={ACM Trans. Graph.},
  volume={42},
  number={4},
  pages={1--10},
  year={2023},
  publisher={ACM New York, NY, USA}
}

@inproceedings{training2024chen,
  title={Training-free layout control with cross-attention guidance},
  author={Chen, Minghao and Laina, Iro and Vedaldi, Andrea},
  booktitle={WACV},
  pages={5343--5353},
  year={2024}
}

@inproceedings{boxdiff2023xie,
  title={Boxdiff: Text-to-image synthesis with training-free box-constrained diffusion},
  author={Xie, Jinheng and Li, Yuexiang and Huang, Yawen and Liu, Haozhe and Zhang, Wentian and Zheng, Yefeng and Shou, Mike Zheng},
  booktitle={ICCV},
  pages={7452--7461},
  year={2023}
}

@inproceedings{spatext2023avrahami,
   title={SpaText: Spatio-Textual Representation for Controllable Image Generation},
   url={http://dx.doi.org/10.1109/CVPR52729.2023.01762},
   DOI={10.1109/cvpr52729.2023.01762},
   booktitle={CVPR},
   publisher={IEEE},
   author={Avrahami, Omri and Hayes, Thomas and Gafni, Oran and Gupta, Sonal and Taigman, Yaniv and Parikh, Devi and Lischinski, Dani and Fried, Ohad and Yin, Xi},
   year={2023},
   month=jun
}

@article{midas2020ranftl,
  title={Towards robust monocular depth estimation: Mixing datasets for zero-shot cross-dataset transfer},
  author={Ranftl, Ren{\'e} and Lasinger, Katrin and Hafner, David and Schindler, Konrad and Koltun, Vladlen},
  journal={IEEE TPAMI},
  volume={44},
  number={3},
  pages={1623--1637},
  year={2020},
  publisher={IEEE}
}

@article{cfg2022ho,
  title={Classifier-free diffusion guidance},
  author={Ho, Jonathan and Salimans, Tim},
  journal={arXiv preprint arXiv:2207.12598},
  year={2022}
}

@article{loveutgve2023wu,
  title={Cvpr 2023 text guided video editing competition},
  author={Wu, Jay Zhangjie and Li, Xiuyu and Gao, Difei and Dong, Zhen and Bai, Jinbin and Singh, Aishani and Xiang, Xiaoyu and Li, Youzeng and Huang, Zuwei and Sun, Yuanxi and others},
  journal={arXiv preprint arXiv:2310.16003},
  year={2023}
}

@article{davis2017pont,
  title={The 2017 davis challenge on video object segmentation},
  author={Pont-Tuset, Jordi and Perazzi, Federico and Caelles, Sergi and Arbel{\'a}ez, Pablo and Sorkine-Hornung, Alex and Van Gool, Luc},
  journal={arXiv preprint arXiv:1704.00675},
  year={2017}
}

@inproceedings{vipseg2022miao,
  title={Large-scale video panoptic segmentation in the wild: A benchmark},
  author={Miao, Jiaxu and Wang, Xiaohan and Wu, Yu and Li, Wei and Zhang, Xu and Wei, Yunchao and Yang, Yi},
  booktitle={CVPR},
  pages={21033--21043},
  year={2022}
}

@inproceedings{llava1-5_2024liu,
  title={Improved baselines with visual instruction tuning},
  author={Liu, Haotian and Li, Chunyuan and Li, Yuheng and Lee, Yong Jae},
  booktitle={CVPR},
  pages={26296--26306},
  year={2024}
}

@article{llama2024dubey,
  title={The llama 3 herd of models},
  author={Dubey, Abhimanyu and Jauhri, Abhinav and Pandey, Abhinav and Kadian, Abhishek and Al-Dahle, Ahmad and Letman, Aiesha and Mathur, Akhil and Schelten, Alan and Yang, Amy and Fan, Angela and others},
  journal={arXiv preprint arXiv:2407.21783},
  year={2024}
}

@article{deepcens2023yuksel,
  title={DeepCens: A deep learning-based system for real-time image and video censorship},
  author={Yuksel, Asim Sinan and Tan, Fatma Gulsah},
  journal={Expert Systems},
  volume={40},
  number={10},
  pages={e13436},
  year={2023},
  publisher={Wiley Online Library}
}

@inproceedings{eve2024singer,
  title={Video editing via factorized diffusion distillation},
  author={Singer, Uriel and Zohar, Amit and Kirstain, Yuval and Sheynin, Shelly and Polyak, Adam and Parikh, Devi and Taigman, Yaniv},
  booktitle={ECCV},
  pages={450--466},
  year={2024},
  organization={Springer}
}

@article{modelscope2023wang,
  title={Modelscope text-to-video technical report},
  author={Wang, Jiuniu and Yuan, Hangjie and Chen, Dayou and Zhang, Yingya and Wang, Xiang and Zhang, Shiwei},
  journal={arXiv preprint arXiv:2308.06571},
  year={2023}
}

@article{p2p2022hertz,
  title={Prompt-to-prompt image editing with cross attention control},
  author={Hertz, Amir and Mokady, Ron and Tenenbaum, Jay and Aberman, Kfir and Pritch, Yael and Cohen-Or, Daniel},
  journal={arXiv preprint arXiv:2208.01626},
  year={2022}
}

@inproceedings{tifa2023hu,
  title={Tifa: Accurate and interpretable text-to-image faithfulness evaluation with question answering},
  author={Hu, Yushi and Liu, Benlin and Kasai, Jungo and Wang, Yizhong and Ostendorf, Mari and Krishna, Ranjay and Smith, Noah A},
  booktitle={ICCV},
  pages={20406--20417},
  year={2023} 
}

@inproceedings{crepe2024ma,
  title={Crepe: Can vision-language foundation models reason compositionally?},
  author={Ma, Zixian and Hong, Jerry and Gul, Mustafa Omer and Gandhi, Mona and Gao, Irena and Krishna, Ranjay},
  booktitle={CVPR},
  pages={10910--10921},
  year={2023}
}

@article{reflectingreality2024dhiman,
  title={Reflecting Reality: Enabling Diffusion Models to Produce Faithful Mirror Reflections},
  author={Dhiman, Ankit and Shah, Manan and Parihar, Rishubh and Bhalgat, Yash and Boregowda, Lokesh R and Babu, R Venkatesh},
  journal={arXiv preprint arXiv:2409.14677},
  year={2024}

}

@inproceedings{coco2014lin,
  title={Microsoft coco: Common objects in context},
  author={Lin, Tsung-Yi and Maire, Michael and Belongie, Serge and Hays, James and Perona, Pietro and Ramanan, Deva and Doll{\'a}r, Piotr and Zitnick, C Lawrence},
  booktitle={ECCV},
  pages={740--755},
  year={2014},
  organization={Springer}
}

@inproceedings{videograin2025yang,
  title={VideoGrain: Modulating Space-Time Attention for Multi-grained Video Editing}, 
  author={Xiangpeng Yang and Linchao Zhu and Hehe Fan and Yi Yang},
  year={2025},
  booktitle={ICLR}
}

@article{animatediff2023guo,
  title={AnimateDiff: Animate Your Personalized Text-to-Image Diffusion Models without Specific Tuning},
  author={Guo, Yuwei and Yang, Ceyuan and Rao, Anyi and Liang, Zhengyang and Wang, Yaohui and Qiao, Yu and Agrawala, Maneesh and Lin, Dahua and Dai, Bo},
  journal={ICLR},
  year={2024}
}

@inproceedings{preservecorrelationnoiseprior2024ge,
  title={Preserve Your Own Correlation: A Noise Prior for Video Diffusion Models}, 
  author={Songwei Ge and Seungjun Nah and Guilin Liu and Tyler Poon and Andrew Tao and Bryan Catanzaro and David Jacobs and Jia-Bin Huang and Ming-Yu Liu and Yogesh Balaji},
  year={2023},
  booktitle={ICCV}
}

@inproceedings{dmt2023yatim,
  title={Space-Time Diffusion Features for Zero-Shot Text-Driven Motion Transfer}, 
  author={Danah Yatim and Rafail Fridman and Omer Bar-Tal and Yoni Kasten and Tali Dekel},
  year={2024},
  booktitle={CVPR}
}

@misc{zeroscope2023cerspense,
  title={\href{https://huggingface.co/cerspense/zeroscope_v2_576w}{\texttt{https://huggingface.co/cerspense /zeroscope\_v2\_576w}}},
  author={cerspense},
  year={2023},
  url={https://huggingface.co/cerspense/zeroscope_v2_576w},
}

@inproceedings{gentron2024chen,
      title={GenTron: Diffusion Transformers for Image and Video Generation}, 
      author={Shoufa Chen and Mengmeng Xu and Jiawei Ren and Yuren Cong and Sen He and Yanping Xie and Animesh Sinha and Ping Luo and Tao Xiang and Juan-Manuel Perez-Rua},
      year={2024},
      booktitle={CVPR}
}

@inproceedings{slicedit2024cohen,
      title={Slicedit: Zero-Shot Video Editing With Text-to-Image Diffusion Models Using Spatio-Temporal Slices}, 
      author={Nathaniel Cohen and Vladimir Kulikov and Matan Kleiner and Inbar Huberman-Spiegelglas and Tomer Michaeli},
      year={2024},
      booktitle={ICML}
}

@inproceedings{dreammotion2024jeong,
      title={DreamMotion: Space-Time Self-Similar Score Distillation for Zero-Shot Video Editing}, 
      author={Hyeonho Jeong and Jinho Chang and Geon Yeong Park and Jong Chul Ye},
      year={2024},
      booktitle={ECCV}
}

@article{ssim,
    author={Zhou Wang and Bovik, A.C. and Sheikh, H.R. and Simoncelli, E.P.},
    journal={IEEE TIP}, 
    title={Image quality assessment: from error visibility to structural similarity}, 
    year={2004},
    volume={13},
    number={4},
    pages={600-612},
    doi={10.1109/TIP.2003.819861}
}

@inproceedings{cogvideox2025yang,
    title={CogVideoX: Text-to-Video Diffusion Models with An Expert Transformer}, 
    author={Zhuoyi Yang and Jiayan Teng and Wendi Zheng and Ming Ding and Shiyu Huang and Jiazheng Xu and Yuanming Yang and Wenyi Hong and Xiaohan Zhang and Guanyu Feng and Da Yin and Yuxuan Zhang and Weihan Wang and Yean Cheng and Bin Xu and Xiaotao Gu and Yuxiao Dong and Jie Tang},
    year={2025},
    eprint={2408.06072},
    url={https://arxiv.org/abs/2408.06072}, 
    booktitle={ICLR}
}

@inproceedings{conmo2025gao,
      title={ConMo: Controllable Motion Disentanglement and Recomposition for Zero-Shot Motion Transfer}, 
      author={Jiayi Gao and Zijin Yin and Changcheng Hua and Yuxin Peng and Kongming Liang and Zhanyu Ma and Jun Guo and Yang Liu},
      year={2025},
      booktitle={CVPR}
}

@article{wan2025wan,
      title={Wan: Open and Advanced Large-Scale Video Generative Models}, 
      author={Team Wan and Ang Wang and Baole Ai and Bin Wen and Chaojie Mao and Chen-Wei Xie and Di Chen and Feiwu Yu and Haiming Zhao and Jianxiao Yang and Jianyuan Zeng and Jiayu Wang and Jingfeng Zhang and Jingren Zhou and Jinkai Wang and Jixuan Chen and Kai Zhu and Kang Zhao and Keyu Yan and Lianghua Huang and Mengyang Feng and Ningyi Zhang and Pandeng Li and Pingyu Wu and Ruihang Chu and Ruili Feng and Shiwei Zhang and Siyang Sun and Tao Fang and Tianxing Wang and Tianyi Gui and Tingyu Weng and Tong Shen and Wei Lin and Wei Wang and Wei Wang and Wenmeng Zhou and Wente Wang and Wenting Shen and Wenyuan Yu and Xianzhong Shi and Xiaoming Huang and Xin Xu and Yan Kou and Yangyu Lv and Yifei Li and Yijing Liu and Yiming Wang and Yingya Zhang and Yitong Huang and Yong Li and You Wu and Yu Liu and Yulin Pan and Yun Zheng and Yuntao Hong and Yupeng Shi and Yutong Feng and Zeyinzi Jiang and Zhen Han and Zhi-Fan Wu and Ziyu Liu},
      journal = {arXiv preprint arXiv:2503.20314},
      year={2025}
}

@article{kong2024hunyuanvideo,
  title={Hunyuanvideo: A systematic framework for large video generative models},
  author={Kong, Weijie and Tian, Qi and Zhang, Zijian and Min, Rox and Dai, Zuozhuo and Zhou, Jin and Xiong, Jiangfeng and Li, Xin and Wu, Bo and Zhang, Jianwei and others},
  journal={arXiv preprint arXiv:2412.03603},
  year={2024}
}

@article{mou2024revideo,
  title={Revideo: Remake a video with motion and content control},
  author={Mou, Chong and Cao, Mingdeng and Wang, Xintao and Zhang, Zhaoyang and Shan, Ying and Zhang, Jian},
  journal={Advances in Neural Information Processing Systems},
  volume={37},
  pages={18481--18505},
  year={2024}
}

@article{xiao2024moft,
  title={Video diffusion models are training-free motion interpreter and controller},
  author={Xiao, Zeqi and Zhou, Yifan and Yang, Shuai and Pan, Xingang},
  journal={Advances in Neural Information Processing Systems},
  volume={37},
  pages={76115--76138},
  year={2024}
}

@inproceedings{burgert2025go,
  title={Go-with-the-flow: Motion-controllable video diffusion models using real-time warped noise},
  author={Burgert, Ryan and Xu, Yuancheng and Xian, Wenqi and Pilarski, Oliver and Clausen, Pascal and He, Mingming and Ma, Li and Deng, Yitong and Li, Lingxiao and Mousavi, Mohsen and others},
  booktitle={Proceedings of the Computer Vision and Pattern Recognition Conference},
  pages={13--23},
  year={2025}
}

@inproceedings{gao2025unity,
  title={Unity in Diversity: Video Editing via Gradient-Latent Purification},
  author={Gao, Junyu and Yang, Kunlin and Yao, Xuan and Hu, Yufan},
  booktitle={Proceedings of the Computer Vision and Pattern Recognition Conference},
  pages={23401--23411},
  year={2025}
}

\begin{IEEEbiography}[{\includegraphics[width=1in,height=1.25in,clip,keepaspectratio]{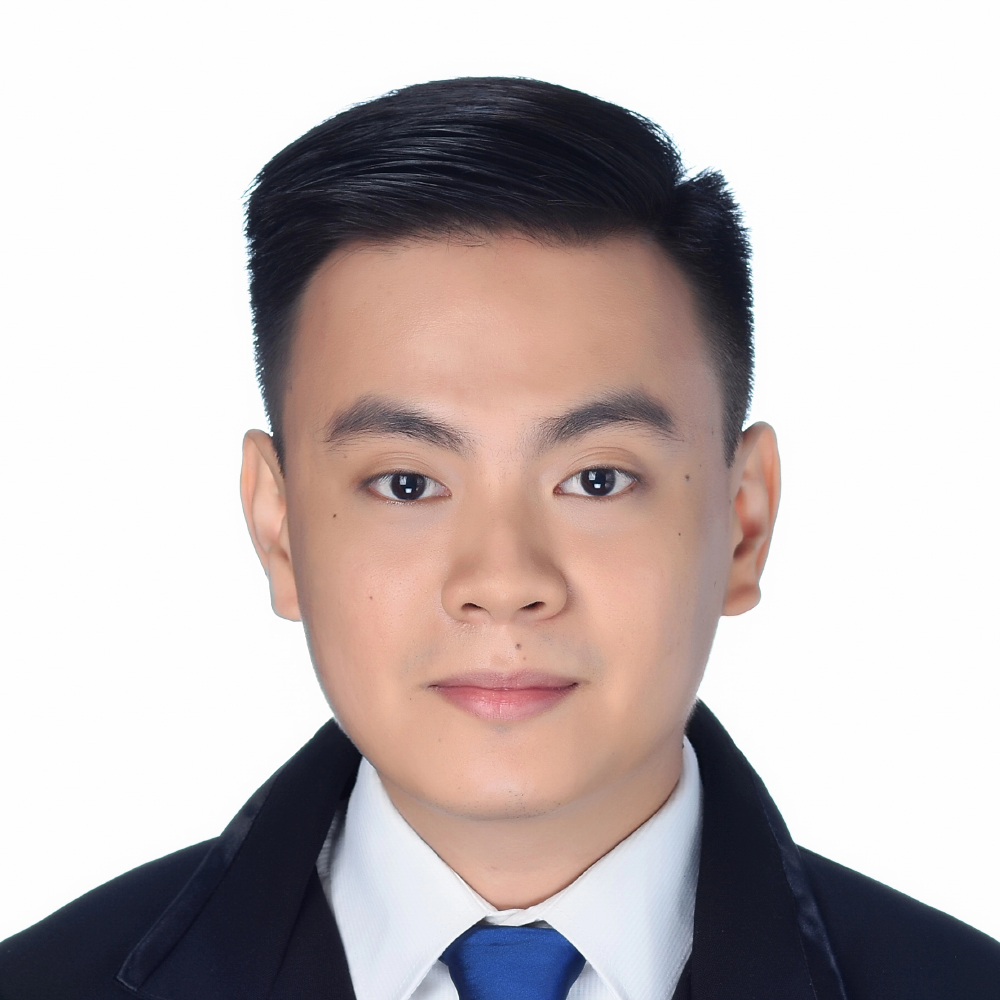}}]{Samuel Teodoro}
is a Ph.D. student at the School of Electrical Engineering, Korea Advanced Institute of Science and Technology (KAIST), Daejeon, South Korea.
He received his B.S. degree in Electrical Engineering from the University of the Philippines Los Baños, Laguna, Philippines in 2018.
His research interests include image and video editing using generative models.
\end{IEEEbiography}

\begin{IEEEbiography}[{\includegraphics[width=1in,height=1.25in,clip,keepaspectratio]{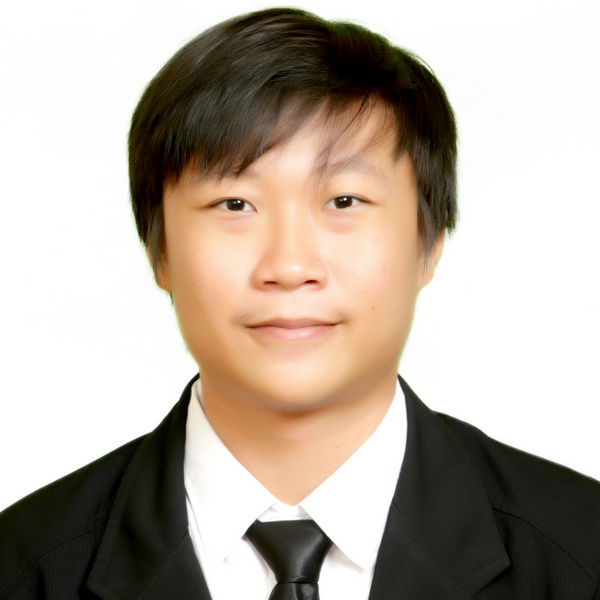}}]{Agus Gunawan} is a Ph.D. student at KAIST, Daejeon, South Korea. He received his B.E. degree in Computer Science from Institut Teknologi Bandung (ITB), Indonesia, in 2019. His research interests lie in low-level vision, including in-the-wild image restoration and style transfer. As of 2025, he has been actively working on image editing using generative models.
\end{IEEEbiography}

\begin{IEEEbiography}[{\includegraphics[width=1in,height=1.25in,clip,keepaspectratio]{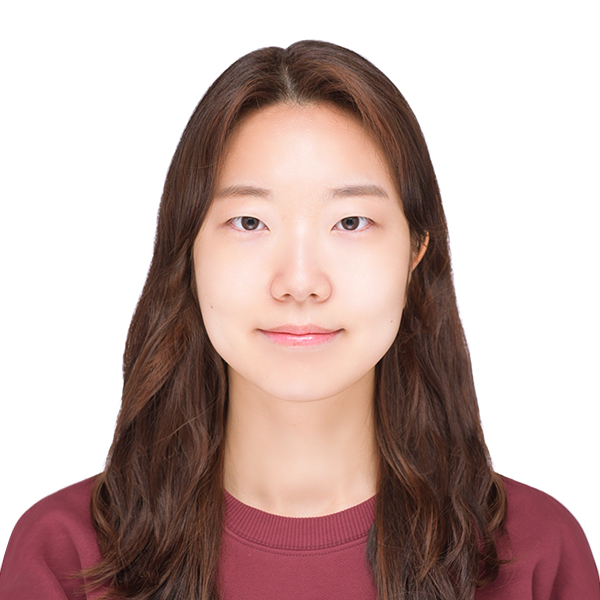}}]{Soo Ye Kim}
is a Research Scientist at Adobe Research (San Jose, USA) since 2022. She received her Ph.D., M.S. and B.S. degrees in Electrical Engineering from KAIST (Daejeon, South Korea) in 2022, 2018 and 2016, respectively. Soo Ye was a Research Intern at Adobe Research in 2021 and a Research Intern and Student Researcher at Google Research in 2020. She is a recipient of the Google PhD Fellowship 2021. As of 2025, Soo Ye has been actively working on image and video editing using generative models. 
\end{IEEEbiography}

\begin{IEEEbiography}[{\includegraphics[width=1in,height=1.25in,clip,keepaspectratio]{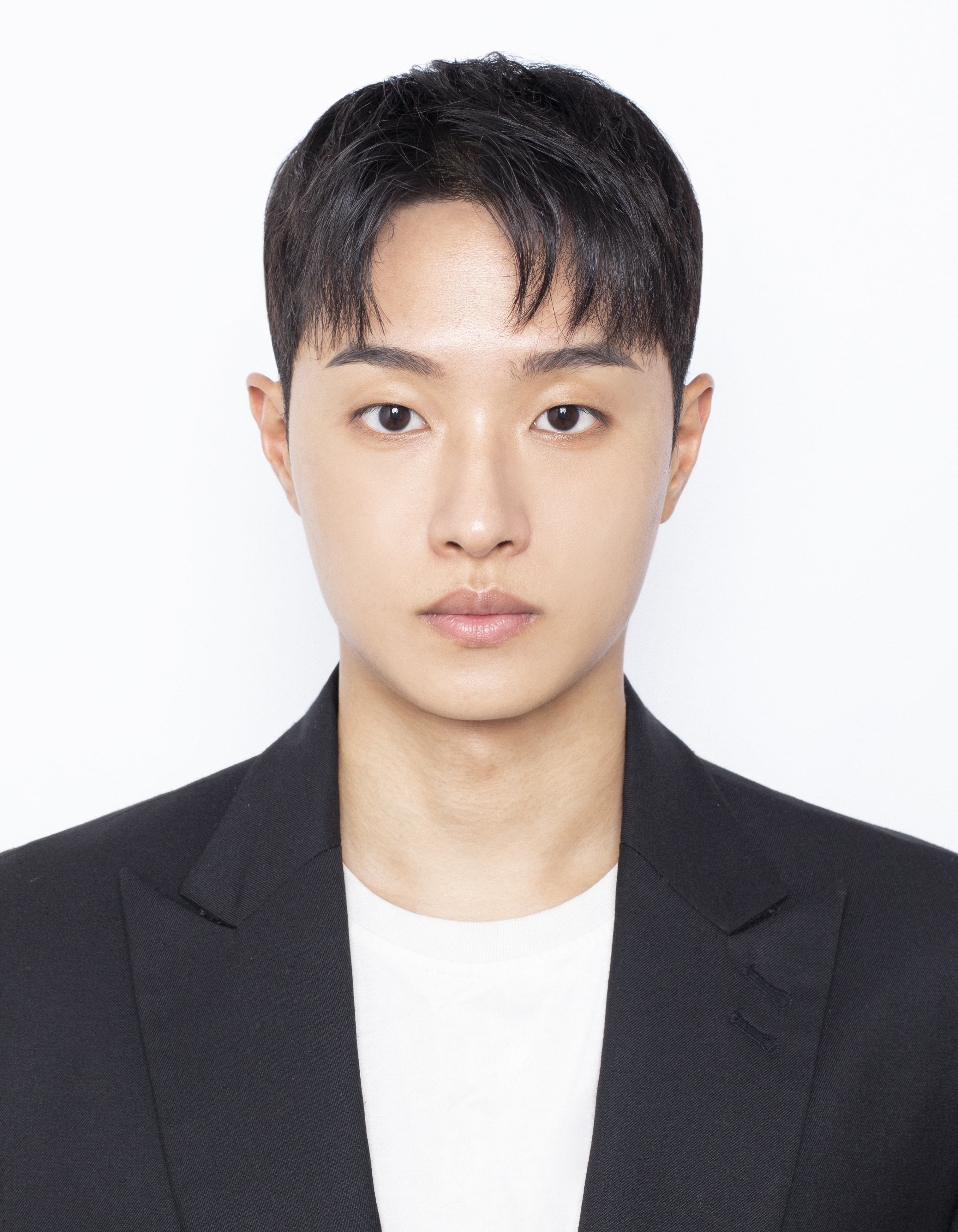}}]{Jihyong Oh}
is an Assistant Professor in the Department of Imaging Science at the Graduate School of Advanced Imaging Science, Multimedia \& Film (GSAIM) at Chung-Ang University (CAU: Seoul, South Korea), and has been leading the Creative Vision and Multimedia Lab (CMLab: https://cmlab.cau.ac.kr/) since September 2023. Prior to this, he was a post-doctoral researcher at VICLAB (https://www.viclab.kaist.ac.kr/) of the Korea Advanced Institute of Science and Technology (KAIST: Daejeon, South Korea). He received his B.E., M.E., and Ph.D. degrees in Electrical Engineering from KAIST, in 2017, 2019, and 2023, respectively. He was a research intern at Meta (Facebook) Reality Labs in 2022. His research primarily focuses on deep-learning-based Low-Level Vision tasks in terms of restoration and enhancement, aiming to provide a visually pleasing experience for users, as well as on 3D, NeRF, Gaussian Splatting, GenAI, etc. He has experience publishing papers at CVPR, ICCV, ECCV, AAAI, TPAMI, TCSVT, Neurocomputing, and Remote Sensing, and has reviewed papers submitted to CVPR, ICCV, ECCV, AAAI, SIGGRAPH, and IEEE TIP/TGRS/Access/TMM. He received outstanding reviewer awards for ICCV 2021 and CVPR 2024.
\end{IEEEbiography}

\begin{IEEEbiography}[{\includegraphics[width=1in,height=1.25in,clip,keepaspectratio]{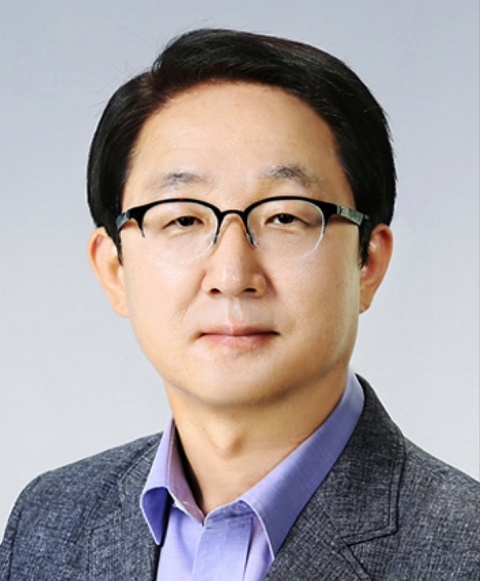}}]{Munchurl Kim}
(M’99–SM’13) received the B.E. degree in electronics from Kyungpook National University, Daegu, South Korea, in 1989, and the M.E. and Ph.D. degrees in electrical and computer engineering from the University of Florida, Gainesville, in 1992 and 1996, respectively. He joined the Electronics and Telecommunications Research Institute, Daejeon, South Korea, as a Senior Research Staff Member, where he led the Realistic Broadcasting Media Research Team. In 2001, he joined the School of Engineering, Information and Communications University (ICU), Daejeon, as an Assistant Professor. Since 2009, he has been with the School of Electrical Engineering, Korea Advanced Institute of Science and Technology (KAIST), Daejeon, where he is currently a Full Professor. He has published 212 (189) international (domestic) journal and conference papers. He holds 243 registered domestic and international patents in image restoration and video coding. He was invited to give a keynote speech on the evolution of conventional and deep video compression technologies in 2020 Multimedia Modeling Conference. He has served the technical program chairs of Visual Communications and Image Processing (VCIP) 2023 and ACM Multimedia Asia 2023.
\end{IEEEbiography}

\vfill

\end{document}